\documentclass[opre,nonblindrev]{informs3}

\DoubleSpacedXI

\usepackage{endnotes}
\let\footnote=\endnote
\let\enotesize=\normalsize
\def\notesname{Endnotes}%
\def\makeenmark{\hbox to1.275em{\theenmark.\enskip\hss}}
\def\enoteformat{\rightskip0pt\leftskip0pt\parindent=1.275em
  \leavevmode\llap{\makeenmark}}

\usepackage[greek,english]{babel}
\usepackage[utf8]{inputenc} 
\usepackage[T1]{fontenc}    
\usepackage{hyperref}       
\usepackage{url}            
\usepackage{booktabs}       
\usepackage{nicefrac}       
\usepackage{microtype}      

\usepackage[export]{adjustbox}
\usepackage{float}
\usepackage{xcolor}
\usepackage{bbm,bm}
\usepackage{mdframed}
\usepackage{relsize}
\usepackage{tikz}
\usepackage{pgfplots}
\usepackage{caption,subcaption}
\usepackage{multirow}
\usepackage{dsfont}
\usepackage{amsmath, amssymb,amsfonts}
\newtheorem{prop}{Proposition}
\newtheorem{theo}{Theorem}
\newtheorem{lem}{Lemma}

\newcommand{\md}[1]{\mathbb{#1}}
\newcommand{\ma}[1]{\mathcal{#1}}
\DeclareMathOperator\diag{Diag}
\def\*#1{\bm{#1}}

\usepackage[ruled,vlined,linesnumbered,resetcount]{algorithm2e}
\RestyleAlgo{boxruled}
\LinesNumbered
\usepackage{algorithmic}

\usepackage{natbib}
 \bibpunct[, ]{(}{)}{,}{a}{}{,}%
 \def\bibfont{\small}%
\TheoremsNumberedThrough     
\ECRepeatTheorems

\EquationsNumberedThrough    

\newcommand{\svc}{\texttt{svar\_cutplane}}
\newcommand{\svh}{\texttt{svar\_heuristic}}
\newcommand{\svhc}{\texttt{svar\_hybrid}}
\newcommand{\sr}{\texttt{static\_cutplane}}
\newcommand{\rsnOne}{\texttt{sum\_of\_norms\_l1}}
\newcommand{\rsnOneLasso}{\texttt{sum\_of\_norms\_l1\_lasso}}
\newcommand{\rsnTwo}{\texttt{sum\_of\_norms\_l2}}
\newcommand{\rsnTwoLasso}{\texttt{sum\_of\_norms\_l2\_lasso}}
\newcommand{\rsnP}{\texttt{sum\_of\_norms\_lp}}
\newcommand{\rsnPLasso}{\texttt{sum\_of\_norms\_lp\_lasso}}
\newcommand{\bisection}{\texttt{bisection}}
\newcommand\E{\mathbb{E}}

\newcommand{\rev}{\color{black}}

\newenvironment{revi}{\color{black}}{}

\begin{document}


\RUNAUTHOR{Bertsimas and Digalakis Jr. and Li and Skali Lami}

\RUNTITLE{Slowly Varying Regression under Sparsity}

\TITLE{Slowly Varying Regression under Sparsity}


\ARTICLEAUTHORS{
\AUTHOR{Dimitris Bertsimas}
\AFF{Sloan School of Management, Massachusetts Institute of Technology, Cambridge, MA, USA}
\AUTHOR{Vassilis Digalakis Jr}
\AFF{Department of Information Systems and Operations Management, HEC Paris, 78350 Jouy-en-Josas, France\\
Operations Research Center, Massachusetts Institute of Technology, Cambridge, MA, USA}
\AUTHOR{Michael Lingzhi Li}
\AFF{Technology and Operations Management Unit,
Harvard Business School, Boston, MA, USA}
\AUTHOR{Omar Skali Lami\footnote{The author participated in discussions of an early version of the paper.}}
\AFF{McKinsey \& Company, Boston, MA, USA}
}

\ABSTRACT{
We introduce the framework of slowly varying regression under sparsity, which allows sparse regression models to vary slowly and sparsely. 
We formulate the problem of parameter estimation as a mixed integer optimization problem and demonstrate that it can be reformulated exactly as a binary convex optimization problem through a novel relaxation.
The relaxation utilizes a new equality on Moore-Penrose inverses that convexifies the non-convex objective function while coinciding with the original objective on all feasible binary points. 
This allows us to solve the problem significantly more efficiently and to provable optimality using a cutting plane-type algorithm. 
We develop a highly optimized implementation of such algorithm, which substantially improves upon the asymptotic computational complexity of a straightforward implementation. 
We further develop a fast heuristic method that is guaranteed to produce a feasible solution and, as we empirically illustrate, generates high-quality warm-start solutions for the binary optimization problem. 
\begin{revi}To tune the framework's hyperparameters, we propose a practical procedure relying on binary search that, under certain assumptions, is guaranteed to recover the true model parameters.
 \end{revi}
We show, on both synthetic and real-world datasets, that the resulting algorithm outperforms competing formulations in comparable times across a variety of metrics including estimation accuracy, predictive power, and computational time,
and is highly scalable, enabling us to train models with 10,000s of parameters.
We make our implementation available open-source at \url{https://github.com/vvdigalakis/SSVRegression.git}.
}%


\maketitle


\section{Introduction} \label{sec:introduction}

We introduce the framework of \textit{slowly varying regression under sparsity} (SSVR), which addresses a large number of problems in machine learning where the underlying model is sparse and varies slowly and sparsely.
This in particular includes problems with \textit{temporally or spatially varying structure}. For example, in the temporal case, the factors important in predicting the energy consumption in a building can vary depending on the hour of the day or the period of the year. In the spatial case, the factors that affect house prices can differ by neighborhood. 

\begin{revi}
In both cases, a modeler may be motivated to use different models for each time period, spatial area, or, more generally, ``vertex.'' However, using separate models ignores the innate dependence across different vertices (e.g., energy consumption is strongly affected by daily and seasonal patterns) and creates interpretability issues if the separate models turn out to be notably different. Further, separate models require substantial data to be available for each vertex, which is often difficult. In this work, we address the need to capture this slowly and sparsely varying structure in a global way. 
\end{revi}

\subsection{Slowly Varying Regression under Sparsity: An Initial Formulation} \label{ssec:initialformulation}

\begin{figure}[ht!] 
    \centering
    \begin{subfigure}[b]{0.4\textwidth}
        \centering
        \resizebox{0.75\linewidth}{!}{
        \begin{tikzpicture}[scale=0.75]
           
            \node[] (0) at (0,0) {}; \node[draw,circle,label=below:$\bm{\beta^1}$] (1) at (0,3) {$t=1$};
            \node[draw,circle] (2) at (2,3) [label=below:$\bm{\beta^2}$] {$t=2$};
            \node[] (3) at (4,3) {$\dots$};
            \node[draw,circle] (4) at (6,3) [label=below:$\bm{\beta^T}$] {$t=T$};
            
            \draw (1) -- (2) -- (3) -- (4);
        \end{tikzpicture}
        }
    \end{subfigure}
    \hfill
    \begin{subfigure}[b]{0.55\textwidth}
        \centering
        \resizebox{0.5\linewidth}{!}{
        \begin{tikzpicture}[scale=0.6]
                \node[draw,circle,label=below:$\bm{\beta^S}$] (1) at (0,2) {S};
                \node[draw,circle,label=below:$\bm{\beta^W}$] (2) at (2,0) {W};
                \node[draw,circle,label=below:$\bm{\beta^{NW}}$] (3) at (4,0) {NW};
                \node[draw,circle,label=below:$\bm{\beta^{N}}$] (4) at (6,2) {N};
                \node[draw,circle,label=below:$\bm{\beta^{NN}}$] (5) at (8,2) {NN};
                \node[draw,circle,label=below:$\bm{\beta^{NE}}$] (6) at (4,4) {NE};
                \node[draw,circle,label=below:$\bm{\beta^{E}}$] (7) at (2,4) {E};
                
                \draw (1) -- (2);
                \draw (2) -- (3);
                \draw (3) -- (4);
                \draw (4) -- (5);
                \draw (4) -- (6);
                \draw (6) -- (7);
                \draw (1) -- (7);
            \end{tikzpicture}
            }
    \end{subfigure}
    \caption{Examples of similarity graphs. 
    In the temporal case (left), the different regressions are applied across T consecutive time periods. (Such examples are considered in Section \ref{ssec:realworld-temporal}.)
    In the spatial case (right), the different regressions are applied across 7 spatial areas (S, E, W, NE, NW, N, NN) with the given similarity structure. (The graph corresponds to one of the experiments described in Section \ref{ssec:realworld-spatial}.)
    }
    \label{fig:similarity-graphs}
\end{figure}
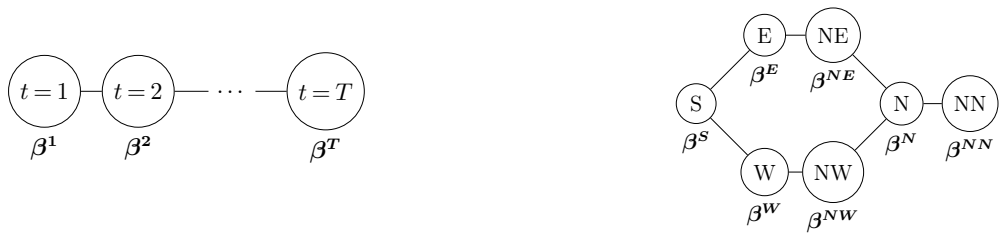

Formally, we consider a multiple regression problem with $N$ cases having features $\bm{X}^1,\dots, \bm{X}^T$, where $\bm{X}^t \in \md{R}^{N \times D}$ for $t \in [T] := \{1,\dots,T\},$ and outcomes $\bm{y}^1,\dots, \bm{y}^T$, where $\bm{y}^t \in \md{R}^N$ for $t \in [T]$. An SSVR model assumes that the regression coefficients and relevant features (i.e., features that correspond to nonzero coefficients) change slowly between pairs of regressions $(s,t) \in E \subseteq [T] \times [T]$ that are considered \emph{similar}. Two prominent applications include temporally varying regression and spatially varying regression. In the \emph{temporal} case, the regressions are scattered over $T$ consecutive time periods, and regressions between two consecutive time periods are considered to be similar. In the \emph{spatial} case, the regressions are conducted over $T$ spatial areas, some of which are adjacent to each other, and it is common to assume that regressions in adjacent areas have to be similar. 
More generally, the $T$ regressions are conducted over a graph $G$ with vertices $V$ of size $|V|=T$. For $v,w \in V$, the edge $(v,w)$ is in the set of edges $E$ if and only if $v$ and $w$ are considered to be similar. Figure \ref{fig:similarity-graphs} presents examples of similarity graphs.

The SSVR problem can be formulated as below:
\begin{align}
        \underset{\bm{\beta}^1,\cdots,\bm{\beta}^T}{\min} \quad & 
         \sum_{t=1}^T \left\|\bm{y}^t-\bm{X}^t\bm{\beta}^t\right\|_2^2
        + \lambda_{\beta} \sum_{t=1}^T \| \bm{\beta}^t \|_2^2 
        + \lambda_{\delta} \sum_{(s,t) \in E} \| \bm{\beta}^t - \bm{\beta}^s \|_2^2 \label{eq:obj_orig}\\
       \text{s.t.} \qquad &
        \left| \text{Supp}(\bm{\beta}^t) \right| \leq K_{\text{L}}, \quad  \forall t\in [T], \label{eq:indsparse}\\
        & \left| \bigcup_{t=1}^T \text{Supp}(\bm{\beta}^t) \right| \leq K_{\text{G}} , \label{eq:globsparse}\\
        &  \sum_{(s,t) \in E} \left| \text{Supp}(\bm{\beta}^t) \triangle \text{Supp}(\bm{\beta}^s) \right| \leq K_{\text{C}}, \label{eq:sparsevary}
\end{align}
where $\text{Supp}(\bm{\beta})$ denotes the set that corresponds to the support of vector $\bm{\beta}$ and $S_1 \triangle S_2$ denotes the symmetric difference of sets $S_1,S_2$. The objective function \eqref{eq:obj_orig} penalizes both the least-squares loss of the $T$ regressions and the $\ell_2$ coefficient distance between regressions that are similar with magnitude $\lambda_\delta$. We also introduce a further $\ell_2$ regularization term of magnitude $\lambda_\beta$ for robustness purposes (see, e.g., \cite{xu2009robust}). There are three types of constraints on the regression coefficients $\bm{\beta}^t$:
\begin{itemize}
    \item[-] \textit{Local Sparsity}: Each regression has at most $K_{\text{L}}$ relevant features (constraint \eqref{eq:indsparse}).
    \item[-] \textit{Global Sparsity}: There are at most $K_{\text{G}}$ relevant features across all $T$ regressions (constraint \eqref{eq:globsparse}).
    \item[-] \textit{Sparsely Varying Support}: There is a difference of at most $K_{\text{C}}$ relevant features among similar regressions $s,t$ across all pairs of similar regressions (constraint \eqref{eq:sparsevary}).
\end{itemize}

\begin{revi}
For consistency, $K_{\text{L}}, K_{\text{G}}, K_{\text{C}}$ satisfy $K_{\text{L}} \leq K_{\text{G}} \leq D$ and $K_{\text{C}} \leq 2 K_{\text{L}} T$ and $2(K_{\text{G}}-K_{\text{L}}) \leq K_{\text{C}}$. 
\end{revi}
This exact formulation is generally considered infeasible beyond toy scales ($D \leq 10^2,\ T \leq 10$) due to the combinatorial complexity of the sparsity constraints. Therefore, many authors have proposed various relaxations in order to solve variants of this problem, including fused lasso \citep{tibshirani2005sparsity} and sum-of-norms regularization \citep{ohlsson2010segmentation}; we review such approaches in Section \ref{ssec:literature-slowlyvarying}. 
Our key contribution in this paper is to show that this general problem can be reformulated as a binary convex optimization problem, which then can be solved efficiently using a cutting plane-type algorithm. This reformulation is primarily enabled by an exact smooth relaxation of the solution under sparsity constraints, which, to the best of our knowledge, has not appeared in prior literature. Furthermore, we discuss in Section \ref{ssec:literature-sparse} how the reformulation directly extends to any sparse quadratic convex problem of a general form, making the relaxation generally applicable. 

\subsection{Contributions and Outline}
\begin{revi}
We now more concretely summarize our contributions from a modeling, theoretical, algorithmic, and computational (practitioner's) perspective:
\begin{itemize}
	\item From a \textit{modeling} standpoint, we introduce the slowly varying regression under sparsity framework, which addresses regression problems with sparse and slowly varying structure.

	\item From a \textit{theoretical} standpoint, we propose a new way of solving the underlying optimization problem, which extends to a more general class of sparse quadratic problems. We reformulate the problem exactly as a binary convex optimization problem through a novel relaxation of the objective function. The proposed relaxation relies upon a new equality on Moore-Penrose inverses that convexifies the nonconvex objective function while coinciding on all feasible binary points. 
	
	\item From an \textit{algorithmic} standpoint, firstly, leveraging the convexity of the reformulated problem, we develop a cutting plane-type algorithm that enables us to solve the binary convex optimization problem at hand to provable optimality. By exploiting the structure of the problem, we efficiently implement the proposed algorithm and substantially improve upon the asymptotic computational complexity of a straightforward implementation. Secondly, we develop a fast heuristic algorithm, which is guaranteed to produce a feasible solution and, as we empirically show, computes high-quality, warm-start solutions to the binary convex optimization problem. Thirdly, we propose a practical hyperparameter tuning procedure relying on binary search that, under certain assumptions, is guaranteed to recover the true model hyperparameters.
	
	\item From a \textit{computational} standpoint, we thoroughly evaluate the proposed method on both synthetic and real-world data. We show that the proposed algorithm outperforms competing formulations across a variety of metrics including estimation accuracy, predictive power, and computational time, and is highly scalable, enabling us to train models with 10,000s of parameters. In real-world experiments, we further illustrate how the resulting SSVR model can provide insights into the problem at hand. We make our implementation available \textit{open-source} at \url{https://github.com/vvdigalakis/SSVRegression.git}. To facilitate the use of the proposed framework by practitioners, all proposed algorithms can be run through a single line of code, and the learned models are provided in an intuitive and interpretable way.
\end{itemize}

The outline of the paper is as follows. In the remainder of Section \ref{sec:introduction}, we summarize the relevant literature. In Section \ref{sec:mioformulation}, we formulate the SSVR problem as a mixed-integer optimization problem. In Section \ref{sec:binaryreformulation}, we develop the proposed relaxation of the objective function and reformulate the problem exactly as a binary convex optimization problem. In Section \ref{sec:relaxation}, we explore the properties of the proposed relaxation, build intuition on why it works, and studiy how it can be extended to general quadratic models. In Section \ref{sec:cuttingplane}, we develop the proposed exact cutting plane type algorithm and discuss how to efficiently implement it. In Section \ref{sec:heuristic}, we design the proposed fast heuristic algorithm. In Section \ref{sec:tuning}, we explain how to tune the SSVR model practically, while also providing theoretical guarantees albeit in limited scenarios. Finally, in Sections \ref{sec:synthetic} and \ref{sec:realworld}, we present our experimental evaluations on synthetic and real-world data, respectively. 

\end{revi}

\subsection{Relevant Literature: Slowly Varying Regression} \label{ssec:literature-slowlyvarying}

The practical relevance of the notion of \textit{slowly (or smoothly) varying regression} is evident by the significant amount of impactful work in the field, dating back to at least \cite{hastie1993varying}, who study linear regression models whose coefficients are allowed to change smoothly with the value of other variables, and \cite{bertsimas1999estimation}, who solve the nonparametric regression estimation problem when the underlying regression function is Lipschitz continuous; see, e.g., the book by \cite{eubank1999nonparametric} for a comprehensive review and e.g. \cite{phillips2007regression, 9060978} for applications in Econometrics and Electronics.

The popularity of slowly varying regression models peaked following the work of \cite{tibshirani2005sparsity} on the \textit{fused lasso}, which proposed to augment the standard lasso \citep{tibshirani1996regression} objective with an $\ell_1$ penalty term on the difference between successive regression coefficients $|\beta^t-\beta^{t-1}|$ to account for pairwise similarity. The algorithms for solving the resulting problems were improved by many works including \citep{tibshirani2011solution,wytock2014fast} while other works considered different convex regularizers or extended the formulation to other settings such as change point detection \citep{alaiz2013group, rojas2014change, bleakley2011group}.

The works that are most closely related to ours include the sum-of-norms regularization approaches by \cite{ohlsson2010segmentation,hallac2015network,hallac2017snapvx}, the total variation regularization approach by \cite{wytocktime}, and the heuristic splicing approach by \cite{zhang2023splicing}. \cite{ohlsson2010segmentation}, in particular, consider the time-varying linear regression optimization problem and their work can naturally be extended to the (more general) graph case that we consider in this paper as follows:
\begin{equation} \label{eq:boyd}
    \underset{\bm{\beta}^1,\cdots,\bm{\beta}^T}{\min} \quad 
         \sum_{t=1}^T \left\|\bm{y}^t-\bm{X}^t\bm{\beta}^t\right\|_2^2
        + \lambda_{\delta} \sum_{(s,t) \in E} \| \bm{\beta}^t - \bm{\beta}^s \|_p,    
\end{equation}
where $p \in \{1,2\}$. Our work significantly differs from this line of work by exactly imposing sparsity and sparse variation in the coefficients, hence providing more control to the modeler, and, further, by utilizing a smooth $\ell_2$ penalty term on the difference between the regression coefficients $\|\bm{\beta}^t-\bm{\beta}^s\|_2^2$.

Another stream of related work in the application of \textit{spatially varying regression} are spatially varying coefficient (SVC) models. Instead of imposing a strict constraint on the degree of variability, SVC models focus on identifying the heterogeneity in coefficient estimates varying across space. Notable methods include the spatial expansion method \citep{casetti1972generating}, geographically weighted regression \citep{brunsdon1996geographically}, and Bayesian SVC models \citep{besag1991bayesian}. 

\subsection{Relevant Literature: Solving Sparse Quadratic Models} \label{ssec:literature-sparse}

Problems with sparsity constraints have long been of interest in many areas ranging from machine learning to facility location and portfolio selection, as sparsity improves robustness to data noise and increases interpretability for better decision-making. However, due to their combinatorial complexity, it has long been thought that exact sparse formulations are not scalable, and $\ell_1$ regularization formulations (e.g. fused lasso in Section \ref{ssec:literature-slowlyvarying}) have been widely used as surrogates.

In recent years, a growing volume of work has challenged the aforementioned paradigm.
\cite{bertsimas2016best,bertsimas2020sparse,hazimeh2022sparse} solve standard sparse regression problems with design matrix $\bm{X} \in \md{R}^{N \times D}$, responses $\bm{y} \in \md{R}^N$, and $\ell_2$ regularization, outlined as
\begin{align}
    \min_{\bm{\beta}:\|\bm{\beta}\|_0\leq K} \frac{1}{2}\|\bm{y}-\bm{X}\bm{\beta}\|_2^2 + \lambda_\beta \|\bm{\beta}\|_2^2,\label{eq:sparse_reg}
\end{align}
at scale, using techniques from mixed-integer optimization. Beyond sparse regression, \cite{wei2022ideal} study the convexification of a class of convex optimization problems with indicator variables and combinatorial constraints on the indicators, whereas, in a recent work motivated by probabilistic graphical models, \cite{liu2023graph} study convex quadratic optimization problems with indicator variables when the matrix defining the quadratic term in the objective is sparse.

The work that is most closely related to ours is by \cite{bertsimas2020sparse}, who utilize binary variables $\bm{z} \in \{0,1\}^D$ and reformulate Problem \eqref{eq:sparse_reg} as
\begin{align}
    \min_{\bm{z}\in\{0,1\}^D,\sum_{i=1}^D z_i \leq K} \min_{\bm{\beta}} \frac{1}{2}\|\bm{y}-\bm{X}\bm{Z}\bm{\beta}\|_2^2 + \lambda_\beta \|\bm{\beta}\|_2^2, \label{eq:reformulation_sparse_reg}
\end{align}
where $\bm{Z}=\diag(z_1,\dots,z_D)$. The authors then show that the inner minimization problem can be solved in a closed form that results in a convex binary formulation for the outer problem:
\begin{align}
    \min_{\bm{z}\in\{0,1\}^D,\sum_{i=1}^D z_i \leq K} \bm{y}^T\left(\bm{I}_N+\frac{1}{2\lambda_\beta}\sum_{i=1}^D z_i\bm{X}_i\bm{X}_i^T\right)^{-1}\bm{y}, \label{eq:inner_solved_sparse_reg}
\end{align}
where $\bm{X}_i$ is the $i$th column of design matrix $\bm{X}$. The resulting problem can then be solved efficiently to very large scales ($N,D \approx 10^4$) using a cutting plane-type algorithm. This significant breakthrough raised the limits of scaling exact sparse methods by multiple orders of magnitude. 
However, the transformation presented above seems to be quite fortuitous. For example, the reformulation in \cite{bertsimas2020sparse} relied on rewriting $\bm{\beta}$ as $\bm{Z}\bm{\beta}$ in the first term but not the second term of Equation \eqref{eq:reformulation_sparse_reg}. There appears to be no systematic reason why doing so is necessary to result in the final convex binary formulation in Equation \eqref{eq:inner_solved_sparse_reg}, and even fewer hints on how we could systematically apply this methodology to other problems. 

This paper aims to uncover the underlying key ingredients to allow such transformations through the study of the SSVR problem, which is a generalization of the standard sparse regression problem of Equation \eqref{eq:sparse_reg}. Moreover, the framework we present here directly extends to any sparse quadratic convex problem of the form
\begin{equation}
    \min_{\bm{x} \in \mathcal{X}} \quad \frac{1}{2}\bm{x}^T(\bm{M}+\lambda \bm{I})\bm{x}-\bm{\mu}^T\bm{x},
    \label{eq:sparse-quadratic}
\end{equation}
where $\bm{\mu} \in \md{R}^D$, $\bm{x} \in \md{R}^D$ are the decision variables, $\bm{M}\in \md{R}^{D \times D}$ is any positive semidefinite matrix, and $\mathcal{X}$ is the feasible set with sparsity-imposing constraints, e.g. $\mathcal{X} = \{ \bm{x} \in \md{R}^D: \|\bm{x}\|_0 \leq K \}$. 

Overall, the proposed framework lies between \cite{bertsimas2020sparse} and \cite{bertsimas2021unified}. 
\textit{On one extreme}, we consider a general quadratic regression framework that contains the sparse regression problem in \cite{bertsimas2020sparse} as a special case and draw insights from optimization and convex relaxations to efficiently solve the resulting inner problem.
\textit{On the other extreme}, \cite{bertsimas2021unified} investigate general sparse mixed-integer optimization problems with logical constraints and develop an outer approximation scheme whereby the solution to the inner problem involves solving an optimization problem in each iteration; in contrast, we focus on sparse mixed-integer optimization problems where the inner problem is an unconstrained quadratic optimization problem and develop a general efficient procedure for optimizing the resulting problem.

\section{An MIO Formulation} \label{sec:mioformulation}
In this section, we develop a mixed-integer optimization (MIO) formulation for the problem defined in (\ref{eq:obj_orig})--(\ref{eq:sparsevary}). To do so, we encode each of the aforementioned constraints using auxiliary binary variables and new constraints.

\paragraph{Local Sparsity.}
First, we introduce binary variables $\bm{z}^t$ encoding the support of the coefficients $\bm{\beta}^t,\ \forall t,$ as $ z^t_d = 0 \Rightarrow \beta^t_d = 0, \ \forall t \in [T],d \in [D].$ Then, the requirement that the number of nonzero coefficients at each vertex is less than $K_{\text{L}}$ can be expressed as $ \sum_{d=1}^D z^t_d \leq K_{\text{L}}, \ \forall t \in [T]. $

\paragraph{Global Sparsity.}
Similarly, we introduce binary variables encoding the union of supports over vertices. We require that $s_d$ is set to $1$ if $z^t_d$ is set to $1$ at least once over all vertices, i.e., $s_d \geq z^t_d, \ \forall t \in [T],d \in [D].$ Then, we have $ \sum_{d=1}^D s_{d} \leq K_{\text{G}}.$

\paragraph{Sparsely Varying Support.}
To be able to capture the sparsely varying support requirement, we introduce another set of binary variables
$ w^{t,s}_d = 0 \Rightarrow \|\beta^t_d\|_0 = \|\beta^s_d\|_0 \Rightarrow z^t_d = z^s_d, \ \forall (s,t) \in E, d \in [D].$ This can be rewritten as
$w^{t,s}_d \geq z^t_d - z^s_d$ and $w^{t,s}_d \geq z^s_d - z^t_d, \ \forall (s,t) \in E, d \in [D].$ We then require that $ \sum_{(s,t) \in E} \sum_{d=1}^D w^{t,s}_d \leq K_{\text{C}}.$

\paragraph{Overall Formulation.}
With these helper binary variables and constraints, we can now rewrite the original problem defined in (\ref{eq:obj_orig})--(\ref{eq:sparsevary}) as follows:
\begin{align}
        \underset{\substack{\bm{z} \in \{0,1\}^{TD}, \bm{s} \in \{0,1\}^{D},\\ \bm{w}  \in \{0,1\}^{|E| D}}}{\min}\;\;\underset{\bm{\beta}}{\min} \quad & 
         \sum_{t=1}^T \left\|\bm{y}^t-\bm{X}^t\bm{Z}^t\bm{\beta}^t\right\|_2^2
        + \lambda_{\beta} \sum_{t=1}^T \| \bm{Z}^t\bm{\beta}^t \|_2^2 
        + \lambda_{\delta} \sum_{(s,t) \in E} \| \bm{Z}^t\bm{\beta}^t - \bm{Z}^s\bm{\beta}^s \|_2^2 \label{eq:obj_mio}\\
       \text{s.t.} \qquad &
        \sum_{d=1}^D z^t_d \leq K_{\text{L}}, \qquad \forall t \in [T], \label{eq:c1_mio}\\
        &     s_d \geq z^t_d, \qquad \forall t \in [T],d \in [D]
 , \label{eq:c2_mio}\\
        &         \sum_{d=1}^D s_{d} \leq K_{\text{G}}, \label{eq:c3_mio}\\ 
        & w^{t,s}_d \geq z^t_d - z^s_d,  \qquad \forall (s,t) \in E, d \in [D], \label{eq:c4_mio}\\ 
        & w^{t,s}_d \geq z^s_d - z^t_d,  \qquad \forall (s,t) \in E, d \in [D], \label{eq:c5_mio}\\
        &  \sum_{(s,t) \in E} \sum_{d=1}^D w^{t,s}_d \leq K_{\text{C}}, \label{eq:c6_mio}
\end{align}
where $\bm{Z}^t=\diag(z^t_1,\cdots,z^t_D)$ are diagonal binary matrices of $\bm{z}$ variables. For convenience, we denote the optimization problem over $\bm{z},\bm{s},\bm{w}$ as the outer optimization problem, while the optimization over $\bm{\beta}$ as the inner optimization problem.


\section{The Binary Convex Reformulation} \label{sec:binaryreformulation}

In this section, we reformulate the mixed-integer optimization problem defined in \eqref{eq:obj_mio}-\eqref{eq:c6_mio} as a pure-binary convex optimization problem. First, we note the following lemma:
\begin{lem}
\label{lem:reform}
The MIO optimization problem defined in \eqref{eq:obj_mio}-\eqref{eq:c6_mio} is equivalent to the following optimization problem:
\begin{align*}
        \underset{\bm{z},\bm{s}, \bm{w} \in \ma{Z}}{\min}\;\;\underset{\bm{\beta}}{\min} \quad & 
         c(\bm{z},\bm{\beta}) := \frac{1}{2}\bm{\beta}^{\top}(\bm{Z}(\bm{M}+\lambda_\beta \bm{I})\bm{Z})\bm{\beta} -\bm{\mu}^{\top} \bm{Z} \bm{\beta}, 
\end{align*}
where $\bm{\beta}=(\bm{\beta}^1,\cdots,\bm{\beta}^T)$, $\bm{Z}=\diag(\bm{z}^1,\cdots,\bm{z}^T)$, and $\ma{Z}$ is the polyhedral feasible set as defined by the binary constraints on $\bm{z},\bm{s}, \bm{w}$ and \eqref{eq:c1_mio}-\eqref{eq:c6_mio}. $\bm{M} \in \md{R}^{TD \times TD}$ and $\bm{\mu}=(\bm{\mu}^1,\cdots,\bm{\mu}^T)$ are defined as:
\begin{align*}
    \bm{M}^{t,s}_{i,j}& =
    \left[ \sum_n \left( X^t_{n,i}\right)^2 \ + \ d^t\lambda_{\delta} \right] \ \mathbbm{1}_{\left( s=t \text{ and } i = j \right)} \ + \
    \left[ \sum_n X^t_{n,i} X^t_{n,j} \right] \ \mathbbm{1}_{\left(s=t \text{ and } i \neq j\right)} \ - \
    \lambda_{\delta} \ \mathbbm{1}_{\left((s,t) \in E \text{ and } i=j\right)}, \\
    \bm{\mu}^t & = 
    (\bm{X}^{t})^{\top}\bm{y}^t,
\end{align*}
where $d^t$ denotes the degree of vertex $t$. Furthermore, $\bm{M}$ is a positive semi-definite matrix. 
\end{lem} 
The proof is given in Appendix \ref{sec:appendix-lem:reform}. With this formulation, we can solve the inner problem easily using the first order condition and reduce the problem to a binary optimization problem. Recall that the Moore-Penrose pseudoinverse $\bm{A}^\dagger \in \md{R}^{n\times m}$ of $\bm{A} \in \md{R}^{m\times n}$ is the unique matrix that satisfies: 1. $\bm{A}^\dagger\bm{A}\bm{A}^\dagger = \bm{A}^\dagger$, 2. $\bm{A}\bm{A}^\dagger\bm{A} = \bm{A}$, 3. $(\bm{A} \bm{A}^\dagger)^*=\bm{A}\bm{A}^\dagger$, 4. $(\bm{A}^\dagger\bm{A})^*=\bm{A}^\dagger\bm{A}$, where $*$ is the Hermitian operator with $\bm{A}^*_{ij}=\overline{\bm{A}_{ji}}$. We have:
\begin{lem}
\label{lem:first_order}
Denote $\bm{\beta}^*(\bm{z}) =\argmin_{\bm{\beta}}c(\bm{z},\bm{\beta})$. Then, we have
\begin{equation} \label{eq:beta_star_pseudoinv}
    \bm{\beta}^*(\bm{z}) = (\bm{Z}(\bm{M}+\lambda_\beta \bm{I})\bm{Z})^\dagger \bm{Z}\bm{\mu}.
\end{equation}
Furthermore, it holds that
\[\min_{\bm{z},\bm{s}, \bm{w} \in \ma{Z}} \min_{\bm{\beta}} c(\bm{z},\bm{\beta})=\min_{\bm{z},\bm{s}, \bm{w} \in \ma{Z}}-\frac{\bm{\mu}^{\top}\bm{\beta}^*(\bm{z})}{2}.\]
\end{lem} 
The proof is given in Appendix \ref{sec:appendix-lem:first_order}. Unfortunately, the resulting formulation
$\min_{\bm{z}, \bm{s}, \bm{w} \in \ma{Z}}-\frac{\bm{\mu}^{\top}\bm{\beta}^*(\bm{z})}{2}$
is neither convex nor differentiable in $\bm{z}$, when $\bm{\beta}^*(\bm{z}) = (\bm{Z}(\bm{M}+\lambda_\beta \bm{I})\bm{Z})^\dagger \bm{Z}\bm{\mu}$, making the problem intractable. However, observe that we only care about $\bm{\beta}^*(\bm{z})$ for binary vectors $\bm{z}$. Therefore, we proceed to consider convex relaxations of $(\bm{Z}(\bm{M}+\lambda_\beta \bm{I})\bm{Z})^\dagger \bm{Z}$ such that it agrees with $(\bm{Z}(\bm{M}+\lambda_\beta \bm{I})\bm{Z})^\dagger \bm{Z}$ on all binary points $\bm{z}$. Specifically, we prove the following proposition, which allows us to convexify the expression above:
\begin{prop}
\label{prop:convexify}
Let $\bm{M}$ be a positive semi-definite matrix. Then, we have, for $\bm{z} \in \{0,1\}^{TD}$, $\bm{Z}=\diag(\bm{z}^1,\cdots,\bm{z}^T)$, and $\lambda_\beta >0$:
\begin{equation}
(\bm{Z}(\bm{M}+\lambda_\beta \bm{I})\bm{Z})^\dagger\bm{Z} = (\lambda_\beta \bm{I}+  \bm{Z}\bm{M})^{-1}\bm{Z}.
\end{equation}
\end{prop} The proof is included in Appendix \ref{sec:appendix-prop:convexify}. Finally, we prove that, using the reformulation in Proposition \ref{prop:convexify}, the problem becomes convex in $\bm{z}$:
\begin{theo}
\label{theo:convexity}
Let $\bm{M},\bm{\mu}$ be defined in Lemma \ref{lem:reform}, and $\lambda_\beta>0$. Then, the optimization problem in (\ref{eq:obj_mio})--(\ref{eq:c6_mio}) is equivalent to the following binary convex optimization problem: 
\begin{equation} \label{eq:bin_conv_reformulation}
    \min_{\bm{z}, \bm{s}, \bm{w} \in \mathcal{Z}}-\frac{\bm{\mu}^{\top}\bm{\beta}^*(\bm{z})}{2},
\end{equation}
where $\bm{\beta}^*(\bm{z})=(\lambda_\beta \bm{I}+  \bm{Z}\bm{M})^{-1}\bm{Z}\bm{\mu}$.
\end{theo} The proof is given in Appendix \ref{sec:appendix-theo:convexity}. Theorem \ref{theo:convexity} shows that the original problem as shown in \eqref{eq:obj_mio}-\eqref{eq:c6_mio} can be reformulated into a binary convex optimization problem over $\bm{z},\bm{s},\bm{w}$, which is amenable to a cutting plane-type algorithm. We point out that the key ingredient that enabled such convex relaxation, Proposition \ref{prop:convexify}, is by no means obvious: there are infinitely many relaxations that match exactly the binary points of $(\bm{Z}(\bm{M}+\lambda_\beta \bm{I})\bm{Z})^\dagger\bm{Z}$. In fact, an arguably more natural construction of a relaxation is the following equality (that can be easily shown using Lemma \ref{lem:psuedo_iden}):
\begin{equation}
    (\bm{Z}(\bm{M}+\lambda_\beta \bm{I})\bm{Z})^\dagger\bm{Z} = (\lambda_\beta \bm{I}+  \bm{Z}\bm{M}\bm{Z})^{-1}\bm{Z}. \label{eq:badrelax}
\end{equation}
However, such a relaxation, unlike the one shown in Proposition \ref{prop:convexify}, results in a \emph{non-convex} reformulation of the problem as stated in \eqref{eq:obj_mio}-\eqref{eq:c6_mio}, making it significantly more difficult to solve. 


\section{Discussion of the Relaxation} \label{sec:relaxation}
\begin{figure}
\centering
\begin{tikzpicture}[
  declare function={
    f1(\x)= (\x<=0) * (0)   +
     (\x>0) * (-1/(20*\x);
         f2(\x)= -\x/(1+19*\x);
             f3(\x)= -\x/(1+19*\x^2);
  }
]
    \begin{axis}[
        axis lines=center,
        xmax = 1.1,
        ymax = 0.05,
        ymin = -0.2,
        ylabel=$f(z)$,
        xlabel=$z$,
          yticklabel style={/pgf/number format/fixed},
        ]
        \addplot [domain=0:1,samples=250, ultra thick, black] {f1(x)}
            node [pos=0.997, below right] {$f_1(z)$};
        \addplot [domain=0:1,samples=250, ultra thick, blue ] { f2(x)}
            node [pos=0.4, above right] {$f_2(z)$};
        \addplot [domain=0:1,samples=250, ultra thick, red ] { f3(x)}
            node [pos=0.25, below left] {$f_3(z)$};
    \end{axis}
\end{tikzpicture}
\caption{Various Relaxations of the Pseudoinverse.}
\label{fig:relaxation}
\end{figure}
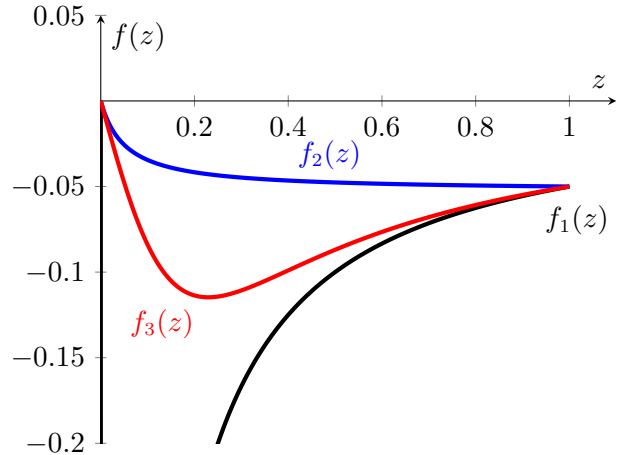
As shown in Section \ref{sec:binaryreformulation}, the key observation that enabled us to create the convex reformulation is Proposition \ref{prop:convexify}. In this section, we provide intuition on why the proposed relaxation works.

For simplicity, we consider the case where we have $T=D=1$, and a single binary variable $z$. Then, by Theorem \ref{theo:convexity}, the objective function for the optimization problem defined in (\ref{eq:obj_mio})--(\ref{eq:c6_mio}) has the form:
\[f_1(z)= -(z(m+\lambda_\beta)z)^\dagger \mu^2 z=\begin{cases}
          0, & z=0,\\
          -\frac{\mu^2}{z(m+\lambda_\beta)}, & z\neq 0.
\end{cases}\]
Proposition \ref{prop:convexify} then reads, for all $m>0$ and $z \in \{0,1\}$, $(z(m+\lambda_\beta)z)^\dagger z=\frac{z}{\lambda_\beta+mz}.$
After reformulation, the objective function has the form
$f_2(z)=-\frac{\mu^2z}{\lambda_\beta + mz}.$
While the other natural relaxation we can construct, as defined in Equation \eqref{eq:badrelax} gives the objective function $f_3(z)=-\frac{\mu^2z}{\lambda_\beta + mz^2}.$
In Figure \ref{fig:relaxation} we plot $f_1(z),f_2(z), f_3(z)$ for $m=19$, $\mu=1$, and $\lambda_\beta = 1$. 

First, we observe that in one dimension, the pseudoinverse is a discontinuous and non-convex function that follows a $-\frac{1}{z}$ type curve everywhere except for $z=0$, where it takes the value of 0. This clearly reflects the difficulty of solving the sparse problem as formulated in the standard way. We then observe that both $f_2(z)$ and $f_3(z)$ agree with $f_1(z)$ when $z\in \{0,1\}$, and therefore $f_2(z)$, $f_3(z)$ are both valid relaxations of the discontinuous function $f_1(z)$ on the binary values of $z$. However, we clearly see that $f_2(z)$ is a convex function in $z$, while $f_3(z)$ is not. This illustrates how the carefully chosen relaxation enables efficient convex algorithms to be utilized. 

We finally note that the relaxation utilized in this paper, $(\bm{Z}(\bm{M}+\lambda_\beta \bm{I})\bm{Z})^\dagger\bm{Z} = (\lambda_\beta \bm{I}+  \bm{Z}\bm{M})^{-1}\bm{Z},$
is not the only convex relaxation possible. For example, the following formula extends our relaxation into a family of relaxations for all $\mu\geq 0$:
\begin{equation*}
(\bm{Z}(\bm{M}+\lambda_\beta \bm{I})\bm{Z})^\dagger\bm{Z} = (\lambda_\beta \bm{I}+  \bm{Z}\bm{M})^{-1}\bm{Z} + \mu\left(\sum_{t=1}^{T} \sum_{d=1}^{D} \left(z_d^t-\frac{1}{2}\right)^2 - \frac{TD}{4}\right).
\end{equation*}
However, our numerical experiments in Appendix \ref{sec:test-different-relaxation} suggest that algorithm performance in both solution time and accuracy decays as $\mu$ increases, so our original relaxation is superior. We believe the development of more effective convex relaxations is a fruitful direction for future research.

\section{An Exact Cutting Plane Algorithm} \label{sec:cuttingplane}
In this section, we propose a cutting plane-type algorithm that solves Problem \eqref{eq:bin_conv_reformulation} to optimality. The proposed Algorithm \ref{alg:cutplane} is based on the outer approximation method by \cite{duran1986outer}, which iteratively tightens a piecewise linear lower approximation of the objective function. Algorithm \ref{alg:cutplane} provides pseudocode for the proposed approach.

\begin{algorithm*}
\caption{Cutting Plane Algorithm (\svc)}
\label{alg:cutplane}
\begin{algorithmic}
{\scriptsize
    \REQUIRE Data $(\*X^t,\*y^t)_{t=1}^T$, similarity graph $G$, sparsity parameters $(K_{\text{L}},K_{\text{G}},K_{\text{C}})$, regularization parameters $(\lambda_{\beta}, \lambda_{\delta}$).
    \ENSURE Learned coefficients $\bm{\beta}^{\star}$.
    
    \vspace{0.5em}
    
    \STATE \texttt{$\triangleright$ Find warm start using Algorithm \ref{alg:heuristic}:}
    
    \STATE $\bm{\beta}^{(0)} \leftarrow \texttt{find\_start}\left((\*X^t,\*y^t)_{t=1}^T, G, (K_{\text{L}},K_{\text{G}},K_{\text{C}}), (\lambda_{\beta}, \lambda_{\delta})\right)$
    
    \vspace{0.5em}
    
    \STATE \texttt{$\triangleright$ Compute corresponding binary variables:}
    
    \STATE $ (\bm{z}^{(0)},\bm{s}^{(0)}, \bm{w}^{(0)}) \leftarrow \texttt{find\_binaries}\left( \bm{\beta}^{(0)} \right)$
    
    \STATE $(i,\eta^{(0)}) \leftarrow (0,0)$
    
    \vspace{0.5em}
    
    \STATE \texttt{$\triangleright$ Cutting plane iterations:}
    
    \WHILE{$c(\bm{z}^{(i)}) > \eta^{(i)}$}
    
        \STATE $ (\bm{z},\bm{s}, \bm{w}, \eta)^{(i+1)} \ \leftarrow \ \underset{\tiny{\substack{\eta \in \mathbb{R}_{+}, \\ \bm{z},\bm{s}, \bm{w} \in \ma{Z}}}}{ \text{argmin} \ \eta} \ \text{s.t.}  \ \eta \geq c(\bm{z}^{(\tau)}) + \nabla_{\bm{z}} c(\bm{z}^{(\tau)})^{\top} (\bm{z} - \bm{z}^{(\tau)}), \ \forall \tau \in [i]$ 
        
        \STATE $ i \leftarrow i + 1$
    \ENDWHILE
    
    \vspace{0.5em}
    
    \STATE \texttt{$\triangleright$ Estimate coefficients using Theorem \ref{theo:convexity}:}
    
    \STATE $\bm{\beta}^{\star} \leftarrow \bm{\beta}^*(\bm{z}^{(i)})$
    
    \vspace{-2em}
    
    \RETURN $\bm{\beta}^{\star}$
}    
\end{algorithmic}
\end{algorithm*}

Recall from Theorem \ref{theo:convexity} that the objective function $c(\bm{z},\bm{\beta})$ is indeed convex in $\bm{z}$ and can, in fact, be written as function only of the binary variables $\bm{z}$ by solving the inner problem to optimality, i.e., 
\begin{equation} \label{eq:cutplane-cost}
    \min_{\bm{\beta}} c(\bm{z},\bm{\beta})
    : = c(\bm{z}) 
    = -\frac{\bm{\mu}^{\top}\bm{\beta}^*(\bm{z})}{2}
    = -\frac{1}{2}\bm{\mu}^{\top}(\lambda_\beta \bm{I}+  \bm{Z}\bm{M})^{-1}\bm{Z}\bm{\mu}.
\end{equation}
Algorithm \ref{alg:cutplane} also requires the computation of the gradient of the cost function $\nabla_{\bm{z}} c(\bm{z})$ at every binary point $\bm{z}$ it visits. 
We thus aim to differentiate the loss function $c(\bm{z})$ with respect to the diagonal entries of the matrix $\bm{Z} = \text{Diag}(\bm{z}^1,\dots,\bm{z}^T)$.
The partial derivative  with respect to component $z^t_d$ can be computed numerically using finite differences as $\frac{\partial c(\bm{z})}{\partial z^t_d} = \frac{c(\bm{z})-c(\bm{z}-\varepsilon \bm{e^{t}_{d}} )}{\varepsilon},$ where $\bm{e^{t}_{d}}$ denotes the basis vector with $1$ in position $(t,d)$ and $0$'s elsewhere and $\varepsilon$ is a sufficiently small constant. Such an approach would be highly impractical, as it would require $TD$ evaluations of the cost function \eqref{eq:cutplane-cost}. Instead, we utilize the chain rule to compute the gradient in closed form, as shown below:
\begin{lem} \label{lem:gradient}
Let $\bm{K} = \bm{K}(\bm{z}) := (\lambda_\beta \bm{I} +  \bm{Z}\bm{M})$ and let $\bm{E^{t}_{d}}$ denote a $TD \times TD$ matrix, with $1$ at position $(t,d),(t,d)$ and $0$'s elsewhere. Then, we have:
$\frac{\partial c(\bm z)}{\partial z^t_d}=\frac{1}{2}\bm{\mu}^{\top} \bm{K}^{-1} \left( \bm{E^{t}_{d}} \bm M \bm K^{-1} \bm{Z} \ - \ \bm{E^{t}_{d}} \right)\bm{\mu}.$
\end{lem} The proof is given in Appendix \ref{sec:appendix-lem:gradient}. We next discuss the computational complexity of the cut generation for Algorithm \ref{alg:cutplane}. The cut generation process requires the evaluation of the cost function $c(\bm{z})$ and its gradient $\nabla_{\bm{z}} c(\bm{z})$. Lemma \ref{lem:cut_generation} enables us to generate cuts more efficiently than we would with a naive implementation, which would require $O(T^3D^3)$ operations from inverting matrix $\bm{K}$. 
\begin{revi}
\begin{lem} \label{lem:cut_generation}
Let $\bm{z}$ be a feasible binary vector for Problem  \eqref{eq:bin_conv_reformulation}. Then, the cost function $c(\bm{z})$ and its gradient $\nabla_{\bm{z}} c(\bm{z})$ can be evaluated in $O\left( T^3 K_{\text{L}}^2 + T^2 K_{\text{L}}^3 + T^2 K_{\text{L}}D \right)$ operations.
\end{lem}
The proof, given in Appendix \ref{sec:appendix-lem:cut_generation}, relies on exploiting the sparsity and block tri-diagonal structure of the various matrices involved (e.g., $\bm{E}^t_d$, $\bm{Z}$) to reduce the need for inverting large matrices, and provides guidance on efficiently implementing Algorithm \ref{alg:cutplane}.
\end{revi}

Finally, Theorem \ref{theo:convergence} asserts that Algorithm \ref{alg:cutplane} converges to the optimal value of Problem \eqref{eq:bin_conv_reformulation} within a finite number of iterations. Intuitively, finite termination is guaranteed since the feasible set is finite and the outer-approximation process of Algorithm \ref{alg:cutplane} never visits a point twice. We also remark that we need not solve a new binary optimization problem at each iteration of Algorithm \ref{alg:cutplane} by integrating the entire algorithm within a single branch-and-bound tree, as proposed by \cite{quesada1992lp}, using lazy constraint callbacks.  
\begin{theo}
\label{theo:convergence}
Algorithm \ref{alg:cutplane} terminates and returns an optimal solution to Problem \eqref{eq:bin_conv_reformulation} in a finite number of iterations. 
\end{theo}
Noting that, from Theorem \ref{theo:convexity}, $f(z) = -\frac{\bm{\mu}^{\top}\bm{\beta}^*(\bm{z})}{2}$ is convex in $z$, where $\bm{\beta}^*(\bm{z})=(\lambda_\beta \bm{I}+  \bm{Z}\bm{M})^{-1}\bm{Z}\bm{\mu}$, and that zero is always a feasible solution, we can conclude on termination and convergence of the outer-approximation cutting plane algorithm (described in Algorithm \ref{alg:cutplane}) by application of the classic result from \cite{Fletcher1994SolvingMI}. 

\section{An Efficient Heuristic Algorithm} \label{sec:heuristic}

In this section, we develop a fast heuristic algorithm for solving the MIO formulation defined by Equations \eqref{eq:obj_orig}-\eqref{eq:sparsevary} to obtain good starting points for the cutting plane algorithm (Algorithm \eqref{alg:cutplane}) and assist in hyperparameter tuning.

\begin{revi}
The following lemma provides an upper bound for Problem \eqref{eq:obj_orig}-\eqref{eq:sparsevary}:
\begin{lem} \label{lem:upper-bound}
    Denote by $\mathcal{Z}_\beta$ the feasible set defined by Equations \eqref{eq:indsparse}-\eqref{eq:sparsevary}. Then, we have
    \begin{align}
    & \underset{\bm{\beta} \in \mathcal{Z}_\beta}{\min}
    \quad  \sum_{n=1}^N \sum_{t=1}^T \left(y_n^t- \sum_{d=1}^D X_{n,d}^t \beta_d^t\right)^2
    + \lambda_{\beta} \sum_{t=1}^T \sum_{d=1}^D \left( \beta_d^t\right)^2
    + \lambda_{\delta} \sum_{(s,t) \in E} \sum_{d=1}^D \left( \beta_d^t - \beta_d^s \right)^2 \nonumber \\
    & \leq
    \underset{\bm{\beta} \in \mathcal{Z}_\beta}{\min}
    \quad  
    \frac{1}{D} \sum_{n=1}^N \sum_{t=1}^T \sum_{d=1}^D \left(y_n^t- X_{n,d}^t \beta_d^t\right)^2
    + \lambda_{\beta} \sum_{t=1}^T \sum_{d=1}^D \left( \beta_d^t\right)^2
    + \lambda_{\delta} \sum_{t=1}^T \sum_{d=1}^D 2 d^t \left( \beta_d^t \right)^2.\label{eq:obj_heuristic}
\end{align}
\end{lem} 
The proof, which we relegate to Appendix~\ref{sec:appendix-lem:upper-bound}, is based on the observation that the prediction error of the best multivariate model is less than or equal to the error of any univariate model. The above manipulations enable us to obtain a (possibly loose) upper bound to the original optimization problem, which however is additively separable in the optimization variables. The interpretation of the new optimization problem is as follows: we now fit separate univariate regressions per vertex per feature; we approximate the slow variation penalty with a new regularization term that depends on the degree of each vertex; we keep all sparsity and slow variation constraints. 
\end{revi}

We then proceed similarly to Section \ref{sec:mioformulation}. We introduce binary variables $\bm{z},\bm{s}, \bm{w}$ to capture the local sparsity, global sparsity, and sparsely varying support requirements. Importantly, we now require that the local sparsity requirement is \emph{exactly} enforced, that is, $\sum_{d=1}^D z_d^t = K_\text{L},\ \forall t \in [T]$; we denote by $\ma{Z}_=$ the corresponding binary feasible set. We replace every occurrence of $\beta_d^t$ with $z_d^t \beta_d^t$. The resulting MIO formulation can then be written as:
\begin{align}
      & \underset{\bm{z},\bm{s}, \bm{w} \in \ma{Z}_=}{\min}\;\;\underset{\bm{\beta}}{\min} \quad
    \sum_{t=1}^T \sum_{d=1}^D
        \frac{1}{D} \sum_{n=1}^N \left(y_n^t- X_{n,d}^t z^t_d \beta_d^t\right)^2
        + \lambda_{\beta} (z^t_d\beta_d^t)^2
        + \lambda_{\delta} 2 d^t (z^t_d \beta_d^t )^2 \nonumber\\
    & =  \underset{\bm{z},\bm{s}, \bm{w} \in \ma{Z}_=}{\min}\;\;\underset{\bm{\beta}}{\min} \quad
    \sum_{t=1}^T \sum_{d=1}^D \left\{ \left[
        \frac{1}{D} \sum_{n=1}^N \left(y_n^t- X_{n,d}^t \beta_d^t\right)^2
        + \lambda_{\beta} (\beta_d^t)^2
        + \lambda_{\delta} 2 d^t ( \beta_d^t )^2
    \right] z^t_d + \frac{1}{D} \sum_{n=1}^N (y^t_n)^2(1-z^t_d)\right\}\nonumber\\
    & = \underbrace{\frac{1}{D}\sum_{t=1}^T\sum_{n=1}^N (y_n^t)^2(D-K_L)}_{:=L_0} + \underset{\bm{z},\bm{s}, \bm{w} \in \ma{Z}_=}{\min}\;\; \sum_{t=1}^T \sum_{d=1}^D \underbrace{\underset{\beta_d^t}{\min}  \left[ 
        \frac{1}{D} \sum_{n=1}^N \left(y_n^t- X_{n,d}^t \beta_d^t\right)^2
        + \lambda_{\beta} (\beta_d^t)^2
        + \lambda_{\delta} 2 d^t ( \beta_d^t )^2
    \right]}_{:=L^t_d} z^t_d \nonumber \\
   & := L_0 + 
    \underset{\bm{z},\bm{s}, \bm{w} \in \ma{Z}_=}{\min} \sum_{t=1}^T \sum_{d=1}^D L_d^t z^t_d.   \label{eq:heuristic-closed-form}
\end{align}

\begin{algorithm*}
\caption{Heuristic Algorithm (\svh)}
\label{alg:heuristic}
\begin{algorithmic}
{\scriptsize
    \REQUIRE Data $(\*X^t,\*y^t)_{t=1}^T$, similarity graph $G$, sparsity parameters $(K_{\text{L}},K_{\text{G}},K_{\text{C}})$, regularization parameters $(\lambda_{\beta}, \lambda_{\delta}$).
    \ENSURE Learned coefficients $\tilde{\bm{\beta}}$.
    
    \vspace{0.5em}
    
    \STATE \texttt{$\triangleright$ Compute loss for each vertex-feature pair:}
    
    \FOR{$t\in[T], \ d\in[D]$} 
    
        \STATE $L_d^t \leftarrow \underset{\beta_d^t}{\min} 
            \frac{1}{D} \sum_{n=1}^N \left(y_n^t- X_{n,d}^t \beta_d^t\right)^2
            + \lambda_{\beta} (\beta_d^t)^2
            + \lambda_{\delta} 2 d^t ( \beta_d^t )^2 $
    
    \ENDFOR
    
    \vspace{0.5em}
    
    \STATE \texttt{$\triangleright$ Solve linear relaxation of Problem \eqref{eq:heuristic-closed-form}:}
    
    \STATE $ (\tilde{\bm{z}}, \tilde{\bm{s}}, \tilde{\bm{w}}) \leftarrow \underset{\bm{z},\bm{s}, \bm{w} \in \tilde{\ma{Z}}_=}{\min} L_d^t z^t_d$
    
    \vspace{0.5em}
    
    \STATE \texttt{$\triangleright$ Ensure integrality:}
    
    \IF{$\tilde{\bm{z}} \not\in \{0,1\}^{TD}$}
    
        \STATE $\mathcal{I} \leftarrow \{ (t,d) \in [T]\times[D]:\ 0 < (\tilde{\bm z})^{t}_d < 1 \}$ \hfill \texttt{$\triangleright$ Find non-integral entries in $\tilde{\bm{z}}$.} \hspace{2em}
        
        \STATE $(\tilde{\bm z})^{t}_d \leftarrow 1,\ \forall (t,d) \in \mathcal{I}$ 
    
        \STATE $(\tilde{\bm{s}}, \tilde{\bm{w}}) \leftarrow f(\tilde{\bm{z}})$ \hfill \texttt{$\triangleright$ Accordingly update $\tilde{\bm{s}}, \tilde{\bm{w}}$ (as per Section \ref{sec:mioformulation}).} \hspace{2em} 
    
    \ENDIF
    
    \vspace{0.5em}
    
    \STATE \texttt{$\triangleright$ Ensure feasibility:}
    
    \WHILE{$(\tilde{\bm{z}}, \tilde{\bm{s}}, \tilde{\bm{w}}) \not\in \ma{Z}_=$}
    
        \STATE $\mathcal{S} \leftarrow \{ d \in [D]:\ \sum_{t=1}^T (\tilde{\bm z})^{t}_d > 0 \}$ \hfill \texttt{$\triangleright$ Find global support.} \hspace{2em}
    
        \STATE $d_0 \leftarrow \argmax_{d \in \mathcal S} \frac{1}{T}\sum_{t=1}^T L_d^t$  \hfill \texttt{$\triangleright$ Find feature $d \in \mathcal{S}$ with largest average loss across all vertices.} \hspace{2em}
        
        \STATE $(\tilde{\bm z})^{t}_d \leftarrow 0,\ \forall t \in [T]$
    
        \STATE $(\tilde{\bm{s}}, \tilde{\bm{w}}) \leftarrow f(\tilde{\bm{z}})$ \hfill \texttt{$\triangleright$ Accordingly update $\tilde{\bm{s}}, \tilde{\bm{w}}$ (as per Section \ref{sec:mioformulation}).} \hspace{2em} 
        
    \ENDWHILE
    
    \vspace{0.5em}
    
    \STATE \texttt{$\triangleright$ Estimate coefficients using Theorem \ref{theo:convexity}:}
    
    \STATE $\tilde{\bm{\beta}} \leftarrow \bm{\beta}^*(\tilde{\bm{z}})$
    
    \vspace{-2em}
    
    \RETURN $\bm{\beta}^{\star}$
}    
\end{algorithmic}
\end{algorithm*}
As shown in Equation \eqref{eq:heuristic-closed-form}, due to separability, we can solve the inner problem in closed form and obtain an integer linear optimization problem in the binary variables. The polyhedral feasible set of the corresponding linear relaxation, which we denote by $\tilde{\ma{Z}}_=$, can unfortunately be shown to not be integral. Nevertheless, we empirically show in Appendix \ref{sec:integrality-test-heuristic-algorithm} that the solution to the linear relaxation is, in fact, integral or near-integral, for a variety of realistic, non-pathological problems. Thus, we proceed by solving the linear relaxation of Problem \eqref{eq:heuristic-closed-form}. In case the solution $\tilde{\bm z}$ to the linear relaxation is not binary feasible, we round up all non-integral entries. Finally, to ensure feasibility in terms of the local sparsity, global sparsity, and sparsely varying constraints, we iteratively remove features from the global support until the resulting solution is indeed feasible. The overall Algorithm \ref{alg:heuristic} gives a feasible solution for Problem \eqref{eq:obj_orig}-\eqref{eq:sparsevary} in polynomial time (proof in Appendix \ref{sec:appendix-prop:heuristic}):
\begin{prop} \label{prop:heuristic}
Algorithm \ref{alg:heuristic} terminates and provides a feasible solution to Problem \eqref{eq:obj_orig}-\eqref{eq:sparsevary} in time $\tilde{O}\left(NTD  + (TD)^{2+\nicefrac{1}{6}} + T^2 K_{\text{L}}^2 (T + K_{\text{L}})\right)$. 
\end{prop} 

{\rev
\section{A Practical Hyperparameter Tuning Procedure} \label{sec:tuning}

The SSVR formulation involves five model hyperparameters, $\boldsymbol{\lambda} = (\lambda_\beta, \lambda_\delta,K_{\text{L}}, K_{\text{G}}, K_{\text{C}})$, which can be challenging to tune via a naive grid search procedure. In this section, we develop a practical hyperparameter tuning procedure relying on binary search and inspired by \cite{kenney2021mip}. 

For a given hyperparameter combination $\boldsymbol{\lambda}$ and training data $\bm{X}, \bm{y}$, we denote by $\bm{\beta}^*(\boldsymbol{\lambda})$ the learned coefficients from Algorithm \ref{alg:cutplane} and by $\tilde{\bm{\beta}}(\boldsymbol{\lambda})$ the learned coefficients from Algorithm \ref{alg:heuristic}. Then, given validation data $\bm{X_V},\bm{y_V}$, we can define the following validation cost functions for the exact and heuristic algorithms utilizing, respectively, the expressions in Equations \eqref{eq:obj_orig} and \eqref{eq:obj_heuristic}:
{\small \begin{align*}
    c_V^*(\bm{K}; \lambda_\beta, \lambda_\delta)
    & := \sum_{t=1}^T \left\|[\bm{y_V}]^t-[\bm{X_V}]^t[\bm{\beta}^*(\boldsymbol{\lambda})]^t\right\|_2^2
        + \lambda_{\beta} \sum_{t=1}^T \| [\bm{\beta}^*(\boldsymbol{\lambda})]^t \|_2^2 
        + \lambda_{\delta} \sum_{(s,t) \in E} \| [\bm{\beta}^*(\boldsymbol{\lambda})]^t - [\bm{\beta}^*(\boldsymbol{\lambda})]^s \|_2^2,\\
    \tilde{c}_V(\bm{K}; \lambda_\beta, \lambda_\delta)
    & := \frac{1}{D} \sum_{n=1}^N \sum_{t=1}^T \sum_{d=1}^D \left([y_V]_n^t- [X_V]_{n,d}^t [\tilde{{\beta}}(\boldsymbol{\lambda})]_d^t\right)^2
    + \lambda_{\beta} \sum_{t=1}^T \sum_{d=1}^D \left( [\tilde{\beta}(\boldsymbol{\lambda})]_d^t\right)^2
    + \lambda_{\delta} \sum_{t=1}^T \sum_{d=1}^D 2 d^t \left( [\tilde{\beta}(\boldsymbol{\lambda})]_d^t \right)^2,
\end{align*}}
where $\bm{K}=(K_L, K_G, K_C)$ are the sparsity parameters. We aim to show that this function as a function of $\bm{K}$ is ``elbow-shaped,'' as defined below: 
\begin{definition}
    A function $c(K)$ is elbow-shaped if there exists $K^*$ and $\delta>0$ such that:
     \begin{itemize}
         \item For all $K_2>K_1>K^*$, we have that $0\leq \frac{c(K_1)-c(K_2)}{c(K_1)(K_2-K_1)}\leq \delta$.
         \item For all $K_2>K^*>K_1$, we have that $\frac{c(K_1)-c(K_2)}{c(K_1)(K_2-K_1)}>\delta$.
     \end{itemize}
\end{definition}
An elbow-shaped function decreases quickly before reaching the optimal value $K^*$ and then stays roughly flat. One property of elbow-shaped functions is that a modified bisection algorithm, as stated in Algorithm \ref{alg:bisection_search}, can find the optimum quickly:
\begin{lem}[\cite{kenney2021mip}]\label{lem:bisection_search}
     Assume a function $c(K)$ is elbow-shaped, then, Algorithm \ref{alg:bisection_search} with improvement tolerance $\delta$ recovers $K^*$. 
\end{lem}
\begin{algorithm*}
\caption{Bisection Algorithm (\bisection($c(\cdot), K_l^0, K_u^0, \delta$))}
\label{alg:bisection_search}
\begin{algorithmic}
\begin{scriptsize}
    \REQUIRE Validation cost function $c(K)$, initial lower bound $K_l^0$, initial upper bound $K_u^0$, improvement tolerance $\delta$.
    \ENSURE Optimal sparsity parameters $K^*$.
    
    \STATE $K_{l},K_{u} \gets K_l^0, K_u^0$
    \STATE $K_{u} \gets D$
    \STATE $K_{m} \gets \lfloor \frac{K_{u} + K_{l}}{2} \rfloor$
    \WHILE{$K_{u} - K_{l} > 1$}
        \STATE $c_m, c_l, c_h \gets c(K_{m}),c(K_{l}),c(K_{u})$
         \STATE \texttt{$\triangleright$ If the \emph{percentage} cost improvement for every unit of parameter is larger than $\delta$:}
        \IF{$\frac{c_m - c_h}{c_m \cdot (K_{u} - K_{m})} > \delta$ \& $\frac{c_l - c_m}{c_l \cdot (K_{m} - K_{l})} > -\delta$}
            \STATE $K_{l} \gets K_{m}$
        \ELSE
            \STATE $K_{u} \gets K_{m}$
        \ENDIF
    \ENDWHILE
    \STATE $c_l, c_h \gets c(K_{l}), c(K_{u})$
    \IF{$\frac{c_h - c_l}{c_l \cdot (K_{u} - K_{l})} > \delta$}
        \STATE $K^* \gets K_{u}$
    \ELSE
        \STATE $K^* \gets K_{l}$
    \ENDIF
    \vspace{-2em}
    \RETURN $K^*$
\end{scriptsize}
\end{algorithmic}
\end{algorithm*}
The following proposition guarantees that, under a setting where the model is correctly specified, the cost functions $c^*$ and $\tilde{c}$ do admit this elbow shape as the number of samples grows large:
\begin{prop}\label{prop:elbow_shape}
    Assume that, for all $t=1,\cdots, T$, the following hold:
    \begin{itemize}
    \item The rows of $\bm{X}^t$, $\bm{x}_n^t \in \mathbb{R}^D, n\in[N]$, are i.i.d. with distribution $P_t$ with finite second moments. 
    \item For some $\bm{{\beta^0}}^{t} \in \md{R}^D$, $\bm{y}^t$ satisfies ${y}_n^t=\bm{x}_n^t\bm{{\beta^0}}^{t}+\epsilon_n^t$ with $\E[\epsilon_n^t]=0$ and $\epsilon_n^t$ has finite second moments. 
\end{itemize}
Define the true sparsity parameters as: 
$K_L^* = \max_{t\in[T]} \|\bm{{\beta^0}}^{t}\|_0,
\quad K_G^* = \left| \bigcup_{t\in [T]} \text{Supp}(\bm{{\beta^0}}^{t}) \right|,
\quad K_C^* = \max_{(s,t) \in E}  \left| \text{Supp}(\bm{{\beta^0}}^{t}) \triangle \text{Supp}(\bm{{\beta^0}}^{s}) \right|$.
Then, for every combination of $(\lambda_\beta, \lambda_\delta)$ values, as the number of training and validation samples $N\to \infty$ and $N_V \to \infty$, $c^*_V$ and $\tilde{c}_V$ exhibit the following behavior:
\begin{align*}
 c_V^*(\bm{K}; \lambda_\beta, \lambda_\delta)- \sum_{t=1}^T N_V\E[(\epsilon_n^t)^2] &\rightarrow \begin{cases}
   \infty, & \text{if } K_L< K_L^* \text{ or } K_G< K_G^* \text{ or } K_C< K_C^*,\\
   \displaystyle C,  & \text{otherwise,}
\end{cases}\\
\tilde{c}_V(\bm{K}; \lambda_\beta, \lambda_\delta)- \sum_{t=1}^T N_V\E[(\epsilon_n^t)^2] &\rightarrow \begin{cases}
   \infty, & \text{if } K_L< K_L^* \text{ or } K_G< K_G^* \text{ or } K_C< K_C^*,\\
   \tilde{C},  & \text{otherwise,}
\end{cases}
\end{align*}
where $C,\tilde{C}<\infty$. In particular, it tends to an elbow-shaped function in each of its arguments $K_L, K_G, K_C$ when fixing the remaining arguments above its true values ($K_L^*, K_G^*, K_C^*$ respectively). 
\end{prop} %
The proof is contained in Appendix \ref{sec:appendix-prop:elbow_shape}. Proposition \ref{prop:elbow_shape} states that, as $N, N_V \to \infty$, both $c^*_V$ and $\tilde{c_V}$ satisfy the elbow-shaped condition for $K_L, K_G, K_C$ \emph{individually} at the true parameters $K_L^*, K_G^*, K_C^*$ provided that the other sparsity parameters are at or above their true value. Thus, Lemma \ref{lem:bisection_search} implies that the bisection routine can be used \textit{sequentially} to discover the optimal parameters. Using this fact, we can construct the full algorithm for tuning hyperparameters as presented in Algorithm \ref{alg:hyperparameter}. Specifically, we search $\lambda_\beta$ and $\lambda_\delta$ over a grid; for each $(\lambda_\beta,\lambda_\delta)$, we discover the optimal sparsity parameters $K_G, K_L, K_C$ one at a time using the bisection routine and holding any undiscovered sparsity parameter at its maximum possible value (so that the elbow condition is satisfied). 

From a practical standpoint, practitioners do not need to pre-specify any hyperparameter combination for the sparsity parameters: the bisection procedure will efficiently search over the entire range of possible values. Using this hyperparameter tuning routine, we develop three versions of our algorithm for training SSVR models:
\begin{enumerate}
    \item We select the hyperparameters using the validation cost function $c_V^*$ associated with Algorithm \ref{alg:cutplane} and run Algorithm \ref{alg:cutplane} for the final model. We denote this algorithm \svc.
    \item We select the hyperparameters using the validation cost function $\tilde{c_V}$ associated with Algorithm \ref{alg:heuristic} and run Algorithm \ref{alg:cutplane} for the final model. We denote this algorithm \svhc.
    \item We select the hyperparameters using the validation cost function $\tilde{c_V}$ associated with Algorithm \ref{alg:heuristic} and run Algorithm \ref{alg:heuristic} for the final model. We denote this algorithm \svh.    
\end{enumerate}

\begin{algorithm*}
\caption{SSVR Hyperparameter Tuning Algorithm}
\label{alg:hyperparameter}
\begin{algorithmic}
\begin{scriptsize}
    \REQUIRE Validation cost function $c_V(\lambda_\beta, \lambda_\delta, K_L, K_G, K_C)$, Grid for regularization parameters $\mathcal{S}_\lambda$, improvement tolerance $\delta$.
    \ENSURE Optimal hyperparameters $\lambda_\beta^*, \lambda_\delta^*,K_L^*, K_G^*, K_C^*$.
    \FOR{$(\lambda_\beta, \lambda_\delta) \in \mathcal{S}_\lambda$}
        \STATE $K_G^* \gets \bisection(c_V(\lambda_\beta, \lambda_\delta, D, \cdot, D|E|),1, D, \delta)$
        \STATE $K_L^* \gets \bisection(c_V(\lambda_\beta, \lambda_\delta, \cdot, K_G^*, D|E|),1, D, \delta)$
        \STATE $K_C^* \gets \bisection(c_V(\lambda_\beta, \lambda_\delta, K_L^*,  K_G^*, \cdot),1, D|E|, \delta)$
        \STATE $c(\lambda_\beta, \lambda_\delta)<- c_V(\lambda_\beta, \lambda_\delta, K_L^*,  K_G^*,  K_C^*)$
    \ENDFOR
    
    $\lambda_\beta^*, \lambda_\delta^* \gets \argmin_{\lambda_\beta, \lambda_\delta} c(\lambda_\beta, \lambda_\delta)$
    \vspace{-2em}
    \RETURN $\lambda_\beta^*, \lambda_\delta^*, K_L^*,  K_G^*,  K_C^*$
\end{scriptsize}
\end{algorithmic}
\end{algorithm*}
}
\section{Experiments on Synthetic Datasets} \label{sec:synthetic}

In this section, we evaluate the proposed SSVR framework using synthetic data. For ease of exposition, we present a high-level description of our experimental methodology and a small set of aggregated and selected computational results; we defer the details to Appendix \ref{sec:appendix-expers}.


\begin{revi}

\paragraph{Data Generation Methodology.} For our synthetic data experiments, we generate a number of datasets as follows. We create a matrix of ground truth sparse and slowly varying regression coefficients $\bm \beta \in \mathbb{R}^{T\times D}$ according to the desired sparsity and slow variation parameters and over a (known) Erdos-Renyi similarity graph $G$. 
We then create a random data matrix $\bm X \in \mathbb{R}^{N\times T \times D}$ with Toeplitz correlation structure across features, and use the ground truth coefficients $\bm \beta$ to generate noisy responses $\bm Y \in \mathbb{R}^{N\times T}$. 
Our data generation methodology involves a number of problem parameters: the number of data points $(N)$, the number of features $(D)$, the number of vertices in the similarity graph $(T)$, the density of the similarity graph $d_G$, the level of variation of the regression coefficients between adjacent vertices $(\sigma_V)$, the sparsity parameters $(K_L, K_G, K_C)$, the correlation between features $(\rho_d)$, and the signal-to-noise ratio $(\xi)$ in the generated data. We generate different datasets by varying the above parameters. We provide more details on data generation in Appendix \ref{sec:appendix-expers-data-generation}.

\paragraph{Algorithms.} We implement the proposed algorithms (\svc, \svhc, and \svh\ as per Section \ref{sec:tuning}), the cutting plane algorithm of \cite{bertsimas2020sparse} that solves the standard sparse regression formulation shown in Problem \eqref{eq:sparse_reg} (referred to as \sr), and a suite of 4 variants of the sum-of-norms regularization framework of \cite{ohlsson2010segmentation, hallac2017snapvx} shown in Problem \eqref{eq:boyd} with and without lasso regularization (referred to as \rsnOne, \rsnOneLasso, \rsnTwo, \rsnTwoLasso; for synthetic experiments, we only report $\ell_1$ methods as they strictly outperform $\ell_2$ methods). For each method, we tune all hyperparameters ($\lambda_\beta$, $\lambda_\delta$, $K_\text{L}$, $K_\text{G}$, and $K_\text{C}$) using a standard holdout validation procedure; we use grid search over the same range of values for the $\lambda$'s (informed by the works of \cite{chu2015warm,wu2020optimal}) and bisection search for the $K$'s (as per Section \ref{sec:tuning}). We describe the implementation details of all methods, e.g., programming language and software used, and the computing environment in which we run our experiments in Appendix \ref{sec:appendix-expers-algorithms-software}. 

\paragraph{Evaluation Metrics.} We use various evaluation metrics to assess different aspects of the regression problem: out-of-sample R$^2$ statistic (Test R$^2$) to assess each method's predictive power; mean absolute error in the estimated coefficients (MAE) and number of differences in support expressed as a percentage of the total support size (DS) to assess each method's estimation accuracy; mean absolute change in coefficients across adjacent vertices relative to the ground truth coefficients (MAC) to assess each method's ability to capture the underlying slowly varying structure; computational time for hyperparameter tuning and refitting the final model (Time-Tune, Time-Refit) to assess each method's computational efficiency; optimality gap (Gap), number of cuts (Cut Count), and average cut time (ACT) to compare the performance of the proposed cutting plane method (Algorithm \ref{alg:cutplane}) against the cutting plane method of \cite{bertsimas2020sparse}. We outline the above in more detail in Table \ref{tab:metrics} in Appendix \ref{sec:appendix-expers-data-generation}.

\paragraph{Aggregated Results.} We first report aggregated results from our sensitivity analysis obtained by setting up a series of $26$ experiments, each of which corresponds to a fixed setting of the problem parameters $(N,T,D,K_{\text{L}},K_{\text{G}},K_{\text{C}}, \sigma_v, d_G, \rho_d, \xi)$. The details of the 26 parameter settings can be found in Appendix \ref{sec:appendix-expers}. For each problem parameter setting, we independently generate $10$ datasets (resulting in a total of 260 synthetic datasets) and, for each method, we compute the mean and standard deviation of each evaluation metric across those $10$ datasets. Then, we rank the methods according to their performance. In Table \ref{tab:synthetic-results}, we report the mean and standard deviation of the rank of each method across all 260 datasets. We provide detailed results in Appendix \ref{sec:appendix-expers-additional}.

{
\begin{table}[htbp]
\caption{Mean and standard deviation of each method's ranking across all 260 synthetic datasets. Note that Gap, Cut Count, and ACT only apply to cutting plane-based methods.}
\label{tab:synthetic-results}
\resizebox{\textwidth}{!}{
\begin{tabular}{lrrrrrrrrr}
\toprule
 \textbf{Algorithm} & \textbf{Test R$^2$} & \textbf{MAE} & \textbf{DS} & \textbf{MAC} & \textbf{Time Tune} & \textbf{Time Refit} & \textbf{Gap} & \textbf{Cut Count} & \textbf{ACT} \\ \midrule
\svc & \textbf{1.0 (0.0)} & 2.1 (0.89) & 3.19 (0.75) & \textbf{1.0 (0.0)} & 5.57 (1.16) & 2.19 (0.51) & \textbf{1.1 (0.3)} & \textbf{1.05 (0.22)} & 1.95 (0.22) \\
\svhc & 1.62 (0.5) & 2.52 (0.93) & 3.67 (0.8) & 1.05 (0.22) & \textbf{1.86 (0.36)} & 3.1 (0.62) & 2.0 (0.45) & 1.95 (0.22) & 3.0 (0.0) \\
\svh & 4.86 (0.91) & 4.67 (1.11) & 4.67 (0.86) & 1.33 (0.73) & 1.0 (0.0) & \textbf{1.0 (0.0)} & - & - & - \\
\sr & 5.71 (0.78) & 1.86 (1.2) & 2.05 (0.74) & 5.71 (0.96) & 3.29 (0.72) & 5.57 (1.25) & 2.9 (0.3) & 3.0 (0.0) & \textbf{1.05 (0.22)} \\
\rsnOne & 3.33 (0.86) & 5.62 (0.92) & 6.0 (0.45) & 4.14 (0.48) & 5.14 (0.91) & 5.38 (0.59) & - & - & - \\
\rsnOneLasso & 4.24 (0.89) & \textbf{1.76 (1.45)} & \textbf{1.43 (1.36)} & 4.29 (0.64) & 4.38 (0.97) & 4.43 (0.75) & - & - & - \\ \bottomrule
\end{tabular}
}
\end{table}
}
In general, our proposed methods (\svc, \svhc, \svh) demonstrate superior or comparable performance in multiple dimensions—ranging from accuracy and interpretability to computational efficiency. Under the distributional assumptions of our synthetic experiments, incorporating Lasso regularization, as in the \rsnOneLasso\ method, appears to offer particular advantages in capturing the ground truth sparsity pattern, hinting at the potential integration of similar techniques in future versions of our algorithms. More specifically:
\begin{itemize}
    \item Predictive Power: The Test R$^2$ statistics show that \svc\ and \svhc\ have the best out-of-sample predictive capabilities, with the former consistently showcasing the best predictive performance across all 26 problem parameter settings. The inferior performance of \sr\ and \rsnOne\ emphasize, respectively, the importance of capturing the underlying slowly varying structure and imposing sparsity (thereby avoiding overfitting).
    \item Estimation Accuracy: \rsnOneLasso\ performs best in terms of MAE and DS owing, in part, to the more exhaustive grid search-based hyperparameter tuning approach that we combine it with. We note that the variation in \rsnOneLasso\'s ranking is large and our methods tend to be very close in absolute performance. In terms of MAC, our proposed methods are best at capturing the right amount of variation in coefficients across adjacent vertices. We emphasize the existence of a tradeoff between the above metrics: increasing the slowly varying penalty results in better capturing the variation in coefficients but slightly deteriorates estimation accuracy (especially in low-noise regimes).
    \item Computational Efficiency: \svh\ stands out as the most efficient algorithm, solving problems with 10,000s of parameters in a few seconds. \svhc\ comes second, with an additional computational burden of 70 seconds on average (corresponding to the average time to refit the selected model); this is a crucial finding, particularly for practitioners who require quick and efficient solutions without compromising much on accuracy.
    \item Cutting Plane Method Performance: When considering the cutting plane-specific metrics like Gap and Cut Count, \svc\ and \svhc\ outperform \sr, which highlights the advances our method brings to the cutting plane algorithm techniques. \svc\ is always better at proving optimality and generates remarkably fewer cuts. Although \sr\ shows a slightly faster average cut time, it is at the expense of accuracy and optimality, as evidenced by its performance in other metrics. To identify the boundary of the proposed cutting plane algorithm, we consider larger problems with $N \in \{1,000, ..., 7,000\}$, $T \in \{1, ..., 40\}$, and $D \in \{1, ..., 900\}$, thereby having problem sizes of up to $10,000$ decision variables. Figure \ref{fig:scalability} (left column) shows that, by exploiting the problem structure, \svc\ generates vastly fewer cuts than \sr. Figure \ref{fig:scalability} (right column) suggests that the average cut generation time for \svc\ is insensitive to $N$, increases quadratically with $T$, and increases linearly with $D$, in agreement with Lemma \ref{lem:cut_generation}; for \sr, the increase is linear in the total number of data points $N'=NT$ (which increases with both $N$ and $T$) and linear in $D$, in agreement with \cite{bertsimas2020sparse}.

    \begin{figure}[htbp] 
    \begin{adjustbox}{minipage=\linewidth,scale=0.82}
        \centering
        \begin{subfigure}[b]{0.49\linewidth}
            \centering\includegraphics[width=\textwidth]{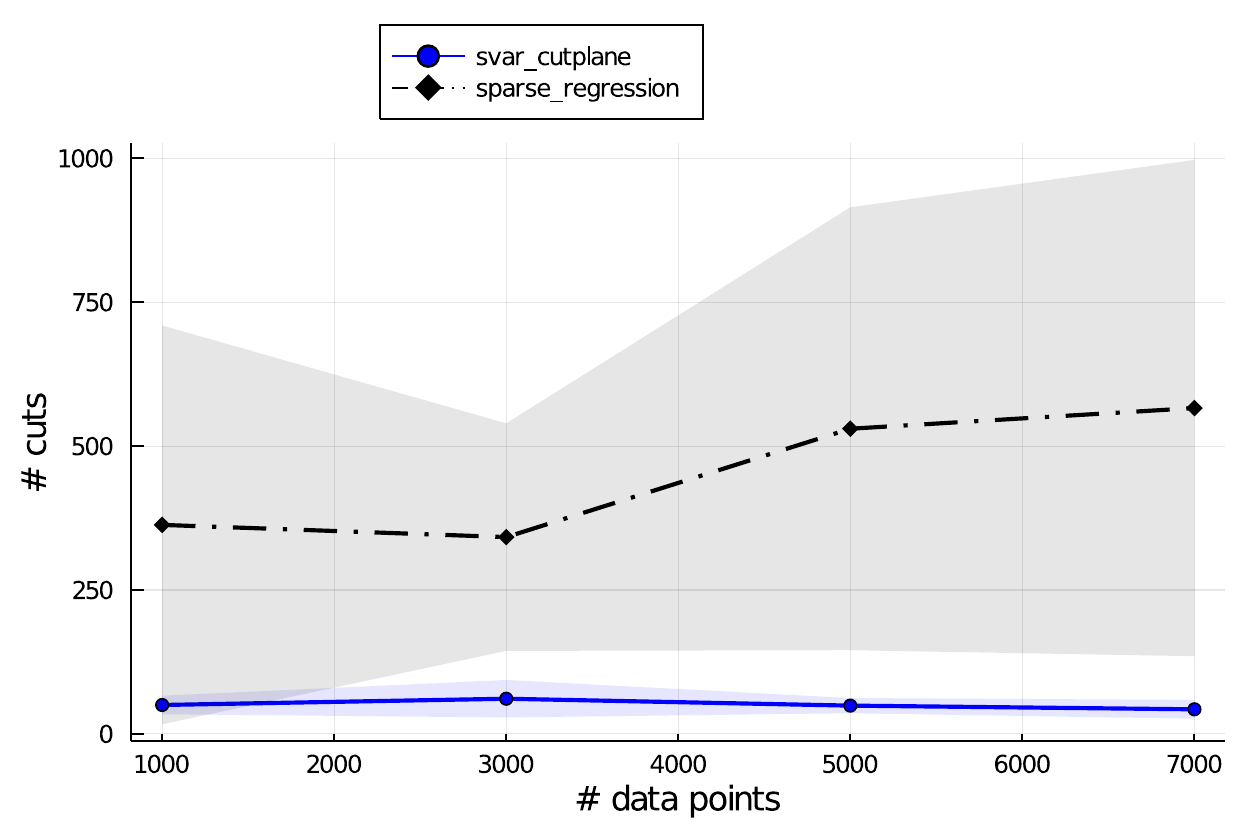}
            \caption{Number of cuts as function of N.\label{fig:synthetic_N_cut_count}}
        \end{subfigure}
        \hfill
        \begin{subfigure}[b]{0.49\linewidth}
            \centering\includegraphics[width=\textwidth]{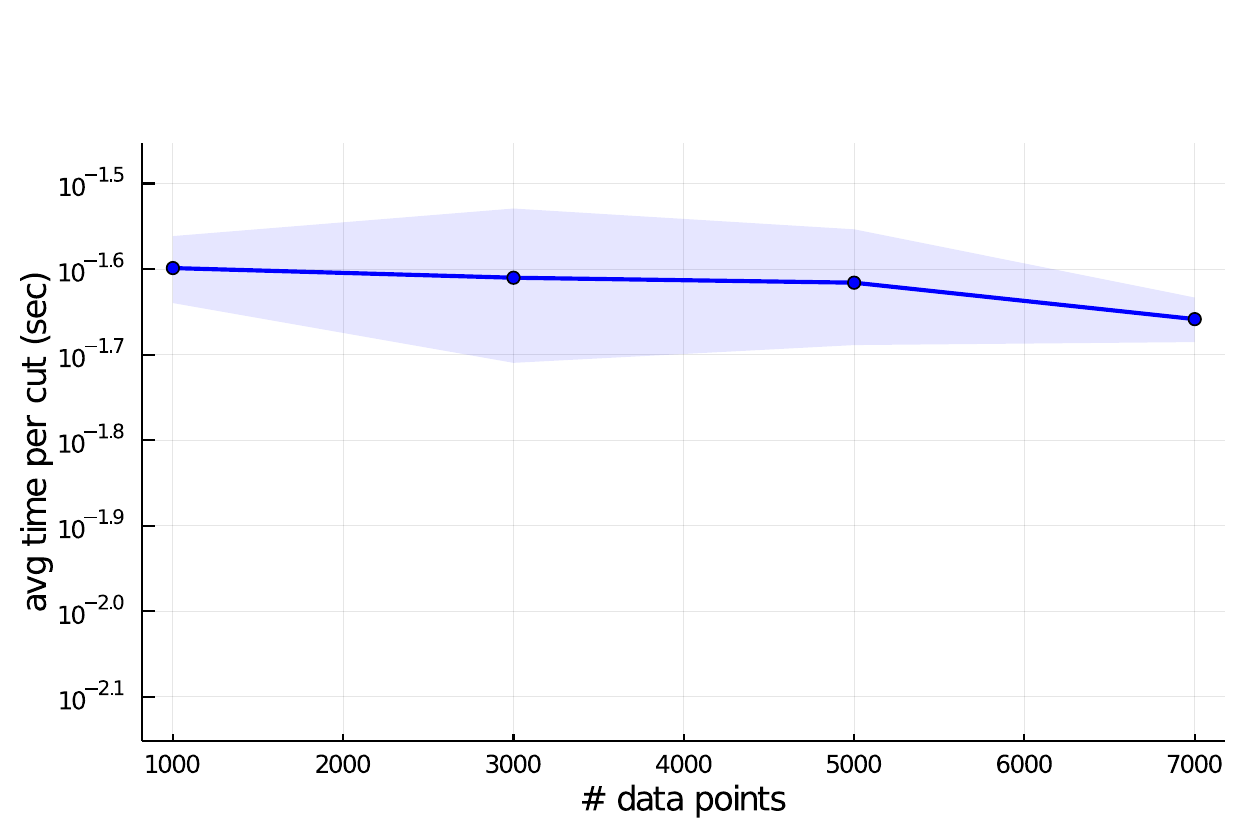}
            \caption{Average time per cut as function of N.\label{fig:synthetic_N_t_cut_avg}}
        \end{subfigure}%
        
        \begin{subfigure}[b]{0.49\linewidth}
            \centering\includegraphics[width=\textwidth]{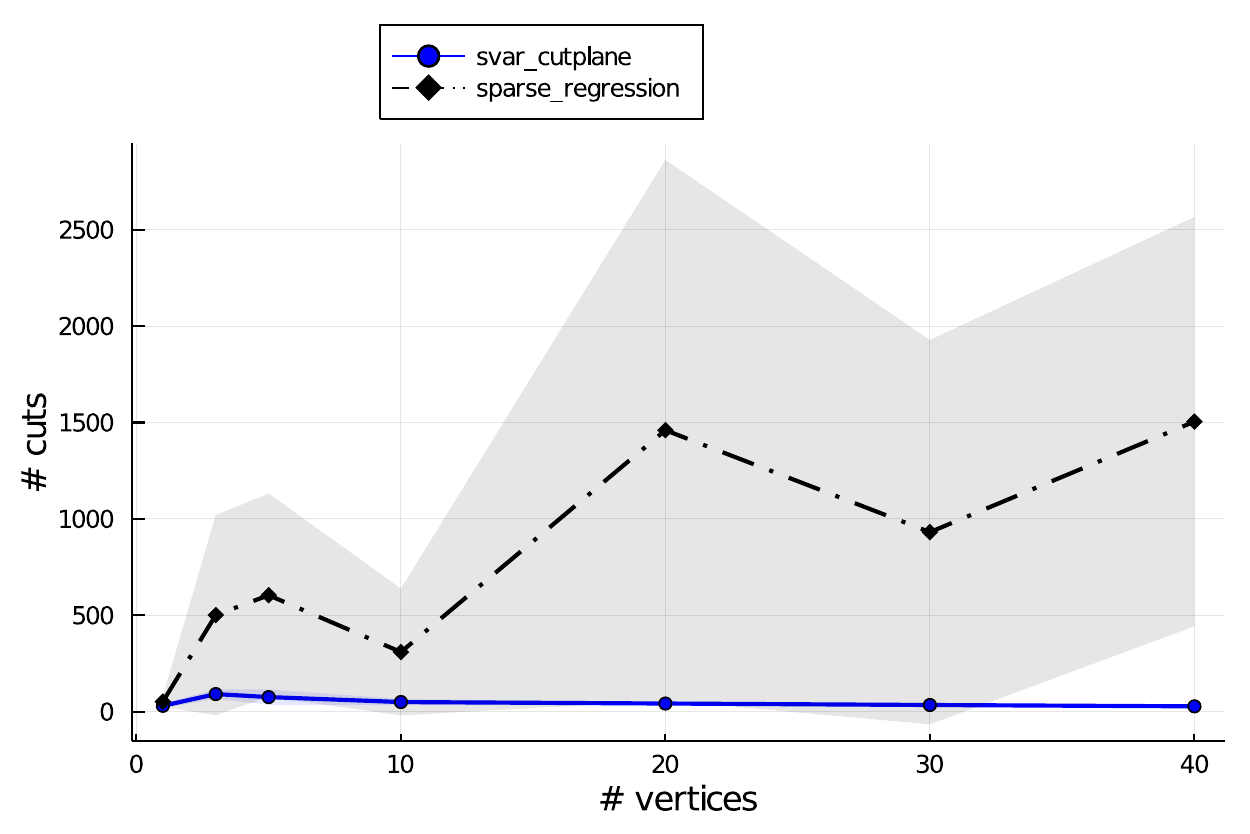}
            \caption{Number of cuts as function of T.\label{fig:synthetic_T_cut_count}}
        \end{subfigure}
        \hfill
        \begin{subfigure}[b]{0.49\linewidth}
            \centering\includegraphics[width=\textwidth]{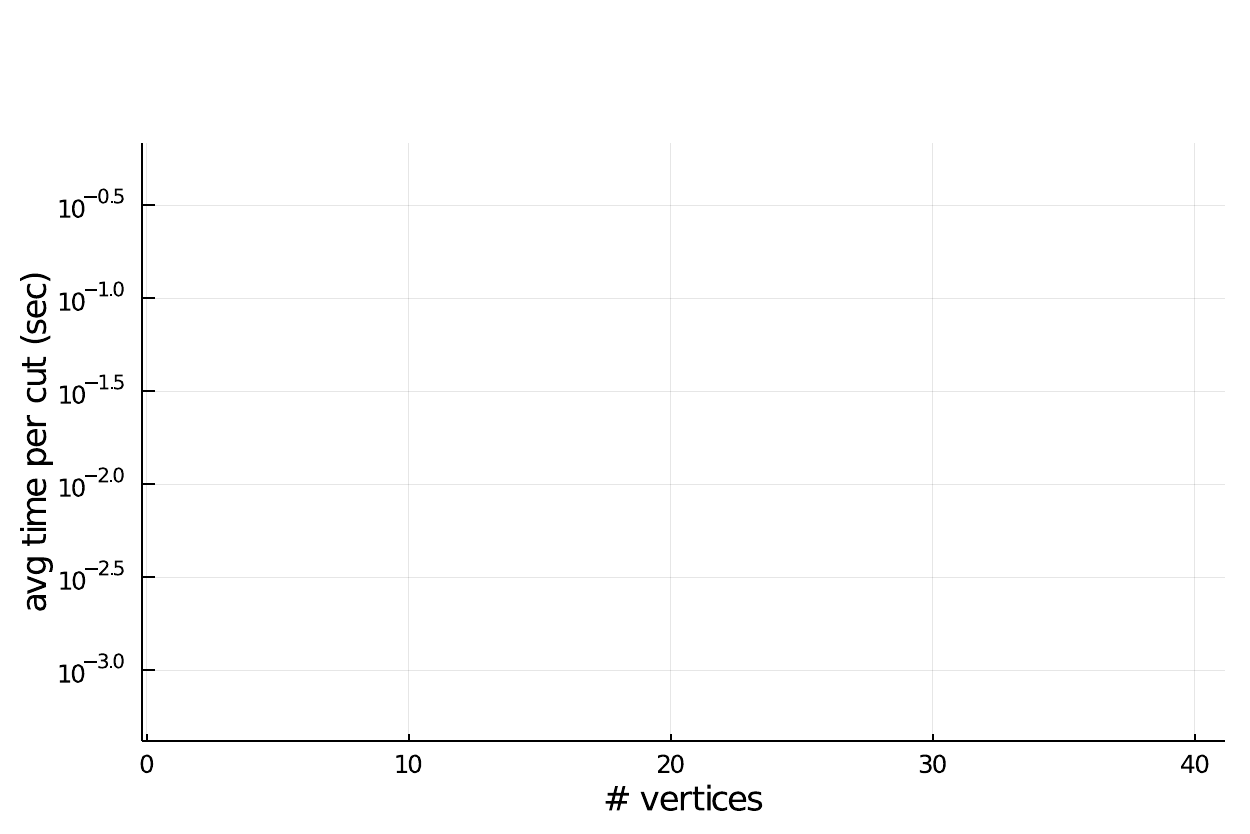}
            \caption{Average time per cut as function of T.\label{fig:synthetic_T_t_cut_avg}}
        \end{subfigure}%
        
        \begin{subfigure}[b]{0.49\linewidth}
            \centering\includegraphics[width=\textwidth]{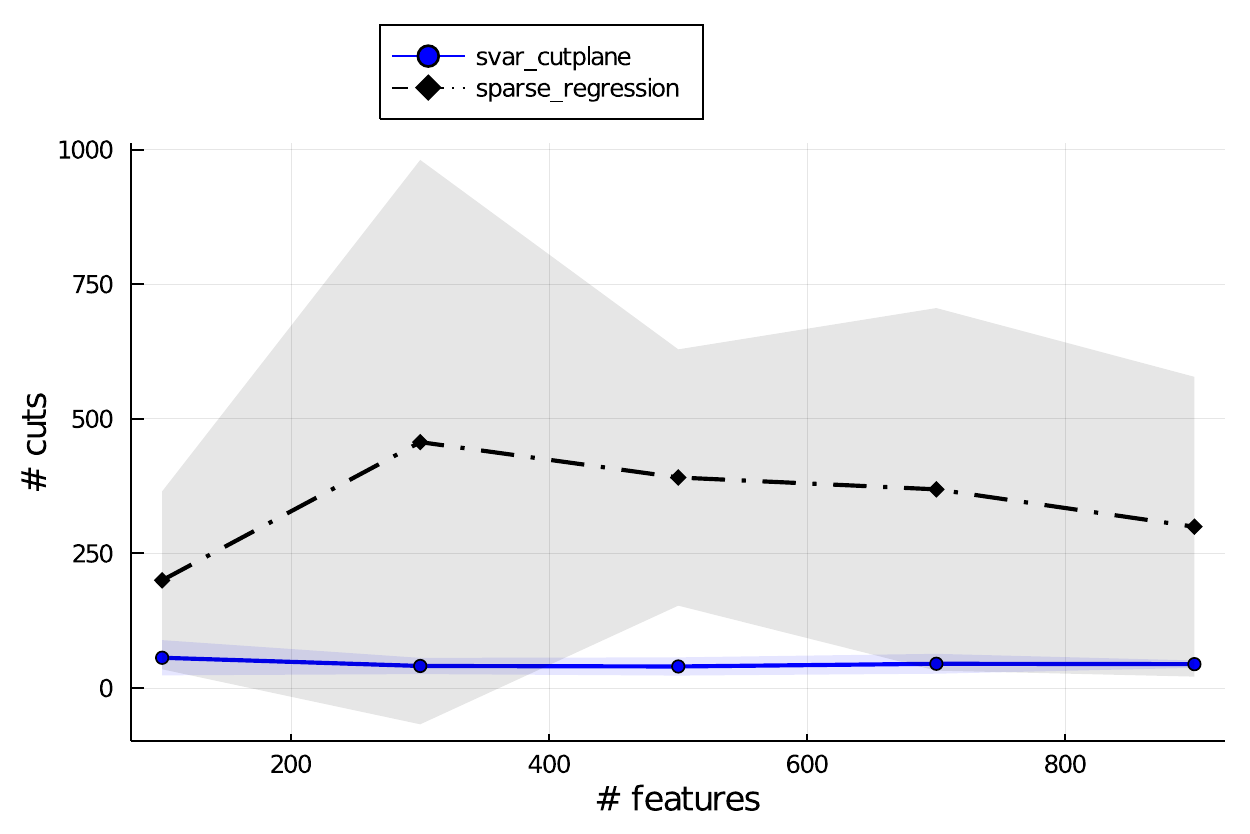}
            \caption{Number of cuts as function of D.\label{fig:synthetic_D_cut_count}}
        \end{subfigure}
        \hfill
        \begin{subfigure}[b]{0.49\linewidth}
            \centering\includegraphics[width=\textwidth]{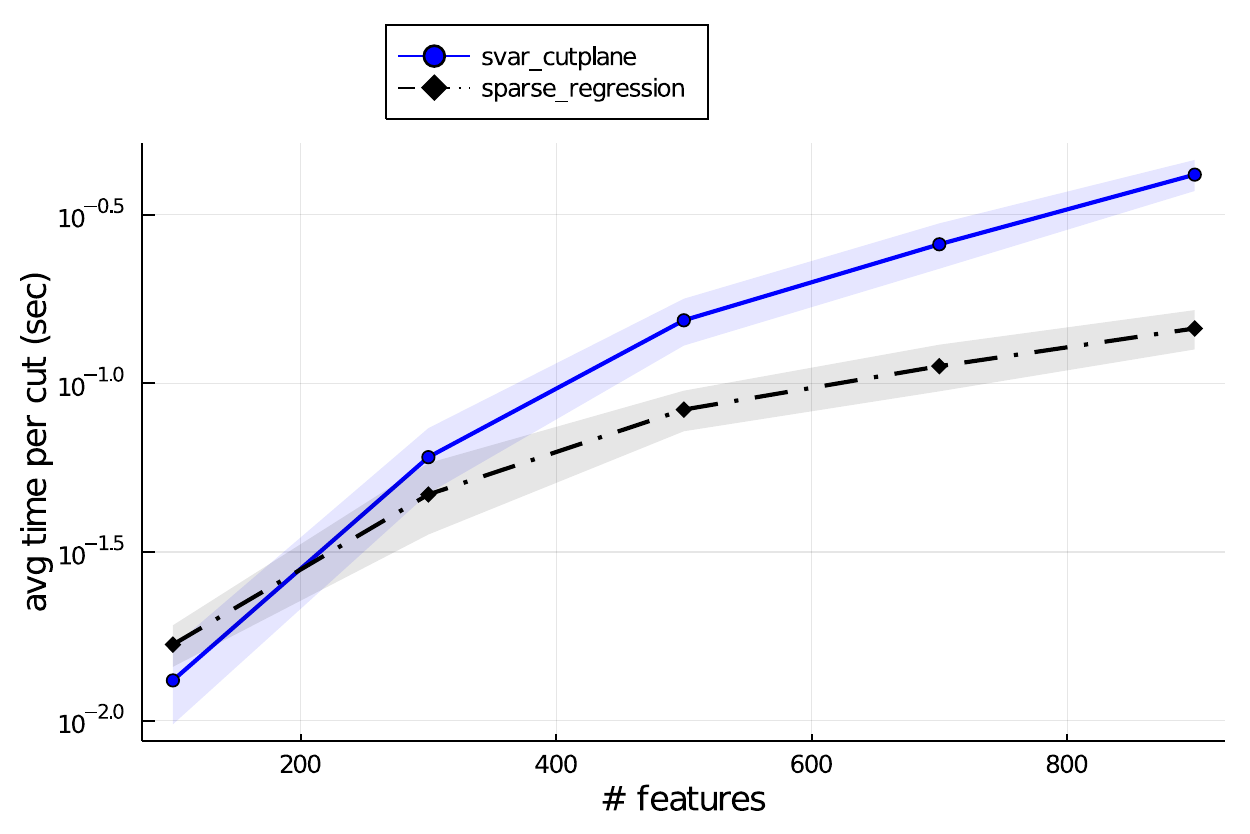}
            \caption{Average time per cut as function of D.\label{fig:synthetic_D_t_cut_avg}}
        \end{subfigure}
        \caption{Scalability of the cutting plane method with respect to $N,T,D$.\label{fig:scalability}}
    \end{adjustbox}
    \end{figure}
\end{itemize}

\end{revi}


\section{Experiments on Real-World Datasets} \label{sec:realworld}

In this section, we study the performance of the SSVR framework on publicly available real-world data. We consider two datasets with temporally and three datasets with spatially varying structure. First, we present aggregated computational results; then, we delve deeper into each dataset and discuss the models learned by the proposed framework. In Appendix \ref{sec:appendix-expers-real}, we provide further information on the datasets and our preprocessing methodology, and detailed computational results.  

\paragraph{Aggregated Results.} To obtain the aggregated results we report in this section, we randomly split each dataset 10 times into training ($60\%$), validation ($20\%$), and test ($20\%$) sets (respecting the temporal structure if such exists). For each dataset and each metric, we compute the mean and standard deviation of each method across those 10 splits. We rank the methods according to their performance and report the mean and standard deviation of the rank of each method across all experiments. Table \ref{tab:real-results} presents the aggregated results (obtained over $50$ datasets).

\begin{revi}
In agreement with our conclusions from Section \ref{sec:synthetic}, \svc, \svhc, and \rsnOne\ outperform in terms of their \textit{predictive power}, with \svc\ and \svhc\ having notably lower variance in their performance. As in real-world problems there is no way to assess estimation accuracy, we instead focus on model \textit{interpretability}; we report each method's estimated local sparsity ($\hat K_{\text{L}}$), global sparsity ($\hat K_{\text{G}}$), and number of changes in support $\hat K_{\text{C}}$. \svc, \svhc, and \svh\ always produce simpler and hence more interpretable models using, in general, the smaller number of features. Similar to the synthetic experiments, \svhc\ and \svh\ are the clear winners on \textit{computational time}. Finally, the proposed methods significantly outperform \sr\ on proving optimality and on generating fewer and faster cuts, i.e., on the \textit{evaluation of the cutting plane method}.
\end{revi}

{

\begin{table}[ht!]
\caption{Aggregated results for real-world data: mean and std. of each method's ranking across all experiments. Note that Gap, Cut Count, and ACT only apply to cutting plane-based methods.}
\label{tab:real-results}
\resizebox{\textwidth}{!}{
\begin{tabular}{lrrrrrrrr}
\toprule
\textbf{Algorithm} & \textbf{Test R$^2$} & \textbf{Local Sparsity} & \textbf{Global Sparsity} & \textbf{Changes in Support} & \textbf{Time} & \textbf{Gap} & \textbf{ACT} & \textbf{Cut Count} \\ \midrule
\svc & \textbf{2.8 (0.84)} & 2.2 (1.3) & \textbf{2.0 (1.22)} & \textbf{2.0 (1.41)} & 6.2 (1.48) & 2.6 (3.05) & 2.0 (0.0) & \textbf{1.6 (0.89)} \\
\svhc & \textbf{2.8 (0.45)} & 2.2 (2.17) & \textbf{2.0 (1.73)} & 2.2 (1.79) & 2.6 (0.89) & \textbf{1.2 (0.45)} & \textbf{1.8 (1.1)} & 1.8 (0.45) \\
\svh & 5.0 (1.58) & \textbf{2.0 (1.73)} & 7.6 (0.89) & 7.6 (0.89) & \textbf{2.0 (1.41)} & - & - & - \\
\sr & 6.6 (1.34) & 3.6 (2.41) & 2.4 (1.34) & \textbf{1.0 (0.0)} & 6.2 (2.17) & 2.4 (0.89) & 2.2 (1.1) & 2.6 (0.89) \\
\rsnOne & \textbf{2.8 (2.95)} & 6.6 (0.89) & 5.6 (0.89) & 3.0 (1.41) & 3.0 (1.22) & - & - & - \\
\rsnOneLasso & 6.4 (1.52) & 4.0 (1.41) & 3.4 (0.89) & 5.6 (0.55) & 3.2 (2.17) & - & - & - \\
\rsnTwo & 3.2 (2.68) & 7.2 (0.45) & 6.2 (0.45) & 4.6 (2.07) & 7.0 (0.71) & - & - & - \\
\rsnTwoLasso & 5.2 (2.49) & 5.2 (1.48) & 4.8 (1.1) & 6.8 (0.45) & 5.6 (0.89) & - & - & - \\ \bottomrule
\end{tabular}
}
\end{table}

}

\subsection{Datasets with Temporally Varying Structure} \label{ssec:realworld-temporal}
In this section, we delve deeper into the temporal datasets and the corresponding learned models.

\begin{revi}

\paragraph{Appliances Energy Prediction: Hourly.} In this experiment, we focus on a real-world case study concerned with appliances energy prediction \citep{candanedo2017data}. Each observation in the dataset is a vector of measurements (temperature and humidity in various rooms, weather conditions, etc.) in a low energy building, and the goal is to predict the energy consumption of the building's appliances. After preprocessing the dataset, we get $N=822$ data points per vertex, $T=24$ vertices (each corresponding to an hour of the day), and $D=26$ features; to capture the temporal structure of the problem, the similarity graph is a chain (see Figure \ref{fig:similarity-graphs} (left)); see Appendix \ref{ssec:appendix-expers-real-data} for details.

Our open-source implementation outputs the final model as a graph, with a structure matching that of the underlying similarity graph. Each vertex shows the learned regression coefficient for any selected feature at the corresponding vertex of the similarity graph; vertices in yellow (resp. blue) correspond to coefficients below (resp. above) the mean across all vertices.

Figure \ref{fig:real_energy_hour} presents the variation of the regression coefficient with the highest mean absolute magnitude across all vertices in the best \svc\ model. The corresponding feature, T4, corresponds to the temperature in the office room. $\bm \beta_{\text{T4}}$ is zero between 9pm and 10pm and, in general, takes very low values at night, when the office room is likely empty; then, it slowly increases during the day, peaks in the afternoon, and then slowly decreases in the evening. The slowly and sparsely varying structure of the learned model is clear.

\paragraph{Appliances Energy Prediction: Monthly.} In this experiment, we consider the same appliances energy dataset. However, instead of assigning a vertex to each hour of the day, we now assign a vertex to each month. We get $N=2,922$, $T=5$, $D=26$ and a chain similarity graph. Figure \ref{fig:real_energy_month} presents the variation of the same regression coefficient across all vertices in the best \svc\ model. In this case, $\bm \beta_{\text{T4}}$ varies slowly across months, having a higher impact on the model as the summer approaches (when, potentially, the use of ACs increases consumption).

\begin{figure}[!ht] 
\begin{adjustbox}{minipage=\linewidth,scale=1}
    \centering
    \begin{subfigure}[b]{0.45\linewidth}
        \centering\includegraphics[width=\textwidth]{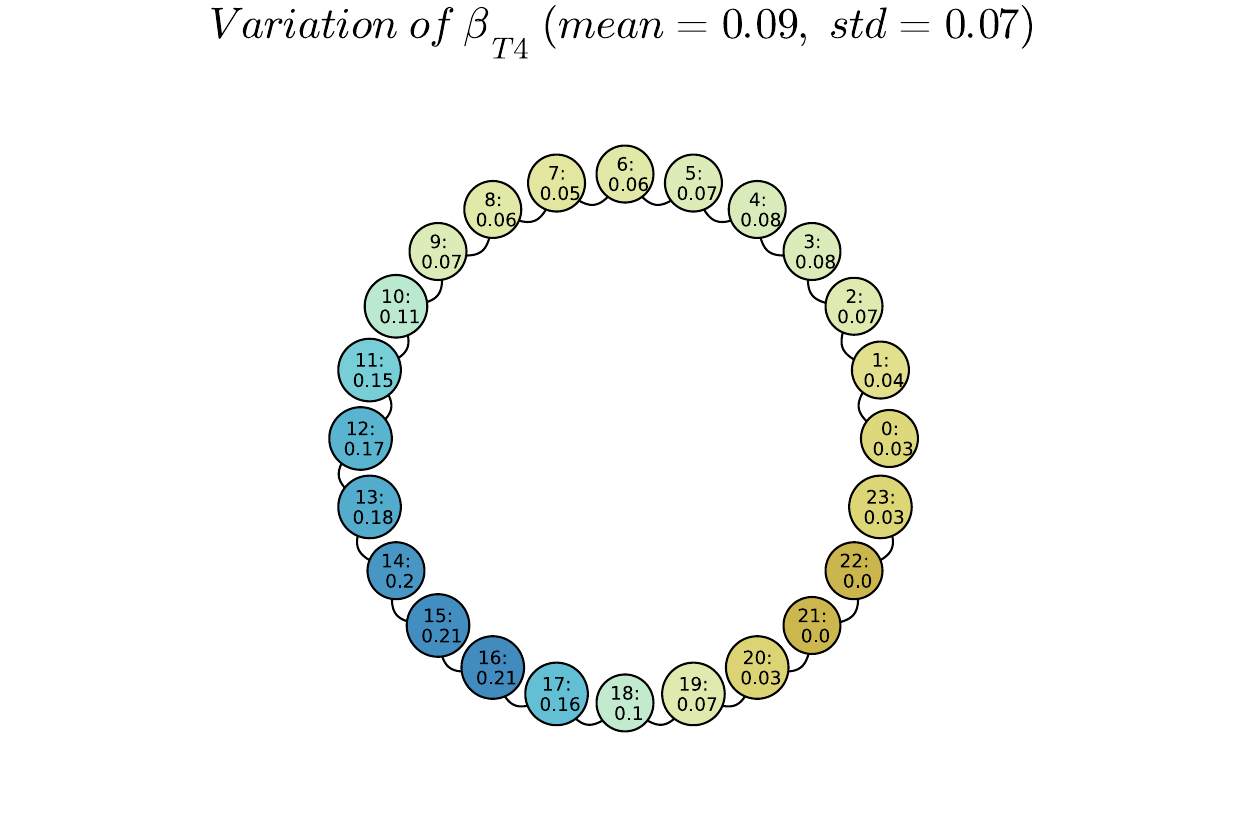}
        \caption{Appliances Energy Prediction: Hourly. \label{fig:real_energy_hour}}
    \end{subfigure}
    \hfill
    \begin{subfigure}[b]{0.45\linewidth}
        \centering\includegraphics[width=\textwidth]{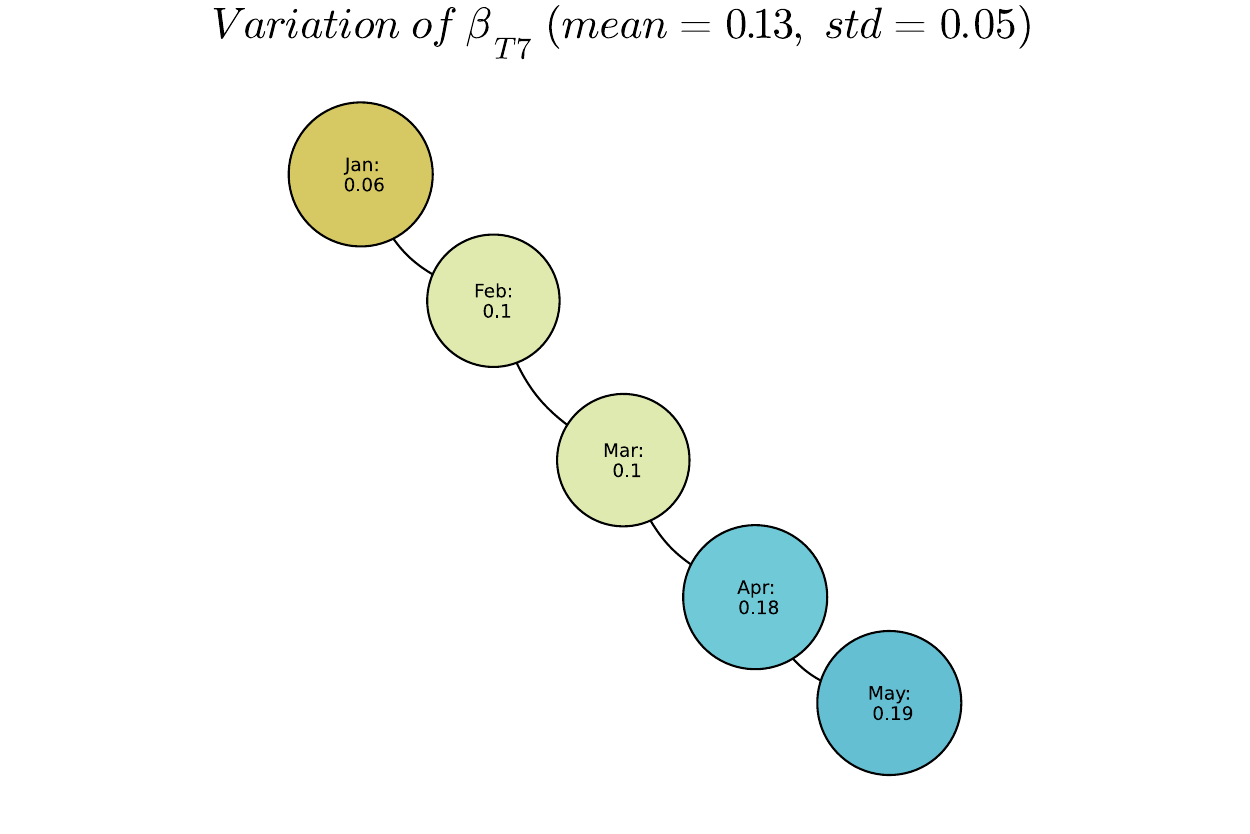}
        \caption{Appliances Energy Prediction: Monthly.\label{fig:real_energy_month}}
    \end{subfigure}
    \caption{Variation of most important feature across vertices on datasets with temporally varying structure.\label{fig:real-temporal}}
\end{adjustbox}
\end{figure}

\subsection{Datasets with Spatially Varying Structure} \label{ssec:realworld-spatial}
In this section, we discuss the details of the spatial datasets and the corresponding learned models.

\paragraph{Housing Price Prediction.} In this experiment, we explore the application of our framework to housing price prediction in Ames, Iowa \citep{de2011ames}. The dataset consists of a number of features involved in assessing home values, and the goal is to predict the selling price of the home. We get $N=822$, $T=7$ (each corresponding to a cluster of neighborhoods in Ames, Iowa; see Figure \ref{fig:housing} for a visualization), $D=199$, and a similarity graph connecting adjacent neighborhood clusters with $E=8$ edges. Figure \ref{fig:real_housing} presents the variation of the regression coefficient with the highest mean absolute magnitude across all vertices in the best \svc\ model. In this case, the corresponding feature, GrLivArea, corresponds to the above-ground living area. $\bm \beta_{\text{GrLivArea}}$ peaks in the northernmost neighborhood clusters and decreases in the southernmost clusters (W, S, E). 

\paragraph{Air Quality.} In this experiment, we consider air quality prediction in 12 air quality monitoring sites in Beijing \citep{zhang2017cautionary}. The original dataset consists of weather (temperature, pressure, dew point temperature, precipitation, wind speed, wind direction) and time-related features, and the goal is to predict PM2.5 concentration - an air pollutant that is a health concern at high levels. We get $N=35,064$, $T=12$ (each corresponding to an air quality monitoring site), $D=25$, and a similarity graph of $E=14$ edges and $4$ connected components. Figure \ref{fig:real_airquality} presents the variation of the regression coefficient with the highest mean absolute magnitude across all vertices in the best \svc\ model. In this case, the corresponding feature, DEWP, corresponds to the dew point temperature. The benefits of the proposed SSVR framework are again clear: the range of values for $\bm \beta_{\text{DEWP}}$ is between 0.38 and 0.48; however, across all connected components, the maximum coefficient variation never exceeds 0.01.

\paragraph{Meteorology.} In this experiment, we consider the task of weather prediction in 30 US and Canadian Cities, as well as 6 Israeli cities. The original dataset contains hourly measurements of weather attributes (temperature, humidity, air pressure, wind direction, and wind speed), and the goal is to predict the temperature half a day in advance. We get $N=45,231$, $T=36$ (each corresponding to a city), $D=50$ features, and a similarity graph of $E=110$ edges and $2$ connected components. Figure \ref{fig:real_meteo} presents the variation of the regression coefficient with the highest mean absolute magnitude across all vertices in the best \svc\ model. In this case, the corresponding feature, T1, corresponds, perhaps unsurprisingly, to the current temperature. The visualization of the learned model clearly shows how $\bm \beta_{\text{T1}}$ varies slowly across the similarity graph.

\begin{figure}[!ht] 
\begin{adjustbox}{minipage=\linewidth,scale=1}
    \centering
    \begin{subfigure}[b]{0.45\linewidth}
        \centering\includegraphics[width=0.6\textwidth]{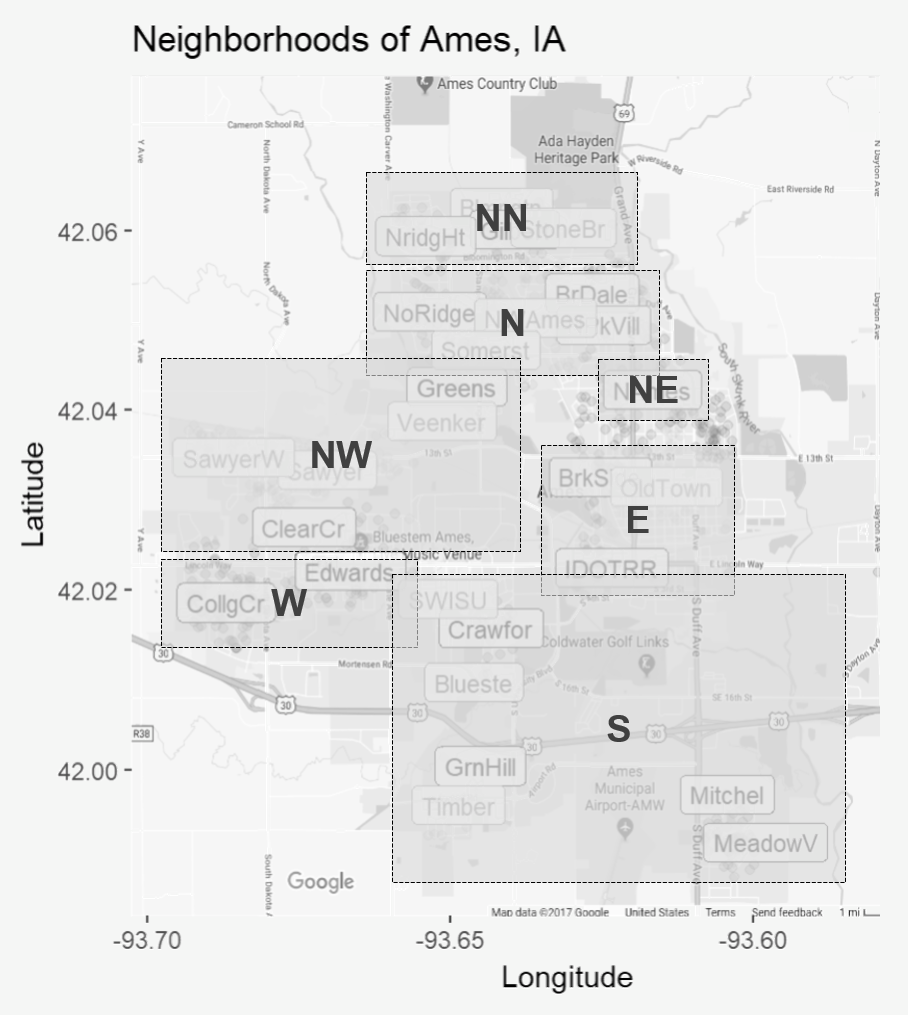}
        \caption{Neighborhood clusters in Ames, IA.\label{fig:housing}}
    \end{subfigure}
    \hfill
    \begin{subfigure}[b]{0.45\linewidth}
        \centering\includegraphics[width=\textwidth]{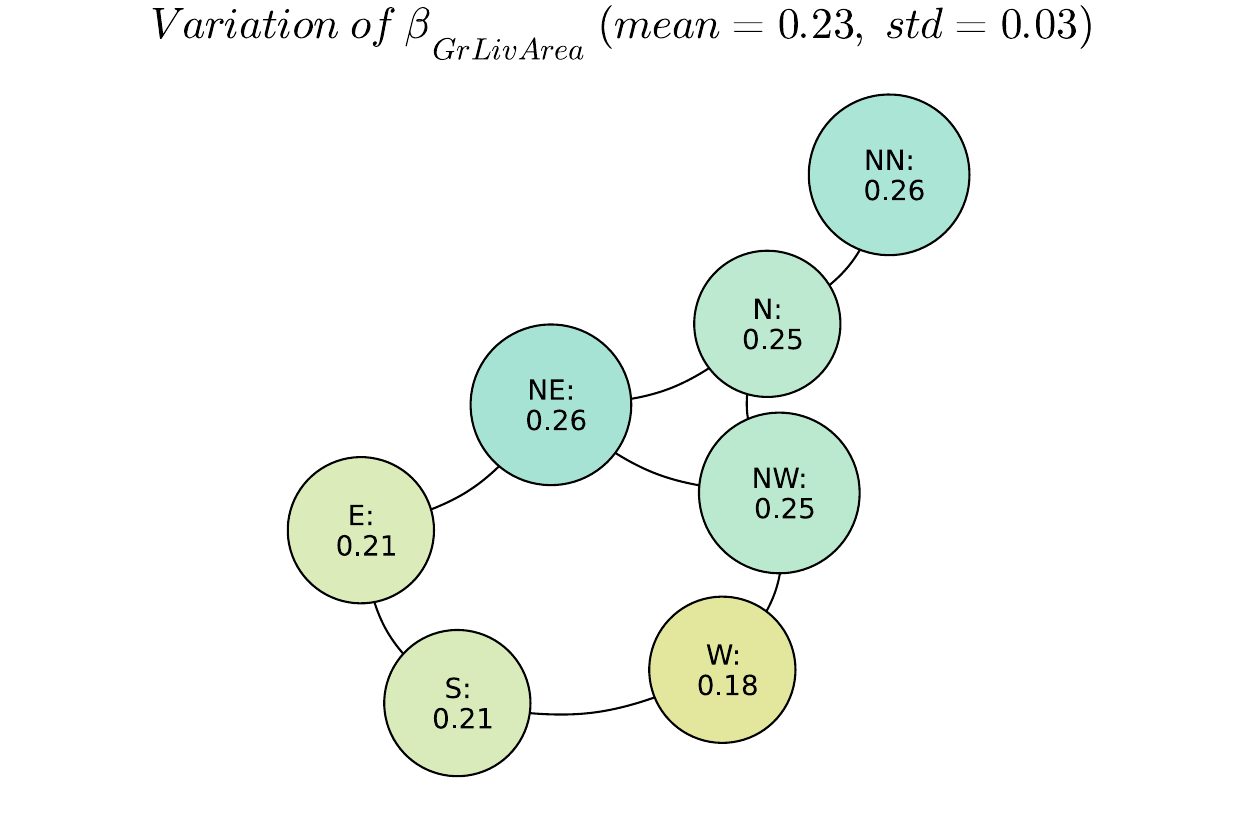}
        \caption{Housing Price Prediction.\label{fig:real_housing}}
    \end{subfigure}%
    
    \vspace{5pt}
    
    \begin{subfigure}[b]{0.45\linewidth}
        \centering\includegraphics[width=\textwidth]{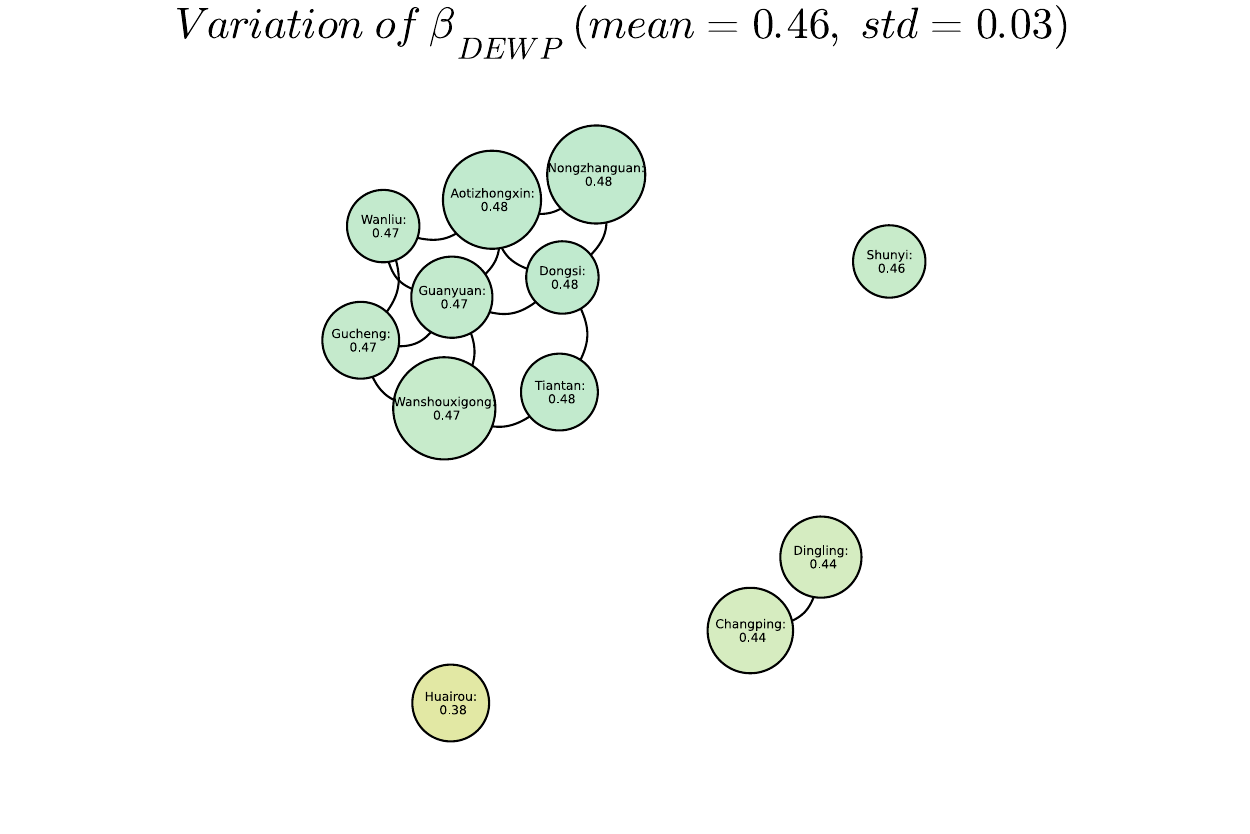}
        \caption{Air Quality. \label{fig:real_airquality}}
    \end{subfigure}
    \hfill
    \begin{subfigure}[b]{0.45\linewidth}
        \centering\includegraphics[width=\textwidth]{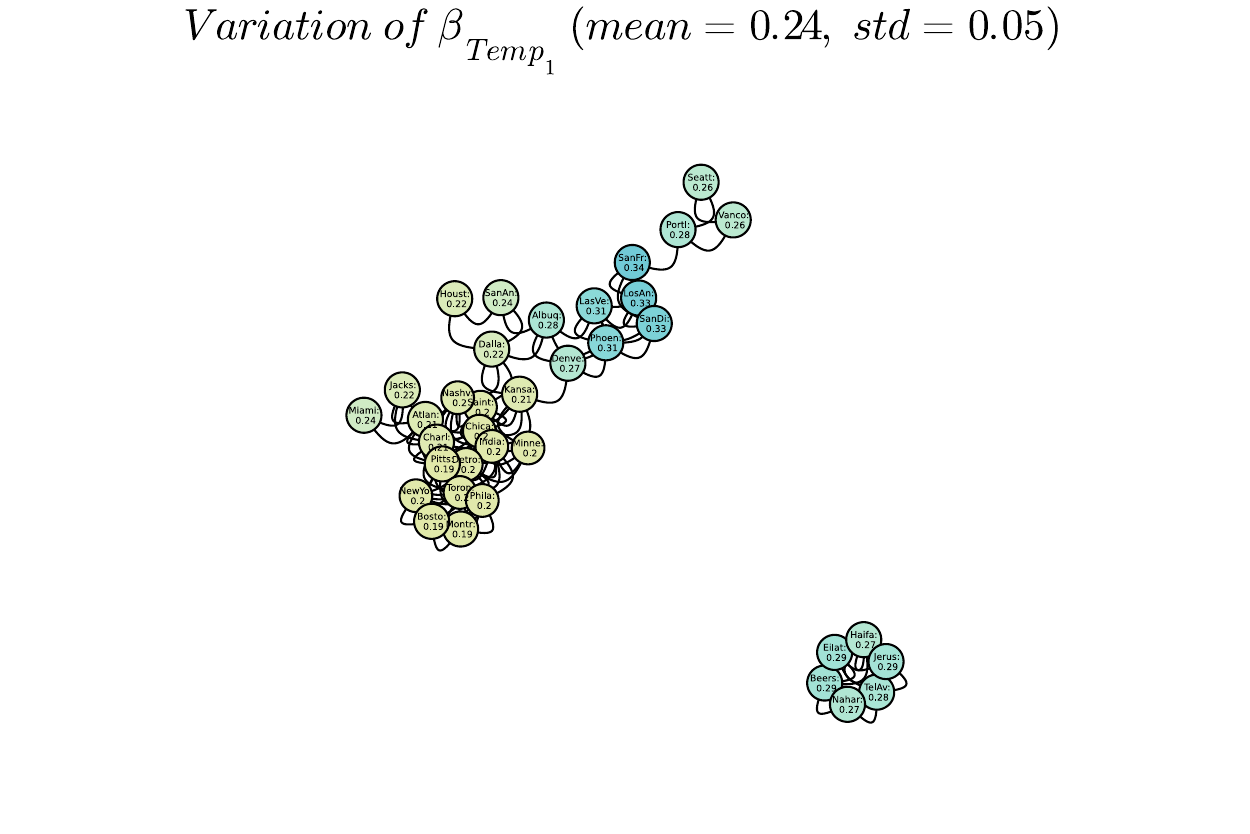}
        \caption{Meteorology.\label{fig:real_meteo}}
    \end{subfigure}
    \caption{Variation of most important feature across vertices on datasets with spatially varying structure.\label{fig:real-spatial}}
\end{adjustbox}
\end{figure}

\section{Conclusion}
In this paper, we have introduced the slowly varying regression under sparsity framework, which addresses regression problems with sparse and slowly varying structure. We have proposed a new way of solving the underlying optimization problem to optimality through a novel relaxation of the objective function. We have developed efficient exact and heuristic algorithms, as well as a practical hyperparameter tuning procedure, and have made our implementation available open-source to facilitate the use of the proposed framework by practitioners. 

Our numerical experiments demonstrate the proposed methods' superior or competitive performance in multiple dimensions—ranging from accuracy and interpretability to computational efficiency.  
Our results affirm the value proposition of our slowly varying methods, particularly \svc\ and \svhc, in handling complex regression scenarios involving sparse and slowly varying coefficients. They offer a compelling blend of accuracy, computational efficiency, and interpretability, thereby establishing their potential for a wide range of applications in sparse regression problems. Future work could explore the integration of additional regularization techniques to further fine-tune these algorithms for specific use cases.

\end{revi}

\bibliographystyle{informs2014}
\bibliography{references.bib}

\begin{APPENDICES}

\section{Technical Proofs}

\subsection{Proof of Lemma \ref{lem:reform}} \label{sec:appendix-lem:reform}

\proof{Proof}
Note that by rearranging the variables, we can rewrite the optimization problem as:
\begin{align*}
        \underset{\bm{z},\bm{s}, \bm{w} \in \ma{Z}}{\min}\;\;\underset{\bm{\beta}}{\min} \quad & 
         c(\bm{z},\bm{\beta}) := \bm{\beta}^{\top}(\bm{Z}(\bm{M}+\lambda_\beta \bm{I})\bm{Z})\bm{\beta} -2\bm{\mu}^{\top} \bm{Z} \bm{\beta} + \sum_{t=1}^T \|\bm{y}^t \|_2^2, 
\end{align*}
where $\bm{\beta}=(\bm{\beta}^1,\cdots,\bm{\beta}^T)$, $\bm{Z}=\diag(\bm{z}^1,\cdots,\bm{z}^T)$, and $\ma{Z}$ is the polyhedral feasible set as defined by the binary constraints on $\bm{z},\bm{s}, \bm{w}$ and (\ref{eq:c1_mio})--(\ref{eq:c6_mio}), and $\bm{M}$ can be defined as:
\begin{align*}
    \bm{M} &= \begin{pmatrix}
    \bm{X}^1 \\
    \bm{X}^2 \\
    \vdots\\
    \bm{X}^T    
    \end{pmatrix}\begin{pmatrix}
    \bm{X}^1 &
    \bm{X}^2 &
    \dots &
    \bm{X}^T    
    \end{pmatrix}+\begin{pmatrix}
    \mathbbm{1}_{\exists t, (1,t) \in E }-\mathbbm{1}_{\exists s, (s,1) \in E } \\
    \mathbbm{1}_{\exists t, (2,t) \in E }-\mathbbm{1}_{\exists s, (s,2) \in E }  \\
    \vdots\\
    \mathbbm{1}_{\exists t, (T,t) \in E }-\mathbbm{1}_{\exists s, (s,T) \in E }      
    \end{pmatrix}  \times \left(\begin{pmatrix}
    \mathbbm{1}_{\exists t, (1,t) \in E }-\mathbbm{1}_{\exists s, (s,1) \in E } \\
    \mathbbm{1}_{\exists t, (2,t) \in E }-\mathbbm{1}_{\exists s, (s,2) \in E }  \\
    \vdots\\
    \mathbbm{1}_{\exists t, (T,t) \in E }-\mathbbm{1}_{\exists s, (s,T) \in E }      
    \end{pmatrix} \right)^T.
\end{align*}
It is clear that the both matrices in the right expression are positive semidefinite, and thus $\bm{M}$ is positive semidefinite. We then reach the final form by dividing the objective by 2 and noting $\sum_{t=1}^T \|\bm{y}^t \|_2^2$ is a constant within the optimization problem and thus can be removed. \hfill $\square$
\endproof

\subsection{Proof of Lemma \ref{lem:first_order}} \label{sec:appendix-lem:first_order}

\proof{Proof}
We first note that our minimization problem is equivalent to the sum-of-norms original problem. Combining this with the fact that $\bm{M}$ is positive semi-definite, to solve the inner problem we only need to derive the first order condition, which is:
\begin{equation*}
    \frac{\partial c(\bm{z},\bm{\beta})}{\partial \bm{\beta}} =(\bm{Z}(\bm{M}+\lambda_\beta \bm{I})\bm{Z})\bm{\beta}- \bm{Z}\bm{\mu}=\bm{0}.
\end{equation*}
$\bm{Z}$ is a rank $\leq TK_{\text{L}}<TD$ matrix, so $(\bm{Z}(\bm{M}+\lambda_\beta \bm{I})\bm{Z})$ is rank-deficient. Thus, to solve this first-order condition, we can utilize the Moore-Penrose pseudo-inverse to write $\bm{\beta}^*(\bm{z})  = (\bm{Z}(\bm{M}+\lambda_\beta \bm{I})\bm{Z})^\dagger \bm{Z}\bm{\mu}.$

The second assertion follows from substituting the first order equality $(\bm{Z}(\bm{M}+\lambda_\beta \bm{I})\bm{Z})\bm{\beta}^*(\bm{z}) = \bm{Z}\bm{\mu}$ into the objective expression. Noting that $\bm{Z}\bm{\beta}^*(\bm{z})=\bm{\beta}^*(\bm{z})$ by definition, we have $\frac{1}{2}\bm{\beta}^*(\bm{z}) ^{\top}(\bm{Z}(\bm{M}+\lambda_\beta \bm{I})\bm{Z})\bm{\beta}^*(\bm{z})  -\bm{\mu}^{\top} \bm{Z} \bm{\beta}^*(\bm{z}) =-\frac{1}{2}\bm{\beta}^*(\bm{z})^{\top}\bm{Z}\bm{\mu}=-\frac{1}{2}\bm{\beta}^*(\bm{z})^{\top}\bm{\mu}$. \hfill $\square$
\endproof

\subsection{Proof of Proposition \ref{prop:convexify}} \label{sec:appendix-prop:convexify}

\proof{Proof}
We prove this in two steps. First, we establish the following relation for the pseudoinverse:

\begin{lem}
\label{lem:psuedo_iden}
$
(\bm{Z}(\bm{M}+\lambda_\beta \bm{I})\bm{Z})^\dagger = (\lambda_\beta \bm{I}+  \bm{Z}\bm{M}\bm{Z})^{-1}-\lambda_\beta(\bm{I}-\bm{Z}).
$
\end{lem}
\proof{Proof of Lemma \ref{lem:psuedo_iden}}
    We verify that the expression on the right satisfies the definition of a Moore-Penrose pseudoinverse for $\bm{A} := \bm{Z}(\bm{M}+\lambda_\beta \bm{I})\bm{Z}$. The Moore-Penrose pseudoinverse $\bm{A}^\dagger$ is the unique matrix that satisfies: 1. $\bm{A}^\dagger\bm{A}\bm{A}^\dagger = \bm{A}^\dagger$, 2. $\bm{A}\bm{A}^\dagger\bm{A} = \bm{A}$, 3. $(\bm{A} \bm{A}^\dagger)^*=\bm{A}\bm{A}^\dagger$, 4. $(\bm{A}^\dagger\bm{A})^*=\bm{A}^\dagger\bm{A}$, where $*$ is the Hermitian operator with $\bm{A}^*_{ij}=\overline{\bm{A}_{ji}}$. The assertions follow immediately if we have $\bm{A}^\dagger \bm{A}=\bm{A} \bm{A}^\dagger = \bm{Z}$, which we next prove:
\begin{align*}
    \bm{A}^\dagger \bm{A}&=[(\lambda_\beta \bm{I}+\bm{Z}\bm{M}\bm{Z})^{-1}-\lambda_\beta(\bm{I}-\bm{Z})]\bm{Z}(\bm{M}+\lambda_\beta \bm{I})\bm{Z}\\&=(\lambda_\beta \bm{I} + \bm{Z}\bm{M}\bm{Z})^{-1}(\bm{Z}\bm{M}\bm{Z}+\lambda_\beta \bm{Z})\\& = (\bm{I}-(\lambda_\beta \bm{I} + \bm{Z}\bm{M}\bm{Z})^{-1}\lambda_\beta(\bm{I}-\bm{Z}))\\&=(\bm{I}-\left(\frac{1}{\lambda_\beta}(\bm{I}-\bm{Z}(\lambda_\beta\bm{I}+\bm{M}\bm{Z})^{-1}\bm{M}\bm{Z})\right)\lambda_\beta(\bm{I}-\bm{Z}))\\&=\bm{Z},
\end{align*}
where on the second last line we utilized the binomial inverse theorem \citep{henderson1981deriving}. Here, we have $\bm{Z}^2=\bm{Z}$ as $\bm{Z}$ is a binary diagonal matrix. The case for $\bm{A}\bm{A}^\dagger$ is identical.  \hfill $\square$
\endproof

Then, we note the following equivalence:
\begin{lem}
\label{lem:binary_equiv}
$
    (\lambda_\beta \bm{I}+  \bm{Z}\bm{M}\bm{Z})^{-1}\bm{Z}=(\lambda_\beta \bm{I}+  \bm{Z}\bm{M})^{-1}\bm{Z}.
$
\end{lem}
\proof{Proof of Lemma \ref{lem:binary_equiv}}
    By the binomial inverse theorem \citep{henderson1981deriving}, we have:
    \[(\lambda_\beta \bm{I}+  \bm{Z}\bm{M}\bm{Z})^{-1}\bm{Z}=\frac{1}{\lambda_\beta}(\bm{I}-\bm{Z}(\lambda_\beta\bm{I}+\bm{M}\bm{Z})^{-1}\bm{M}\bm{Z})\bm{Z}=\frac{1}{\lambda_\beta}(\bm{Z}-\bm{Z}(\lambda_\beta\bm{I}+\bm{M}\bm{Z})^{-1}\bm{M}\bm{Z}).\]
    Similarly, we have:
    \[(\lambda_\beta \bm{I}+  \bm{Z}\bm{M})^{-1}\bm{Z}=\frac{1}{\lambda_\beta}(\bm{I}-\bm{Z}(\lambda_\beta\bm{I}+\bm{M}\bm{Z})^{-1}\bm{M})\bm{Z}=\frac{1}{\lambda_\beta}(\bm{Z}-\bm{Z}(\lambda_\beta\bm{I}+\bm{M}\bm{Z})^{-1}\bm{M}\bm{Z}).\]    
    This proves the statement required.  \hfill $\square$ 
\endproof

We now prove the final desired statement, utilizing Lemmata \ref{lem:psuedo_iden} and \ref{lem:binary_equiv}:
\begin{align*}
    &(\bm{Z}(\bm{M}+\lambda_\beta \bm{I})\bm{Z})^\dagger\bm{Z}=((\lambda_\beta \bm{I}+  \bm{Z}\bm{M}\bm{Z})^{-1}-\lambda_\beta(\bm{I}-\bm{Z}))\bm{Z}=(\lambda_\beta \bm{I}+  \bm{Z}\bm{M})^{-1}\bm{Z}.
\end{align*}  \hfill $\square$
\endproof

\subsection{Proof of Theorem \ref{theo:convexity}} \label{sec:appendix-theo:convexity}

\proof{Proof}
The equivalence of the two optimization problems follows immediately from Lemma \ref{lem:first_order} and Proposition \ref{prop:convexify}. We proceed to prove that $f(\bm{z})=-\frac{\bm{\mu}^{\top}\bm{\beta}^*(\bm{z})}{2}$ is convex in $\bm{z}$. Now, denote the element-wise products $(\bm{\beta} \cdot \bm{M})_{ij}=\beta_iM_{ij}$ and $(\bm{\beta} \cdot \bm{\mu})=\beta_i\mu_i$. Then, by direct calculation, the Hessian of $f(\bm{z})$ in the direction of $\bm{\beta}$ can be calculated as:
\begin{align*}
    &\bm{\beta}^{\top}\frac{\partial f(\bm{z})}{\partial \bm{z}\partial \bm{z}^{\top}} \bm{\beta}\\&=\bm{\mu}^{\top}(\lambda_\beta \bm{I}+\bm{Z}\bm{M})^{-1}(\bm{\beta}\cdot\bm{M})(\lambda_\beta\bm{I}+\bm{Z}\bm{M})^{-1}(\bm{\mu}\cdot \bm{\beta})\\&-\bm{\mu}^{\top}(\lambda_\beta\bm{I}+\bm{Z}\bm{M})^{-1}(\bm{\beta}\cdot\bm{M})(\lambda_\beta\bm{I}+\bm{Z}\bm{M})^{-1}((\bm{\beta}\cdot \bm{M})(\lambda_\beta\bm{I}+\bm{Z}\bm{M})^{-1}\bm{Z}\bm{\mu})\\&=\bm{\mu}^{\top}(\lambda_\beta\bm{I}+\bm{Z}\bm{M})^{-1}(\bm{\beta}\cdot\bm{M})(\lambda_\beta\bm{I}+\bm{Z}\bm{M})^{-1}(\bm{\beta}\cdot (\bm{I}-\bm{M}(\lambda_\beta\bm{I}+\bm{Z}\bm{M})^{-1}\bm{Z})\bm{\mu})\\&=\frac{1}{\lambda_\beta}\bm{\mu}^{\top}(\lambda_\beta \bm{I}+\bm{Z}\bm{M})^{-1}(\bm{\beta}\cdot\bm{M})(\lambda_\beta\bm{I}+\bm{Z}\bm{M})^{-1}(\bm{\beta}\cdot (\lambda_\beta\bm{I}+\bm{M}\bm{Z})^{-1}\bm{\mu})\\&=\frac{1}{\lambda_\beta}(\bm{\beta}\cdot (\lambda_\beta\bm{I}+\bm{M}\bm{Z})^{-1}\bm{\mu})^{\top}\bm{M}(\bm{I}+\bm{Z}\bm{M})^{-1}(\bm{\beta}\cdot (\lambda_\beta\bm{I}+\bm{M}\bm{Z})^{-1}\bm{\mu})\\&=\frac{1}{\lambda_\beta}(\bm{\beta}\cdot (\lambda_\beta\bm{I}+\bm{M}\bm{Z})^{-1}\bm{\mu})^{\top}\bm{M}(\bm{M}+\bm{M}\bm{Z}\bm{M})^\dagger \bm{M}(\bm{\beta}\cdot (\lambda_\beta\bm{I}+\bm{M}\bm{Z})^{-1}\bm{\mu})\\&\geq 0,
\end{align*}
where, in the second last step, we utilized:
\begin{align*}
    &\bm{M}(\bm{I}+\bm{Z}\bm{M})^{-1}=\bm{M}\bm{M}^\dagger\bm{M}(\bm{I}+\bm{Z}\bm{M})^{-1}=\bm{M}\bm{M}^\dagger(\bm{I}+\bm{M}\bm{Z})^{-1}\bm{M}= \bm{M}(\bm{M}+\bm{M}\bm{Z}\bm{M})^\dagger \bm{M}.
\end{align*}
Since $\bm{M}, \bm{Z}$ are both positive semi-definite, it is clear that $(\bm{M}+\bm{M}\bm{Z}\bm{M})^\dagger$ is positive semi-definite, and thus the Hessian of $f(\bm{z})$ in the direction of $\bm{\beta}$ is always non-negative. 
Since this inequality holds for any $\bm{\beta}$, the Hessian matrix of $f(\bm{z})$ is positive semidefinite and hence $f(\bm{z})$ is convex in $\bm{z}$.   \hfill $\square$
\endproof

\subsection{Proof of Lemma \ref{lem:gradient}} \label{sec:appendix-lem:gradient}

\proof{Proof}
We begin by differentiating matrix $\bm{K}$ with respect to $\bm{Z}$'s diagonal component $z^t_d$:
\begin{equation} \label{eqn:derivative_K}
    \begin{split}
        \frac{\partial \bm K}{\partial z^t_d}
        = \frac{\partial \bm K(\bm z)}{\partial z^t_d} 
        = \frac{\partial \left( \lambda_{\beta} \mathbb{I}+ \bm Z\bm M \right) }{\partial z^t_d}
        = \bm{E^{t}_{d}} \bm M.
    \end{split}
\end{equation}
The partial derivative of the inverse of $\bm K$ is then given by
\begin{equation} \label{eqn:derivative_K_inv}
    \begin{split}
        \frac{\partial \bm K^{-1}}{\partial z^t_d}
        = - \bm K^{-1} \frac{\partial \bm K}{\partial z^t_d} \bm K^{-1}
        = - \bm K^{-1} \bm{E^{t}_{d}} \bm M \bm K^{-1}.
    \end{split}
\end{equation}
Finally, we have
\begin{equation} \label{eqn:derivative_c}
        \frac{\partial c(\bm z)}{\partial z^t_d}
        = \frac{\partial \left( -\frac{1}{2}\bm{\mu}^{\top}\bm{K}^{-1}\bm{Z}\bm{\mu} \right)}{\partial z^t_d} = \frac{1}{2}\bm{\mu}^{\top} \bm{K}^{-1} \left( \bm{E^{t}_{d}} \bm M \bm K^{-1} \bm{Z} \ - \ \bm{E^{t}_{d}} \right)\bm{\mu}.
\end{equation}  \hfill $\square$
\endproof

\subsection{Proof of Lemma \ref{lem:cut_generation}}
\proof{Proof}
\label{sec:appendix-lem:cut_generation}
We first introduce some notation: given any vector (matrix) $\bm a$ ($\bm A$) and a binary vector $\bm{z}$ (feasible for Problem  \eqref{eq:bin_conv_reformulation}), $\bm{a}_{\bm{z}}$ ($\bm{A}_{\bm{z},:}$ or $\bm{A}_{:,\bm{z}}$) is formed by selecting all entries $(t,d)$ of vector $\bm a$ (all rows $(t,d)$ or all columns $(t,d)$ of matrix $\bm{A}$, respectively) for which $z_d^t=1$. Accordingly, the subscript $\bm{z}^c$ selects the entries/rows/columns for which $z_d^t=0$.

\paragraph{Cost function evaluation.} Define $\bm{K} = (\lambda_\beta \bm{I}+  \bm{Z}\bm{M})$. Then, given a feasible binary vector $\bm{z}$, the cost function $c(\bm{z})$ is $-\frac{1}{2}\bm{\mu}^{\top}\bm{K}^{-1}\bm{Z}\bm{\mu}$. To evaluate this equation, we first need to invert matrix $\bm{K}$. The size of matrix $\bm K$ is $TD\times TD$, so a naive implementation would require $O\left( T^3 D^3\right)$ operations. We can reduce the complexity of the inversion by exploiting the structure of the matrix as follows:
\begin{itemize}
    \item We reorder the rows and columns of matrix $\bm{K}$ so that it takes the form:
    \[
    \bm{\tilde{K}} :=\left[
    \begin{array}{cc}
    \lambda_{\beta} \bm{I} + \bm{M}_{\bm{z},\bm{z}} & \bm{M}_{\bm{z},\bm{z}^c} \\
    \bm{0} & \lambda_{\beta} \bm{I}
    \end{array}
    \right],
    \]
    where $\bm{M}_{\bm{z},\bm{z}} \in \mathbb{R}^{TK_{\text{L}}\times TK_{\text{L}}}$
    and $\bm{M}_{\bm{z},\bm{z}^c}  \in \mathbb{R}^{TK_{\text{L}}\times T(D-K_{\text{L}})}$. We then similarly reorder $\bm{\mu}$ and $\bm{Z}$ to $\bm{\tilde{\mu}}=[\bm{\mu}_{\bm{z}},\bm{\mu}_{\bm{z}^c}]$ and $\bm{\tilde{Z}}=\diag(\bm{1}_{\bm{z}}, \bm{0}_{\bm{z}^c})$. Note that the reordering does not change the objective value, and therefore the objective function is now $-\frac{1}{2}\bm{\tilde{\mu}}^{\top}\bm{\tilde{K}}^{-1}\bm{\tilde{Z}}\bm{\tilde{\mu}}.$
    
    \item We perform blockwise inversion, which gives
    \begin{equation}
        \bm{\tilde{K}}^{-1} = \begin{bmatrix}
    (\lambda_{\beta} \bm{I} + \bm{M}_{\bm{z},\bm{z}})^{-1} & \frac{1}{\lambda_\beta} (\lambda_{\beta} \bm{I} + \bm{M}_{\bm{z},\bm{z}})^{-1}\bm{M}_{\bm{z},\bm{z}^c}\\
    \bm{0} & \frac{1}{\lambda_\beta}\bm{I}
    \end{bmatrix}. \label{eq:blockwise_inv}
    \end{equation}
   Since $\bm{Z}$ has zeros on the diagonals for all $\bm{z}^c$ columns, we thus have
        \begin{equation}
        \bm{\tilde{K}}^{-1}\bm{\tilde{Z}} = \begin{bmatrix}
    (\lambda_{\beta} \bm{I} + \bm{M}_{\bm{z},\bm{z}})^{-1} & \bm{0}\\
    \bm{0} & \bm{0}
    \end{bmatrix} 
    \end{equation}
    and therefore the objective function can be now written as
    $-\frac{1}{2}\bm{\tilde{\mu}}^{\top}\bm{\tilde{K}}^{-1}\bm{\tilde{Z}}\bm{\tilde{\mu}}=-\frac{1}{2}\bm{\mu}^{\top}_{\bm{z}}(\lambda_{\beta} \bm{I} + \bm{M}_{\bm{z},\bm{z}})^{-1}\bm{\mu}_{\bm{z}}.$
    Noting that the matrix $(\lambda_{\beta} \bm{I} + \bm{M}_{\bm{z},\bm{z}})$ has block tri-diagonal structure, with $T$ blocks of size $K_{\text{L}} \times K_{\text{L}}$ each, its inverse $(\lambda_{\beta} \bm{I} + \bm{M}_{\bm{z},\bm{z}})^{-1}$ can be computed by recursive application of blockwise inversion in $O\left( T^2 K_{\text{L}}^2 (T+K_{\text{L}})\right)$ operations. 
    \item The remaining operations to evaluate the objective are the vector-matrix multiplications with $\bm{\mu}_{\bm{z}}$, which require $O(T^2K_L^2)$ operations. 
\end{itemize}

\paragraph{Gradient evaluation.} We compute each of the $TD$ gradient entries as per Lemma \ref{lem:gradient}:
$\frac{\partial c(\bm z)}{\partial z^t_d}=\frac{1}{2}\bm{\mu}^{\top} \bm{K}^{-1} \left( \bm{E^{t}_{d}} \bm M \bm K^{-1} \bm{Z} \ - \ \bm{E^{t}_{d}} \right)\bm{\mu}.$
$\bm{v}^{i} \in \mathbb{R}^{TD}$ denotes auxiliary vectors, and we work as follows:
\begin{itemize}
    \item We compute $\bm{K}^{-1} \bm{Z} \bm{\mu}$. Noting that $\bm{K}^{-1} \bm{Z}$ selects the columns $(t,d)$ of $\bm{K}^{-1}$ for which $z_d^t=1$ and sets the remaining columns to $0$, and observing that $\bm{K}^{-1}_{\bm{z}^c,\bm{z}}=\bm 0$ from Equation \ref{eq:blockwise_inv}, we in fact only need to compute 
    $\bm{v^0}_{\bm{z}} = \bm{K}^{-1}_{\bm{z},\bm{z}} \bm{\mu}_{\bm{z}}.$
    The remaining entries of $\bm{v^0} \in \mathbb{R}^{TD}$, i.e., $\bm{v^0}_{\bm{z}^c}$, are set to $0$. This only needs to be performed once independent of which gradient entry is being computed, and we have $\bm{K}^{-1}$ from the evaluation of the cost function.
    Thus, the complexity is $O\left(T^2 K_{\text{L}}^2 \right)$ operations.
    
    \item We compute  $\bm{\mu}^{\top} \bm{K}^{-1}$, namely, $\bm{v^1}_{\bm z} = \left( \bm{\mu}^{\top}_{\bm z} \bm{K}^{-1}_{\bm z, \bm z}\right)^{\top},$
    which requires $O\left(T^2 K_{\text{L}}^2 \right)$ operations.
    To compute the remaining entries of $\bm{v^1} \in \mathbb{R}^{TD}$, i.e., $\bm{v^1}_{\bm{z}^c}$, we again reorder $\bm{\mu}$ to be $\bm{\tilde{\mu}}:=[\bm{\mu}_{\bm{z}},\bm{\mu}_{\bm{z}^c}]$ similar to above, and then use the formula indicated in Equation \eqref{eq:blockwise_inv}. Put together, we have
    \begin{align} \label{eq:v1}
        \bm{v^1}
        & = \begin{bmatrix}
        \bm{\mu}_{\bm{z}}^\top(\lambda_{\beta} \bm{I} + \bm{M}_{\bm{z},\bm{z}})^{-1} 
        & \quad \frac{1}{\lambda_\beta}\left( \bm{\mu}_{\bm{z}}^\top(\lambda_{\beta} \bm{I} + \bm{M}_{\bm{z},\bm{z}})^{-1}\bm{M}_{\bm{z},\bm{z}^c}+\bm{\mu}_{\bm{z}^c}^\top\right)        \end{bmatrix} \nonumber \\
        & = \begin{bmatrix}\bm{v^1}_{\bm z} 
        & \quad \frac{1}{\lambda_\beta}\left(\bm{v^1}_{\bm z}\bm{M}_{\bm{z},\bm{z}^c}+\bm{\mu}_{\bm{z}^c}^\top\right)\end{bmatrix},
    \end{align}
    which requires $O\left(T^2 K_{\text{L}} D \right)$ operations.
    The above steps only need to be performed once, independently of which entry of the gradient is being computed.
    Overall, the complexity is $O\left(T^2 K_{\text{L}} D \right)$ operations.
    
    \item For each $(t,d)$, we compute the multiplication $\left(\bm{E^{t}_{d}} \bm M\right) \left( \bm{K}^{-1} \bm{Z} \bm{\mu} \right)$. Noting that the multiplication $\bm{E^{t}_{d}} \bm M$ yields a matrix that is nonzero only at row $(t,d)$, and since the result is multiplied with the vector $\bm{K}^{-1} \bm{Z} \bm{\mu}$, we implement the multiplication as
    $\bm{v^2}_{(t,d)} = \bm{M}_{(t,d),\bm{z}} \bm{v^0}_{\bm z}.$
    This requires $O\left(T^2 D K_{\text{L}} \right)$ operations in total across all $(t,d)$.
    
    \item For each $(t,d)$, we compute the multiplication $\left(\bm{\mu}^{\top} \bm{K}^{-1} \right) \left( \bm{E^{t}_{d}} \bm M \bm{K}^{-1} \bm{Z} \bm{\mu} \right)$. We  first note that  $\bm{E^{t}_{d}}=\bm{E^{t}_{d}}\bm{E^{t}_{d}}$ and hence the multiplication can be rewritten as:
    $\left(\bm{\mu}^{\top} \bm{K}^{-1} \bm{E^{t}_{d}}\right) \left( \bm{E^{t}_{d}} \bm M \bm{K}^{-1} \bm{Z} \bm{\mu} \right)=\left(\bm{\mu}^{\top} \bm{K}^{-1} \bm{E^{t}_{d}}\right) \left( \bm{E^{t}_{d}} \bm{v}^2\right)$.
    Therefore, the first term selects the $(t,d)$ column of $\bm{\mu}^{\top} \bm{K}^{-1} $ and the second term selects the $(t,d)$ row of $\bm{v}^2$. 
    Using this fact, along with Equation \eqref{eq:v1}, we can now go back and calculate the final product $\bm{v^3}_{(t,d)} 
    = \left(\bm{\mu}^{\top} \bm{K}^{-1} \bm{E^{t}_{d}}\right) \left( \bm{E^{t}_{d}} \bm M \bm{K}^{-1} \bm{Z} \bm{\mu} \right) 
    = \bm{v^1}_{(t,d)} \cdot \bm{v^2}_{(t,d)}.$
    The complexity is $O\left(T D \right)$ operations in total across all $(t,d)$.
    
    
    

    \item For each $(t,d)$, we compute the multiplication $\bm{\mu}^{\top} \bm{K}^{-1} \bm{E^{t}_{d}} \bm{\mu}$ as
    $\bm{v^4}_{(t,d)} = \bm{v^1}_{(t,d)} \cdot \bm{\mu}_{(t,d)} + \mathbbm{1}_{\left( z^t_d = 0 \right)} \frac{(\bm{\mu}_{(t,d)})^2}{\lambda_{\beta}}.$
    This requires $O\left(TD \right)$ operations in total across all $(t,d)$.
    
    
    \item For each $(t,d)$, we compute ($O\left(TD \right)$ operations in total across all $(t,d)$) the corresponding entry of the gradient as
    $\frac{\partial c(\bm z)}{\partial z^t_d} = \frac{\bm{v^3}_{(t,d)}-\bm{v^4}_{(t,d)}}{2}.$
\end{itemize}
After completing the steps outlined above, we have the ingredients to compute all entries of the gradient $\nabla_{\bm z} c(\bm z)$. In total, the cost is $O\left(T^2 D K_{\text{L}} \right)$ operations.

\paragraph{Cut generation.} The complexity of the entire process is $O\left( T^2 K_{\text{L}} [K_{\text{L}}(T + K_{\text{L}}) + D] \right).$ \hfill $\square$

{\rev

\subsection{Proof of Lemma \ref{lem:upper-bound}}
\label{sec:appendix-lem:upper-bound}
\proof{Proof}
Let us denote by $\mathcal{Z}_\beta$ the feasible set defined by Equations \eqref{eq:indsparse}-\eqref{eq:sparsevary}. We upper bound Problem \eqref{eq:obj_orig}-\eqref{eq:sparsevary} as follows:
\begin{align}
    & \underset{\bm{\beta} \in \mathcal{Z}_\beta}{\min}
    \quad  \sum_{n=1}^N \sum_{t=1}^T \left(y_n^t- \sum_{d=1}^D X_{n,d}^t \beta_d^t\right)^2
    + \lambda_{\beta} \sum_{t=1}^T \sum_{d=1}^D \left( \beta_d^t\right)^2
    + \lambda_{\delta} \sum_{(s,t) \in E} \sum_{d=1}^D \left( \beta_d^t - \beta_d^s \right)^2 \label{eqn:separability-1}\\
    & \leq
    \frac{1}{D} \sum_{d=1}^D \quad \underset{\bm{\beta}_d }{\min} 
    \quad  
    \sum_{n=1}^N \sum_{t=1}^T  \left(y_n^t- X_{n,d}^t \beta_d^t\right)^2
    + \lambda_{\beta} \sum_{t=1}^T \left( \beta_d^t\right)^2
    + \lambda_{\delta} \sum_{(s,t) \in E} \left( \beta_d^t - \beta_d^s \right)^2\label{eqn:separability-2}\\
    & =
    \underset{\bm{\beta} }{\min} 
    \quad  \frac{1}{D} \sum_{d=1}^D \left(
    \sum_{n=1}^N \sum_{t=1}^T  \left(y_n^t- X_{n,d}^t \beta_d^t\right)^2
    + \lambda_{\beta} \sum_{t=1}^T \left( \beta_d^t\right)^2
    + \lambda_{\delta} \sum_{(s,t) \in E} \left( \beta_d^t - \beta_d^s \right)^2 \right)\label{eqn:separability-3}\\
    & \leq
    \underset{\bm{\beta} \in \mathcal{Z}_\beta}{\min}
    \quad  
    \frac{1}{D} \sum_{n=1}^N \sum_{t=1}^T \sum_{d=1}^D \left(y_n^t- X_{n,d}^t \beta_d^t\right)^2
    + \lambda_{\beta} \sum_{t=1}^T \sum_{d=1}^D \left( \beta_d^t\right)^2
    + \lambda_{\delta} \sum_{(s,t) \in E} \sum_{d=1}^D \left( \beta_d^t - \beta_d^s \right)^2\label{eqn:separability-4}\\
    & \leq
    \underset{\bm{\beta} \in \mathcal{Z}_\beta}{\min}
    \quad  
    \frac{1}{D} \sum_{n=1}^N \sum_{t=1}^T \sum_{d=1}^D \left(y_n^t- X_{n,d}^t \beta_d^t\right)^2
    + \lambda_{\beta} \sum_{t=1}^T \sum_{d=1}^D \left( \beta_d^t\right)^2
    + \lambda_{\delta} \sum_{(s,t) \in E} \sum_{d=1}^D 2 [\left( \beta_d^t \right)^2 + \left( \beta_d^s \right)^2]\label{eqn:separability-5}\\
    & =
    \underset{\bm{\beta} \in \mathcal{Z}_\beta}{\min}
    \quad  
    \frac{1}{D} \sum_{n=1}^N \sum_{t=1}^T \sum_{d=1}^D \left(y_n^t- X_{n,d}^t \beta_d^t\right)^2
    + \lambda_{\beta} \sum_{t=1}^T \sum_{d=1}^D \left( \beta_d^t\right)^2
    + \lambda_{\delta} \sum_{t=1}^T \sum_{d=1}^D 2 d^t \left( \beta_d^t \right)^2. \label{eqn:separability-6}
\end{align}
For the \textbf{first inequality}, in \eqref{eqn:separability-1} we have the prediction error of the best multivariate model, which by definition is less than or equal to the error of any univariate model. This can therefore be upper bounded by the average error among all univariate models, which is what we have in \eqref{eqn:separability-2}. Observe that the best among these univariate models is indeed feasible for the minimization problem in \eqref{eqn:separability-1}. 
The \textbf{equality} between \eqref{eqn:separability-2} and \ref{eqn:separability-3} is due to separability. 
For the \textbf{second inequality}, in \ref{eqn:separability-3} we have an unconstrained problem, whereas, in \ref{eqn:separability-4} we require that $\bm{\beta} \in \mathcal{Z}_\beta$ hence restricting the feasible set. Moreover, in \ref{eqn:separability-4} we rescale the regularization term and the slowly varying penalty with $D$ so that their relative importance compared to the prediction error is in the same order as in the original problem. 
For the \textbf{third inequality}, we trivially bound the squares of the differences between coefficients in adjacent vertices. \hfill $\square$
\endproof
}

\subsection{Proof of Proposition \ref{prop:heuristic}} \label{sec:appendix-prop:heuristic}

\proof{Proof}
The termination condition of Algorithm \ref{alg:heuristic} guarantees that, at termination, the solution must be a feasible solution for Problem \eqref{eq:obj_orig}-\eqref{eq:sparsevary}. Therefore, we only need to prove that Algorithm \ref{alg:heuristic} terminates in polynomial time, which we do step-by-step.

Algorithm \ref{alg:heuristic}'s first step computes the loss for each vertex-feature pair. This requires solving $TD$ univariate regularized least squares problems and can be done in closed form in time $O(NTD)$.

The second step involves solving a linear optimization problem over $TD$ variables. This can be done in time $\tilde{O}((TD)^{2+\nicefrac{1}{6}})$ using the algorithm by \cite{cohen2021solving}. (We note that $\tilde{O}(\cdot)$ gives the asymptotic complexity ignoring logarithmic factors.) 

The third step involves ensuring integrality by iterating over all entries of the linear optimization problem's solution, and can be done in time $O(TD)$.

The fourth step involves ensuring feasibility by removing one feature at a time as long as the linear optimization problem's solution is infeasible. Let us denote by $S$ the global support (across all vertices) of the estimated regression coefficients.
Since, after each iteration, we remove one feature from $S$, the global sparsity constraint \eqref{eq:globsparse} is guaranteed to be satisfied after at most $D -  K_{\text{G}}$ iterations.
Similarly, after at most $D - (K_{\text{L}}+\frac{K_{\text{C}}}{2})$ iterations, all vertices will be constrained to include the same set of $K_{\text{L}}+\frac{K_{\text{C}}}{2}$ features and hence any pair of similar regressions will differ in at most $K_{\text{C}}$ features.
Therefore, the while loop terminates in at most
\[\max\{D -  K_{\text{G}},D - (K_{\text{L}}+\frac{K_{\text{C}}}{2})\}\]
iterations, which gives an asymptotic complexity of $O(D).$ The complexity of each iteration is $O(T)$: in an efficient implementation, the first two steps inside the while loop are performed once, and, in each iteration, we only update the corresponding data structures in $O(1)$ time. Therefore, the fourth step can be done in time $O(TD).$

The fifth step involves computing $\tilde{\bm{\beta}}$ for the estimated support $\tilde{\bm{z}}$. This can be done in time $O(T^2 K_{\text{L}}^2 (T + K_{\text{L}}))$ using the procedure described in the proof of Lemma \ref{lem:cut_generation}.

Thus, Algorithm \ref{alg:heuristic} terminates in polynomial time with a feasible solution to Problem \eqref{eq:obj_orig}-\eqref{eq:sparsevary}.  \hfill $\square$
\endproof

\begin{revi}
\subsection{Proof of Proposition \ref{prop:elbow_shape}} \label{sec:appendix-prop:elbow_shape}

We first prove the result for Algorithm \ref{alg:cutplane} with validation cost function $c_V^*$. 
For any parameter set $(K_L, K_G, K_C)$, as the number of training samples $N\to \infty$, the squared loss $\sum_{t=1}^T\left\|\bm{y}^t-\bm{X}^t\bm{\beta}^t\right\|_2^2$ within Equation \eqref{eq:obj_orig} becomes the dominant term as it is the only term that scales with $N$.  
For any parameter set $(K_L, K_G, K_C)$ with $K_L\geq K_L^*, K_G\geq K_G^*, \text{ and } K_C\geq K_C^*$, as number of training samples $N \to \infty$, the minimizer $\bm{\beta}^*(\boldsymbol{\lambda}) \to \bm{\beta}^0$ by standard asymptotic theory of linear regression. \cite[e.g.,][]{dasgupta2008asymptotic} Therefore, as $N\to \infty$, the validation cost on $N_V$ samples tend to:
\begin{align*}
    c_V^*(\boldsymbol{\lambda}) \quad 
    \to  \quad N_V\sum_{t=1}^T (\epsilon_n^t)^2
        + \lambda_{\beta} \sum_{t=1}^T \| \bm{\beta}^{0^t} \|_2^2 
        + \lambda_{\delta} \sum_{(s,t) \in E} \| \bm{\beta}^{0^t} - \bm{\beta}^{0^s} \|_2^2.
\end{align*}

Alternatively, for any parameter set $(K_L, K_G, K_C)$ with $K_L< K_L^*$, $ K_G< K_G^*$, \text{ or } $K_C< K_C^*$, as number of training samples $N \to \infty$, the minimizer $\bm{\beta}^*(\boldsymbol{\lambda}) \to \tilde{\bm{\beta}} \neq \bm{\beta}^0$  as $\bm{\beta}^0$ is infeasible. Therefore, as $N\to \infty$, the validation cost on $N_V$ samples tend to:
\begin{align*}
    c_V^*(\boldsymbol{\lambda})\quad 
    \to  \quad \sum_{t=1}^T \|\bm{y}^t-\bm{X}^t\tilde{\bm{\beta}} ^t\|^2
        + \lambda_{\beta} \sum_{t=1}^T \| \tilde{\bm{\beta}}^{t} \|_2^2 
        + \lambda_{\delta} \sum_{(s,t) \in E} \| \tilde{\bm{\beta}}^{t} - \tilde{\bm{\beta}}^{s} \|_2^2.
\end{align*}
Now, by construction, $\bm{\beta}^0$ is the unique optimal solution of the linear regression problem:
\[\min_{\bm{\beta}^t} \sum_{t=1}^T \E[\left\|y_n^t-\bm{x}_n^t\bm{\beta}^t\right\|_2^2]= \sum_{t=1}^T \E[(\epsilon_n^t)^2].\]
Thus, since $\tilde{\bm{\beta}} \neq \bm{\beta}^0$, we have that, as $N_V \to \infty$:
\begin{equation*}
    \frac{\sum_{t=1}^T \|\bm{y}^t-\bm{X}^t\tilde{\bm{\beta}}^t\|^2}{N_V}\to \sum_{t=1}^T \E[\|y_n^t-\bm{x}_n^t\tilde{\bm{\beta}}^t\|^2]> \sum_{t=1}^T \E[(\epsilon_n^t)^2].
\end{equation*}
In particular, we have that:
\begin{equation*}
    \sum_{t=1}^T \|\bm{y}^t-\bm{X}^t\tilde{\bm{\beta}}^t\|^2-\sum_{t=1}^T N_V\E[(\epsilon_n^t)^2]  \to \infty.
\end{equation*}
Then, using the assumptions outlined in Proposition \ref{prop:elbow_shape}, as the number of validation samples $N_V \to \infty$ we have:
\begin{align*}
c^*_V(\boldsymbol{\lambda})- \sum_{t=1}^T N_V\E[(\epsilon_n^t)^2] &\rightarrow \begin{cases}
    \infty, & \text{if } K_L< K_L^* \text{ or } K_G< K_G^* \text{ or } K_C< K_C^*,\\
    \displaystyle C,  & \text{otherwise,}
\end{cases}\\
\tilde{c}_V(\boldsymbol{\lambda})- \sum_{t=1}^T N_V\E[(\epsilon_n^t)^2] &\rightarrow \begin{cases}
    \infty, & \text{if } K_L< K_L^* \text{ or } K_G< K_G^* \text{ or } K_C< K_C^*,\\
    \tilde{C},  & \text{otherwise,}
\end{cases}
\end{align*}
where $C,\tilde{C}<\infty$. The result then follows.

For Algorithm \ref{alg:heuristic} with validation cost function $\tilde{c_V}$, note that the per feature minimization objective function is
$\underset{\beta_d^t}{\min} 
            \frac{1}{D} \sum_{n=1}^N \left(y_n^t- X_{n,d}^t \beta_d^t\right)^2
            + \lambda_{\beta} (\beta_d^t)^2
            + \lambda_{\delta} 2 d^t ( \beta_d^t )^2. $
Therefore, as $N \to \infty$, the squared loss term again dominates, and thus we have $\tilde{\bm{\beta}}(\boldsymbol{\lambda}) \to \bm{{\beta^0}}.$ The remaining proof is identical to the case for Algorithm \ref{alg:cutplane}.

\end{revi}

\section{Algorithms and Software} \label{sec:appendix-expers-algorithms-software}
In this section, we give the implementation details of the algorithms which we compare in our experiments. 
For a fair comparison, we implement all algorithms in \verb|Julia| programming language (version 1.6) and using the \verb|JuMP.jl| modeling language for mathematical optimization (version 0.21). We solve the optimization models using the \verb|Gurobi| commercial solver (version 9.5). All experiments were performed on a standard Intel(R) Xeon(R) CPU E5-2690 @ 2.90GHz running CentOS release 7. We make our code available at \url{https://github.com/vvdigalakis/SSVRegression.git}.

We consider the following algorithms:
\begin{itemize}
    \item \textit{Sparse regression:} We fit a single (static) sparse regression model across all vertices. Note that, as a result, this approach uses $N' = NT$ data points to train $D$ parameters (since the same set of parameters is estimated across all vertices). We solve the sparse regression formulation, as shown in Problem \eqref{eq:sparse_reg}, using the cutting plane algorithm by \cite{bertsimas2020sparse} and the \verb|Gurobi| solver. We refer to this approach as \sr.
    
    \begin{revi}
    \item \textit{Sum-of-norms regularization:} We fit a slowly varying regression model in which penalize the sum across all pairs of adjacent vertices of the $\ell_p$ difference, for $p\in\{0,1\}$, between the corresponding coefficients, as shown in Problem \eqref{eq:boyd} \citep{ohlsson2010segmentation}. In the $p=1$ case, the resulting problem can be reformulated as a quadratic optimization problem. In the $p=2$ case, the resulting problem can be reformulated as a second-order cone optimization problem. In both cases, we directly solve the resulting problems using \verb|Gurobi|. We refer to this approach as \rsnP.

    \item \textit{Sum-of-norms and lasso regularization:} We expand the \rsnP approach with an $\ell_1$ penalty on the coefficients to add robustness and -hopefully- encourage some level of sparsity. We again reformulate the resulting problem and solve either as a linear optimization problem using \verb|Gurobi| (for $\ell_1$) or using ADMM (for $\ell_2$ --- see \cite{hallac2017snapvx}). We refer to this approach as \rsnPLasso.

    \item \textit{SSVR via the heuristic algorithm:} We implement Algorithm \ref{alg:heuristic} using the \verb|Gurobi| solver. We refer to this approach as \svh.
    
    \item \textit{SSVR via the exact cutting plane algorithm:} We implement Algorithm \ref{alg:cutplane} using the \verb|Gurobi| solver. We refer to this approach as \svc.
    
    \item \textit{SSVR via the hybrid algorithm:} We combine \svh\ (for hyperparameter tuning) with \svc\ (for refitting the final model). We refer to this approach as \svhc.
    \end{revi}
\end{itemize}
\begin{revi}
For all methods, we impose a time limit of $900$ seconds; if no solution is returned when the solver terminates, we return the all-zeros solution. We remark that the solver may not stop immediately upon hitting the time limit; it will instead stop after performing the required additional computations of the attributes associated with the terminated optimization \citep{gurobi}. Moreover, if the solution time of a method exceeded 1 hour in preliminary experiments, we did not include this method in our reported experiments. 

Each of the above models is hyperparameter tuned using holdout validation and exhaustive grid search over the cross product of the selected ranges of values of regularization hyperparameters. Specifically, we consider 5 values for $\lambda_\beta$ and 5 values for $\lambda_\delta$, each starting at $N$ and decreasing by a factor of 2 to obtain each next value. For \svc, \svh, and \sr, we estimate the final coefficients using a regularization weight of $\sqrt{\lambda_\beta^\star}$, where $\lambda_\beta^\star$ is the regularization weight selected through the validation process; we empirically observe that such an approach slightly improves the performance of these methods. 

\end{revi}

\section{Extended Numerical Experiments on Synthetic Data} \label{sec:appendix-expers}

In this section, we provide more detailed information on our computational study on synthetic data.

\subsection{Synthetic Data Generation and Evaluation Methodology} \label{sec:appendix-expers-data-generation}

In this section, we provide the details of the data generation and evaluation methodology we use in our synthetic data experiments in Section \ref{sec:synthetic} as well as the remaining results from our sensitivity analysis.

\paragraph*{Ground truth coefficients.} 

We generate a matrix of ground truth coefficients $\boldsymbol{\beta} \in \mathbb{R}^{T \times D}$. Each element $\boldsymbol{\beta}^t \in \mathbb{R}^{D}$ is the vector of coefficients of the regression at vertex $t \in [T]$. We focus on the spatially varying case, where the similarity graph is a general graph, as the temporally varying case is essentially a special case. To generate $\bm \beta$, we control the parameters presented in Table \ref{tab:data-generation}. 

\begin{table}[ht!]
\caption{Data generation parameters.}
\label{tab:data-generation}
\centering
\resizebox{\textwidth}{!}{
\centering
\begin{tabular}{rl}
\toprule
\textbf{Parameter} & \textbf{Explanation} \\ \midrule
$K_{\text{L}} \in \mathbb{Z}^{+}$ & Local sparsity, as detailed in Equation \eqref{eq:indsparse}. \\ 
$K_{\text{G}} \in \mathbb{Z}^{+}$ & Global sparsity, as detailed in Equation \eqref{eq:globsparse}. \\ 
$K_{\text{C}} \in \mathbb{Z}^{+}$ & Number of changes in support, as detailed in Equation \eqref{eq:sparsevary}. \\ 
$\sigma_v \in [0,1]$ & Maximum $\%$ of change in coefficients between similar vertices; drawn uniformly at random from $[-\sigma_v, +\sigma_v]$. \\  
$d_G \in \mathbb{R}^+$ & Similarity graph density. \\ 
$\rho_d \in [0,1]$ & Correlation across features. \\ 
$\xi \in \mathbb{R}^{+}$ & Signal-to-noise ratio for the noise added to the outcome variable. \\ \bottomrule
\end{tabular}
}
\end{table}

Given the data generation parameters, the actual generation of $\bm \beta$ is as follows. We generate a random Erdos-Renyi (ER) graph $G$ with $d_G\frac{ (T-1)\log T}{2}$ edges. The rationale behind this value is the following: 
\begin{itemize}
    \item Consider a random graph $G$ drawn according to the ER model where each edge is included in $G$ with probability $p=d_G\frac{\log T}{T}$, independently from every other edge.
    \item The expected number of edges is then $p\binom{T}{2}=\frac{d_GT(T-1)\log T}{2T}$.
    \item Noting that $p=\frac{\log T}{T}$ is a sharp threshold for the connectedness of $G$, by setting $d_G>1$, the resulting graph will almost surely be connected, whereas, by setting $d_G<1$, the resulting graph will almost surely be disconnected. 
    \item In our experiments, we would like to directly control the number of edges in $G$, so we instead sample $G$ uniformly at random from the collection of all graphs which have $d_G\frac{ (T-1)\log T}{2}$ edges.
\end{itemize}
Therefore, $d_G$ controls the density and connectedness of $G$. 

\begin{revi}
We then randomly choose the global support $S$ according to the desired value of $K_{\text{G}}$, i.e., $|S|=K_{\text{G}}$. For each connected component $C$ of $G$, we generate an initial vector of coefficients $\boldsymbol{\beta}^C \in \mathbb{R}^{T\times D}$, satisfying the local sparsity constraint (note that we allow only features from the global support to be selected). To generate each entry in $\boldsymbol{\beta}^C$, for 21 out of 26 problem parameter settings we use $\beta^t_d = (\beta_0)^t_{d} z^t_d$ where $(\beta_0)^t_{d}$ is drawn from $\mathcal{N}(1,0.25)$ truncated at 0.5 and 1.5; and $z^t_d$ is drawn from $\{-1,0,1\}$ at random according to the desired sparsity. In 5 out of 26 problem parameter settings of our synthetic experiments, we consider having pure binary coefficients as is commonly considered in the sparse regression literature \citep{bertsimas2020sparse, hazimeh2020sparse}. To do this, we set $\beta^t_d = z^t_d$. Then, for each vertex $t \in C$, we construct $\boldsymbol{\beta}^{t}$ by perturbing $\boldsymbol{\beta}^{C}$ according to the desired $\sigma_v.$ The desired number of changes in support is performed by randomly replacing features that originally were in the support, with features that were not, at randomly selected vertices from the global support $S$.
\end{revi}

\paragraph*{Design Matrix and Response.} We create the design matrix $\boldsymbol{X} \in \mathbb{R}^{N \times T \times D}$ as follows.  We assume that, for $t\in[T]$, $\bm X^t = (\bm x_1^t , \dots , \bm x_N^t)$ are i.i.d. realizations from a $D$-dimensional zero-mean normal distribution with covariance matrix $\mathbf{\Sigma}$, i.e., $\bm x_n^t \sim \mathcal{N}(\mathbf{0}_D, \mathbf{\Sigma}), n \in [N]$. The covariance matrix $\mathbf{\Sigma}$ is parameterized by the correlation coefficient $\rho_d \in [0, 1]$ as $\Sigma_{ij} = \rho_d^{|i-j|}, \forall i,j \in [D]$. As $\rho_d \rightarrow 1$, the columns of the data matrix $\bm X^t$, i.e., the features, become more alike.

The outcome vectors $\boldsymbol{Y} \in \mathbb{R}^{N \times T}$ are created by applying $\boldsymbol{\beta}$ on $\boldsymbol{X}$ and adding i.i.d. noise drawn from a normal distribution $\mathcal{N}(0, \sigma^2)$ to each entry in $\boldsymbol{Y}$, where $\sigma^2$ is selected to satisfy  $\xi^2 = \frac{\sum_{t\in T} \|\bm{X^t} \bm{\beta^t}\|^2}{\sigma^2}$ according to the desired signal-to-noise ratio $\xi$.

\paragraph*{Evaluation Tasks and Metrics.} Our task is to estimate $\boldsymbol{\beta}$ and make out-of-sample predictions for unseen data $\boldsymbol{X_{\text{test}}} \in \mathbb{R}^{N_{\text{test}} \times T \times D}$ and $\boldsymbol{Y_{\text{test}}} \in \mathbb{R}^{N_{\text{test}} \times T}$, generated according to the same process as $\bm X$ and $\bm Y$. We consider the evaluation metrics shown in Table \ref{tab:metrics}. 
We perform a full sensitivity analysis with respect to the problem parameters $(N,T,D,K_{\text{L}},K_{\text{G}},K_{\text{C}}, \sigma_v, d_G, \rho_d, \xi)$. For each problem parameter setting, we independently generate $10$ datasets and report the mean and standard deviation of the results for each evaluation metric.
\begin{table}[ht!]
\caption{Evaluation Metrics.} \label{tab:metrics}
\centering
\resizebox{\textwidth}{!}{
\begin{tabular}{rl}
\toprule
\textbf{Metric} & \textbf{Explanation} \\ \midrule
MAE & Mean absolute error in estimated coefficients compared to ground truth. \\ 
DS & Differences in support between estimated and ground truth coefficients (expressed in \%). \\ 
MAC & Mean absolute change in coefficients across adjacent vertices. \\ 
Test R$^2$ & Out-of-sample R$^2$ statistic (evaluated on held-out test set). \\  
Time & Computational time (in seconds). Measures time including hyperparameter tuning.  \\ 
Gap & Optimality gap for MIO-based methods. \\ 
Cut Count & Number of cuts generated by cutting plane method. \\ 
ACT  & Average time per cut generated by cutting plane method.\\ \bottomrule
\end{tabular}}
\end{table}

\begin{revi}
\subsection{Experiments on Synthetic Data: Extended Results} \label{sec:appendix-expers-additional}

In this section, we provide extended computational results from our experiments on synthetic data. For each metric, we report results from our sensitivity analysis with respect to each problem parameter, and setting the remaining problem parameters to the following default values: $N=3000, T=10, D=200, K_L=5, K_G=15, K_C=20, \sigma_v=0.33, d_G=3, \rho_d=0.9, \xi=2$. We give the results in Tables \ref{tab:synthetic-results-sensitivity-2} and \ref{tab:synthetic-results-sensitivity-3}. The 21 parameter variations provide the 21 out of 26 problem parameter settings for our main results generated with $\bm{\beta}^C$ following the truncated normal distribution. For the binary $\bm{\beta}^C$, we only consider the 5 parameter variations that changes the number of samples $N$ to reduce computational time. For brevity, we only show the sensitivity analysis results for $R^2$, MAE, DS, and MAC. 

\begin{table}
\centering
\caption{Sensitivity analysis: Test R$^2$}
\label{tab:synthetic-results-sensitivity-1}
\resizebox*{\textwidth}{!}{%
\begin{tabular}{lrrrrrrrr}
\toprule
 & \svc & \svhc & \svh & \sr & \rsnOne & \rsnOneLasso \\ \midrule
T = 5.0 & 0.789 & 0.789 & 0.747 & \textbf{0.631} & 0.768 & 0.763 \\
T = 20.0 & 0.788 & 0.788 & 0.732 & 0.727 & 0.755 & 0.749 \\
$ \xi$ = 0.5 & 0.195 & 0.195 & 0.171 & 0.176 & 0.179 & 0.179 \\
$ \xi$ = 10.0 & 0.985 & 0.977 & 0.905 & 0.876 & 0.95 & 0.945 \\
N = 500.0 & 0.782 & 0.782 & 0.713 & 0.713 & 0.754 & 0.757 \\
N = 1000.0 & 0.784 & 0.78 & 0.733 & 0.708 & 0.756 & 0.755 \\
N = 2000.0 & 0.788 & 0.785 & 0.711 & 0.705 & 0.76 & 0.757 \\
N = 3000.0 & 0.791 & 0.785 & 0.724 & 0.718 & 0.762 & 0.759 \\
N = 5000.0 & 0.791 & 0.79 & 0.732 & 0.717 & 0.763 & 0.758 \\
D = 50.0 & 0.789 & 0.789 & 0.776 & 0.702 & 0.754 & 0.747 \\
D = 500.0 & 0.786 & 0.776 & 0.727 & 0.696 & - & - \\
$K_L$ = 3.0 & 0.789 & 0.789 & 0.755 & 0.68 & 0.766 & 0.764 \\
$K_L$ = 10.0 & 0.792 & 0.791 & 0.633 & 0.72 & 0.753 & 0.743 \\
$K_C$ = 10.0 & 0.787 & 0.787 & 0.734 & 0.703 & 0.761 & 0.755 \\
$K_C$ = 30.0 & 0.789 & 0.787 & 0.727 & 0.709 & 0.755 & 0.751 \\
$d_G$ = 1.0 & 0.784 & 0.783 & 0.717 & 0.513 & 0.778 & 0.775 \\
$d_G$ = 10.0 & 0.79 & 0.79 & 0.746 & 0.705 & 0.742 & 0.567 \\
$\sigma_v$ = 0.1 & 0.792 & 0.78 & 0.681 & 0.732 & 0.78 & 0.776 \\
$\sigma_v$ = 0.67 & 0.776 & 0.775 & 0.725 & 0.623 & 0.734 & 0.731 \\
$\rho_d$ = 0.33 & 0.791 & 0.784 & 0.773 & 0.703 & 0.757 & 0.755 \\
$\rho_d$ = 0.99 & 0.783 & 0.78 & 0.589 & 0.703 & 0.756 & 0.744 \\ \bottomrule
\end{tabular}
}
\end{table}

\begin{table}
\centering
\caption{Sensitivity analysis: MAE}
\label{tab:synthetic-results-sensitivity-2}
\resizebox*{\textwidth}{!}{%
\begin{tabular}{lrrrrrrrr}
\toprule
 & \svc & \svhc & \svh & \sr & \rsnOne & \rsnOneLasso \\ \midrule
T = 5.0 & 0.018 & 0.018 & 0.019 & 0.019 & 0.025 & 0.017 \\
T = 20.0 & 0.017 & 0.018 & 0.018 & 0.016 & 0.02 & 0.017 \\
$ \xi$ =    0.5 & 0.023 & 0.023 & 0.024 & 0.021 & 0.032 & 0.022 \\
$ \xi$ =    10.0 & 0.015 & 0.016 & 0.018 & 0.015 & 0.017 & 0.015 \\
N = 500.0 & 0.019 & 0.019 & 0.02 & 0.017 & 0.03 & 0.016 \\
N = 1000.0 & 0.018 & 0.019 & 0.02 & 0.017 & 0.026 & 0.017 \\
N = 2000.0 & 0.018 & 0.018 & 0.019 & 0.017 & 0.023 & 0.016 \\
N = 3000.0 & 0.018 & 0.018 & 0.019 & 0.018 & 0.022 & 0.017 \\
N = 5000.0 & 0.016 & 0.016 & 0.018 & 0.016 & 0.019 & 0.016 \\
D = 50.0 & 0.064 & 0.064 & 0.067 & 0.062 & 0.064 & 0.064 \\
D = 500.0 & 0.007 & 0.008 & 0.008 & 0.007 & - & - \\
$K_L$ =    3.0 & 0.009 & 0.009 & 0.01 & 0.009 & 0.013 & 0.008 \\
$K_L$ =    10.0 & 0.037 & 0.038 & 0.044 & 0.038 & 0.041 & 0.037 \\
$K_C$ =    10.0 & 0.016 & 0.016 & 0.018 & 0.016 & 0.02 & 0.015 \\
$K_C$ =    30.0 & 0.018 & 0.018 & 0.019 & 0.017 & 0.022 & 0.017 \\
$d_G$ =    1.0 & 0.018 & 0.018 & 0.02 & 0.021 & 0.023 & 0.016 \\
$d_G$ =    10.0 & 0.015 & 0.015 & 0.017 & 0.015 & 0.02 & -0.087 \\
$\sigma_v$ =  0.1 & 0.018 & 0.018 & 0.02 & 0.017 & 0.022 & 0.017 \\
$\sigma_v$ =  0.67 & 0.019 & 0.019 & 0.02 & 0.019 & 0.024 & 0.018 \\
$\rho_d$ =  0.33 & 0.017 & 0.017 & 0.017 & 0.018 & 0.019 & 0.018 \\
$\rho_d$ =  0.99 & 0.02 & 0.019 & 0.024 & 0.015 & 0.03 & 0.016 \\ \bottomrule
\end{tabular}
}
\end{table}

\begin{table}
\centering
\caption{Sensitivity analysis: DS}
\label{tab:synthetic-results-sensitivity-3}
\resizebox*{\textwidth}{!}{%
\begin{tabular}{lrrrrrrrr}
\toprule
 & \svc & \svhc & \svh & \sr & \rsnOne & \rsnOneLasso \\ \midrule
T = 5.0 & 0.14 & 0.14 & 0.143 & 0.045 & 0.97 & 0.015 \\
T = 20.0 & 0.12 & 0.128 & 0.131 & 0.045 & 0.963 & 0.006 \\
$ \xi$ =    0.5 & 0.11 & 0.11 & 0.115 & 0.019 & 0.967 & 0.017 \\
$ \xi$ =    10.0 & 0.13 & 0.134 & 0.133 & 0.04 & 0.954 & 0.005 \\
N = 500.0 & 0.125 & 0.129 & 0.135 & 0.042 & 0.971 & 0.019 \\
N = 1000.0 & 0.142 & 0.141 & 0.146 & 0.045 & 0.968 & 0.015 \\
N = 2000.0 & 0.114 & 0.114 & 0.12 & 0.044 & 0.967 & 0.01 \\
N = 3000.0 & 0.098 & 0.1 & 0.105 & 0.048 & 0.963 & 0.012 \\
N = 5000.0 & 0.111 & 0.119 & 0.124 & 0.053 & 0.96 & 0.009 \\
D = 50.0 & 0.562 & 0.562 & 0.571 & 0.101 & 0.892 & 0.056 \\
D = 500.0 & 0.044 & 0.045 & 0.045 & 0.017 & - & - \\
$K_L$ =    3.0 & 0.106 & 0.104 & 0.104 & 0.033 & 0.975 & 0.007 \\
$K_L$ =    10.0 & 0.132 & 0.139 & 0.172 & 0.103 & 0.942 & 0.016 \\
$K_C$ =    10.0 & 0.118 & 0.134 & 0.14 & 0.042 & 0.963 & 0.007 \\
$K_C$ =    30.0 & 0.115 & 0.125 & 0.132 & 0.06 & 0.966 & 0.013 \\
$d_G$ =    1.0 & 0.139 & 0.146 & 0.157 & 0.069 & 0.968 & 0.005 \\
$d_G$ =    10.0 & 0.129 & 0.139 & 0.145 & 0.06 & 0.962 & - \\
$\sigma_v$ =  0.1 & 0.112 & 0.118 & 0.109 & 0.045 & 0.967 & 0.01 \\
$\sigma_v$ =  0.67 & 0.105 & 0.109 & 0.11 & 0.043 & 0.968 & 0.008 \\
$\rho_d$ =  0.33 & 0.008 & 0.008 & 0.009 & 0.038 & 0.948 & 0.002 \\
$\rho_d$ =  0.99 & 0.128 & 0.115 & 0.136 & 0.063 & 0.972 & 0.036 \\ \bottomrule
\end{tabular}
}
\end{table}

\begin{table}
\centering
\caption{Sensitivity analysis: MAC}
\label{tab:synthetic-results-sensitivity-4}
\resizebox*{\textwidth}{!}{%
\begin{tabular}{lrrrrrrrr}
\toprule
 & \svc & \svhc & \svh & \sr & \rsnOne & \rsnOneLasso \\ \midrule
T = 5.0 & 0.009 & 0.009 & 0.009 & 0.016 & 0.013 & 0.013 \\
T = 20.0 & 0.005 & 0.005 & 0.005 & 0.009 & 0.008 & 0.008 \\
$ \xi$ =    0.5 & 0.008 & 0.008 & 0.008 & 0.01 & 0.01 & 0.01 \\
$ \xi$ =    10.0 & 0.006 & 0.006 & 0.006 & 0.011 & 0.009 & 0.009 \\
N = 500.0 & 0.005 & 0.005 & 0.005 & 0.01 & 0.009 & 0.009 \\
N = 1000.0 & 0.005 & 0.005 & 0.005 & 0.01 & 0.009 & 0.009 \\
N = 2000.0 & 0.005 & 0.005 & 0.005 & 0.01 & 0.009 & 0.009 \\
N = 3000.0 & 0.006 & 0.006 & 0.006 & 0.01 & 0.009 & 0.009 \\
N = 5000.0 & 0.005 & 0.005 & 0.006 & 0.009 & 0.008 & 0.008 \\
D = 50.0 & 0.019 & 0.019 & 0.019 & 0.043 & 0.039 & 0.039 \\
D = 500.0 & 0.003 & 0.003 & 0.003 & 0.004 & - & - \\
$K_L$ =    3.0 & 0.003 & 0.003 & 0.003 & 0.007 & 0.006 & 0.006 \\
$K_L$ =    10.0 & 0.012 & 0.012 & 0.012 & 0.018 & 0.017 & 0.017 \\
$K_C$ =    10.0 & 0.004 & 0.004 & 0.004 & 0.009 & 0.008 & 0.008 \\
$K_C$ =    30.0 & 0.007 & 0.007 & 0.007 & 0.011 & 0.01 & 0.01 \\
$d_G$ =    1.0 & 0.008 & 0.008 & 0.008 & 0.014 & 0.01 & 0.011 \\
$d_G$ =    10.0 & 0.004 & 0.004 & 0.005 & 0.009 & 0.009 & -0.092 \\
$\sigma_v$ =  0.1 & 0.003 & 0.003 & 0.004 & 0.006 & 0.006 & 0.006 \\
$\sigma_v$ =  0.67 & 0.011 & 0.011 & 0.011 & 0.017 & 0.015 & 0.015 \\
$\rho_d$ =  0.33 & 0.007 & 0.008 & 0.008 & 0.011 & 0.01 & 0.01 \\
$\rho_d$ =  0.99 & 0.006 & 0.006 & 0.006 & 0.01 & 0.009 & 0.009 \\ \bottomrule
\end{tabular}
}
\end{table}

\end{revi}

\section{Extended Numerical Experiments on Real-World Data} \label{sec:appendix-expers-real}

In this section, we provide a more detailed discussion of our computational study on real-world data. First, we give more information on the datasets and the preprocessing methodology we apply to each of them. Then, we present the detailed computational results for each dataset, method, and metric combination, which we use to extract the aggregated results shown in Table \ref{tab:real-results}.

\subsection{Datasets and Preprocessing Methodology} \label{ssec:appendix-expers-real-data}
We begin our discussion by outlining the details of the real-world datasets we use in our experiments and the preprocessing methodology we apply to each of them. 

As discussed in Section \ref{sec:realworld}, we randomly split each dataset 10 times into training ($60\%$), validation ($20\%$), and test ($20\%$) sets (respecting the temporal structure if such exists). In all cases, we use the training set to normalize both the validation and the test sets' data matrices $\bm X$ and responses $\bm Y$, so that all features and responses have zero mean and unit variance.

\paragraph{Appliances Energy Prediction: Hourly.} In this experiment, we focus on a real-world case study concerned with appliances energy prediction \citep{candanedo2017data}. The dataset is publicly available at the University of California Irvine (UCI) Machine Learning repository, at \url{https://archive.ics.uci.edu/ml/datasets/Appliances+energy+prediction}. 

Each observation in the dataset is a vector of measurements made by a wireless sensor network in a low-energy building. The features include the temperature and humidity conditions in various rooms in the building, the weather conditions in the nearest weather station, the month in which the measurements were taken, and a couple of noise variables. The goal is to predict the energy consumption of the building's appliances. Measurements are taken every 10 minutes over a 4.5-month period.

We preprocess the dataset as follows. We construct the similarity graph by assigning a vertex to each hour of the day so that $T=24$. To capture the temporal structure in the problem, the graph is a chain, i.e., vertex $t \in [T-1]$ is considered adjacent to vertex $t+1$. For each day $d$ in the data, we create $6$ data points per vertex $t \in [T]$, by collecting all $6$ measurements that were taken at hour $t$ and during day $d$. For example, for $t=15$, we collect the measurements taken at 3pm, 3:10pm, $\dots$, 3:50pm, across all days in the data. By doing so, we get $N = 822$ data points per vertex. Each data point consists of $D = 26$ features. The decision to split the data hourly was the most natural, but it remains an arbitrary decision. The model can be applied to any subdivision depending on the goal of the regression, and this splitting can also be hyper-parameter-tuned for further performance improvement.

\paragraph{Appliances Energy Prediction: Monthly.} In this experiment, we consider the same appliances energy prediction dataset. However, instead of assigning a vertex to each hour of the day, we now assign a vertex to each month in the data, so that $T=5$ (the data covers a 4.5-month period between January and May). For each measurement, we replace in the feature set the month with the hour at which the measurement was taken. Once again, to capture the temporal structure in the problem, we take the similarity graph to be a chain. We now collect all measurements taken during each month as data points for the corresponding vertex, and subsample $N=2,922$ data points so that we get the same $N$ across all months. 

\paragraph{Housing Price Prediction.} In this experiment, we explore the application of our framework to the task of housing price prediction, in Ames, Iowa \citep{de2011ames}. The dataset is publicly available at \url{http://jse.amstat.org/v19n3/decock/DataDocumentation.txt}. 

The original dataset contains $2,930$ observations and a large number of features (23 nominal, 23 ordinal, 14 discrete, and 20 continuous) involved in assessing home values. The sales took place in Ames, Iowa, from 2006 to 2010. The goal is to predict the price at which the house was sold.

We preprocess the dataset as follows. We first drop features with missing values in over $1\%$ of the observations; then, we drop any observation that still has missing features. We use one-hot encoding for the nominal features and integer encoding for ordinal and discrete variables. Each data point consists of $D=199$ features. The dataset contains information on the neighborhood where each house is located, so we could have used these neighborhoods as the vertices in the similarity graph. Nevertheless, such an approach leads to highly imbalanced vertices in terms of the number of data points that fall therein (due to the fact that many sales were performed in some neighborhoods and very few at others). To address this issue, we cluster the neighborhoods into larger groups while requiring that neighborhoods that fall into the same group be adjacent and that the number of data points that fall into each group be relatively balanced. Then, we construct the similarity graph by adding an edge between groups of neighborhoods that are adjacent. In the end, we obtain $T=7$ groups of neighborhoods, each with at least $N=352$ data points (for simplicity, we randomly select exactly $N=352$ data points in each group). The similarity graph consists of $E=8$ edges. 

\paragraph{Air Quality.} In this experiment, we consider the task of air quality prediction, in Beijing \citep{zhang2017cautionary}. The dataset is publicly available at \url{https://archive.ics.uci.edu/ml/datasets/Beijing+Multi-Site+Air-Quality+Data}. 

The original dataset consists of 420,768 observations and features: 5 numerical features (temperature, pressure, dew point temperature, precipitation, wind speed), 1 categorical feature (wind direction), and 3 time-related features. The goal is to predict PM2.5 concentration - an air pollutant that is a concern for people's health when levels in air are high. The data is collected from 12 nationally controlled air quality monitoring sites. The meteorological data in each air quality site are matched with the nearest weather station from the China Meteorological Administration. The time period is from March 1st, 2013, to February 28th, 2017.

We preprocess the dataset as follows. We construct the similarity graph by assigning a vertex to each air quality monitoring station so that $T=12$. We add an edge between each station and the closest station towards each direction (east, north, south, west), provided that their distance does not exceed a pre-defined threshold, for a total of $E=14$ edges. The resulting similarity graph is disconnected and consists of $4$ connected components. We get $N=35,064$ data points per vertex. After one-hot encoding of the wind direction categorical feature into 17 binary features, we get a total of $D=25$ features. We finally perform mean imputation. 

\paragraph{Meteorology.} In this experiment, we consider the task of weather prediction. The dataset is publicly available at \url{https://www.kaggle.com/datasets/selfishgene/historical-hourly-weather-data}. 

The original dataset contains about 5 years of hourly measurements of various weather attributes, including temperature, humidity, air pressure, wind direction, and wind speed. The goal is to predict the temperature half a day in advance. This data is collected from 30 US and Canadian Cities, as well as 6 Israeli cities.

We preprocess the dataset as follows. We construct the similarity graph by assigning a vertex to each city so that $T=36$. We add an edge between two cities provided that their (euclidean) distance is less than $1,000$ kilometers, for a total of $E=110$ edges. The resulting similarity graph is disconnected and consists of $2$ connected components. We get $N=45,231$ data points per vertex. Each data point consists of measurements of the 5 aforementioned weather attributes over the past 10 hours, for a total of $D=50$ features. We note that we made a number of arbitrary decisions in preprocessing the data, including predicting half a day ahead (predicting fewer hours ahead led to extremely high R$^2$ simply by outputting the current temperature value), setting the distance threshold to 1,000 kilometers, and considering the past 10 measurements for each weather attribute; in all cases, we set the above values to what seemed the most natural choice.

\subsection{Experiments on Real-World Data: Extended Results} \label{ssec:appendix-expers-real-results}

In this section, we provide extended computational results from our experiments on real-world data. In particular, for each dataset-method pair, we give the mean and standard deviation of each metric discussed in the aggregated results of Table \ref{tab:real-results}.

\begin{revi}
Table \ref{tab:real-r2} summarizes, for each dataset-method pair, the averaged (across all 10 training-validation-test splits) out-of-sample R$^2$ results. Here, \rsnOne\ and \rsnTwo\ perform best (although \rsnTwo\ still faces serious scalability issues), closely followed by \svc. We note that \svc\ usually improves upon the solution found by \svh, and that \svh performs surprisingly well, outperforming \sr\ and the lasso-based regularized methods. The benefits that sparsity can provide on generalization can be seen through the housing case study, where \rsnOne\ and \rsnOneLasso\ produce much denser models, which fail to generalize out-of-sample.

\begin{table}[ht!]
\caption{Real-world data experiments: out-of-sample R$^2$.}
\label{tab:real-r2}
\resizebox{\textwidth}{!}{
\begin{tabular}{lrrrrr}
\toprule
\textbf{Algorithm} & \textbf{Air Quality} & \textbf{Appliances Energy (hour)} & \textbf{Appliances Energy (month)} & \textbf{Housing Price} & \textbf{Meteorology} \\ \midrule
\svc & 0.515 (0.002) & 0.828 (0.054) & 0.489 (0.018) & 0.949 (0.009) & 0.838 (0.0) \\
\svhc & 0.515 (0.002) & 0.828 (0.054) & 0.491 (0.02) & 0.949 (0.009) & 0.837 (0.0) \\
\svh & 0.515 (0.002) & 0.824 (0.055) & 0.487 (0.021) & 0.949 (0.009) & 0.816 (0.0) \\
\sr & 0.512 (0.002) & 0.592 (0.033) & 0.441 (0.018) & 0.759 (0.4) & 0.83 (0.0) \\
\rsnOne & 0.516 (0.002) & 0.832 (0.054) & 0.495 (0.017) & 0.621 (0.925) & 0.846 (0.0) \\
\rsnOneLasso & 0.495 (0.002) & 0.826 (0.055) & 0.478 (0.02) & 0.663 (0.928) & 0.831 (0.0) \\
\rsnTwo & 0.516 (0.002) & 0.832 (0.053) & 0.496 (0.017) & 0.94 (0.048) & - \\
\rsnTwoLasso & 0.495 (0.002) & 0.826 (0.055) & 0.478 (0.021) & 0.955 (0.012) & - \\ \bottomrule
\end{tabular}
}
\end{table}

In Table \ref{tab:real-sparsity}, we assess, for each dataset-method pair, the learned models' interpretability, through the (average) estimated sparsity-related hyperparameters $\hat K_\text{L}$, $\hat K_\text{G}$, $\hat K_\text{C}$. The edge of \svc\ and \svh\ among the slowly varying methods is evident: the learned models are significantly sparser and hence more interpretable while achieving comparable or even improved predictive performance.

\begin{table}[ht!]
\caption{Real-world data experiments: model sparsity ($\hat K_\text{L}$, $\hat K_\text{G}$, $\hat K_\text{C}$).}
\label{tab:real-sparsity}
\resizebox{\textwidth}{!}{
\begin{tabular}{lrrrrr}
\toprule
\textbf{Algorithm} & \textbf{Air Quality} & \textbf{Appliances Energy (hour)} & \textbf{Appliances Energy (month)} & \textbf{Housing Price} & \textbf{Meteorology} \\ \midrule
\svc & \textbf{7.0 (0.0) | 7.0 (0.0)   | 0.0 (0.0)} & \textbf{12.4 (1.96) | 12.8 (1.81) | 0.8 (1.69)} & 21.9 (2.51) | 22.8 (2.3) | 2.3 (3.56) & \textbf{9.1 (0.32) | 9.1 (0.32) | 0.0 (0.0)} & 18.0 (0.0) | 25.0   (0.0) | 0.0 (0.0) \\
\svhc & \textbf{7.0 (0.0) | 7.0 (0.0) | 0.0 (0.0)} & \textbf{12.4 (1.96) | 12.8 (1.81) | 0.8 (1.69)} & 24.8 (3.49) | 25.8 (2.2) | 2.6 (3.47) & \textbf{9.1 (0.32) | 9.1 (0.32) | 0.0 (0.0)} & \textbf{10.0 (0.0) | 10.0 (0.0) | 0.0 (0.0)} \\
\svh & \textbf{7.0 (0.0) | 7.0 (0.0) | 0.0 (0.0)} & \textbf{12.4 (1.96) | 12.8 (1.81) | 0.8 (1.69)} & 24.3 (3.95) | 25.2 (2.97) | 2.7 (3.4) & \textbf{9.1 (0.32) | 9.1 (0.32) | 0.0 (0.0)} & \textbf{10.0 (0.0) | 10.0 (0.0) | 0.0 (0.0)} \\
\sr & 5.7 (0.82) | 5.7 (0.82) | 0.0 (0.0) & 22.4 (2.84) | 22.4 (2.84) | 0.0 (0.0) & \textbf{15.2 (8.82) | 15.2 (8.82) | 0.0 (0.0)} & 48.5 (8.83) | 48.5 (8.83) | 0.0 (0.0) & 50.0 (0.0) | 50.0 (0.0) | 0.0 (0.0) \\
\rsnOne & 25.0 (0.0) | 25.0 (0.0) | 16.6 (10.51) & 28.0 (0.0) | 28.0 (0.0) | 0.6 (1.35) & 28.0 (0.0) | 28.0 (0.0) | 0.0 (0.0) & 172.0 (60.46) | 172.3 (60.56) | 4.3 (3.68) & 50.0 (0.0) | 50.0 (0.0) | 31.0 (0.0) \\
\rsnOneLasso & 15.0 (0.82) | 17.4 (1.17) | 24.1 (9.54) & 13.3 (1.06) | 27.4 (0.7) | 173.4 (30.73) & 15.3 (1.06) | 22.2 (1.55) | 37.0 (5.72) & 44.5 (20.39) | 98.8 (45.24) | 272.7 (140.41) & 24.0 (0.0) | 43.0 (0.0) | 965.0 (0.0) \\
\rsnTwo & 25.0 (0.0) | 25.0 (0.0) | 39.5 (7.79) & 28.0 (0.0) | 28.0 (0.0) | 1.4 (2.12) & 28.0 (0.0) | 28.0 (0.0) | 0.0 (0.0) & 191.9 (1.6) | 191.9 (1.6) | 12.8 (3.91) & - \\
\rsnTwoLasso & 14.8 (0.42) | 20.2 (1.32) | 49.9 (7.43) & 14.0 (1.33) | 27.5 (0.71) | 203.7 (12.23) & 15.9 (1.29) | 22.3 (1.7) | 38.0 (5.4) & 54.6 (16.8) | 125.8 (35.61) | 390.8 (135.69) & - \\ \bottomrule
\end{tabular}
}
\end{table}

Table \ref{tab:real-time} reports, for each dataset-method pair, the corresponding average computational time in seconds for each dataset-method pair. \svh\ is again the clear winner. The MIO-based methods time out without proving optimality only in one case.

\begin{table}[ht!]
\caption{Real-world data experiments: computational time (in seconds).}
\label{tab:real-time}
\resizebox{\textwidth}{!}{
\begin{tabular}{lrrrrr}
\toprule
\textbf{Algorithm} & \textbf{Air Quality} & \textbf{Appliances Energy (hour)} & \textbf{Appliances Energy (month)} & \textbf{Housing Price} & \textbf{Meteorology} \\ \midrule
\svc & 5899.828 (2718.802) & 5628.365 (2671.271) & 577.013 (278.963) & 3082.032 (760.444) & 30280.938 (1069.892) \\
\svhc & 232.601 (91.224) & 270.152 (162.101) & 118.94 (71.951) & 107.238 (58.342) & 1254.121 (40.703) \\
\svh & 199.779 (1.438) & 164.32 (45.394) & 134.511 (69.629) & 92.79 (25.58) & 964.328 (15.504) \\
\sr & 3480.81 (345.824) & 3652.116 (534.893) & 3842.325 (471.714) & 8175.013 (5.826) & 6562.15 (401.019) \\
\rsnOne & 1559.888 (83.519) & 118.372 (7.512) & 64.287 (2.118) & 357.147 (119.129) & 33693.443 (1269.554) \\
\rsnOneLasso & 1565.859 (74.407) & 114.882 (10.162) & 58.797 (1.887) & 707.304 (938.539) & 35480.862 (1492.15) \\
\rsnTwo & 12195.662 (448.928) & 3516.378 (628.868) & 1310.701 (240.861) & 3689.827 (381.796) & - \\
\rsnTwoLasso & 5833.191 (191.863) & 349.423 (24.617) & 184.182 (7.518) & 791.322 (68.882) & - \\ \bottomrule
\end{tabular}
}
\end{table}
\end{revi}

\section{Numerical Experiments for Section \ref{sec:relaxation}: Testing Different Relaxations} \label{sec:test-different-relaxation}
In this section, we test the performance of the following (extended) convex relaxation of the closed form solution given in Equation \eqref{eq:beta_star_pseudoinv}:
\begin{equation} \label{eq:beta_star_extended_convex}
(\bm{Z}(\bm{M}+\lambda_\beta \bm{I})\bm{Z})^\dagger\bm{Z} = (\lambda_\beta \bm{I}+  \bm{Z}\bm{M})^{-1}\bm{Z} + \mu\left(\sum_{t=1}^{T} \sum_{d=1}^{D} \left(z_d^t-\frac{1}{2}\right)^2 - \frac{TD}{4}\right),
\end{equation}
for  $\mu \in \{0, 0.1, 1, 2, 5\}$, using synthetic data. 

Specifically, we set $N=1000$, $T=1$, $D=100$, $K_{\text{L}}=5$, $\lambda_\beta=10$, hence focusing on the standard sparse regression problem. We generate data of the form $\bm{y}=\bm{X}_0\bm{\beta}_0 +\bm{\varepsilon}$, where $\bm{X}_0 \in \md{R}^{N \times K_L}$ and $(\bm X_0)_{n,d}\sim N(0,1)$, $\bm{\beta}_0 \in \md{R}^{K_L}$ and $(\bm \beta_0)_{d}\sim N(0,1)$, $\bm{\varepsilon} \in \md{R}^{N}$ and $(\bm \varepsilon)_{n}\sim N(0,0.1)$, and we set $\bm{X}=[\bm{X}_0,\bm{Z}]$, where $\bm{Z} \in \md{R}^{N \times (D-K_L)}$ and $(\bm Z)_{n,d}\sim N(0,1)$. We generate 20 instances of the synthetic data.

For each instance, we run 5 versions of Algorithm \ref{alg:cutplane}: in each version, we solve the inner problem using Equation \eqref{eq:beta_star_extended_convex} and a different value for $\mu \in \{0, 0.1, 1, 2, 5\}$. We record the computational time in seconds and the mean absolute error (MAE) in the estimated coefficients, as detailed in Table \ref{tab:relaxation_table}. 
We observe that as $\mu$ increases, both the computational time and MAE suffer, suggesting that such family of relaxations is unlikely to produce stronger cuts than the baseline $\mu=0$ relaxation.
\begin{table}[ht!]
    \caption{Results for extended convex relaxations.}
    \label{tab:relaxation_table}
\centering
    \resizebox{0.13\textwidth}{!}{
    \centering
    \begin{tabular}{lrr}
    \toprule
         $\bm{\mu}$ & \textbf{Time} &  \textbf{MAE}  \\\midrule
         0 & 0.19 & 0.056\\
         0.1 & 0.27 & 0.069\\
         1 & 0.97 & 0.128\\
         2 & 1.61 & 0.112\\
         5 & 10.60 & 0.150\\\bottomrule       
    \end{tabular}
    }
\end{table}

\section{Numerical Experiments for Section \ref{sec:heuristic}: Integrality Test of Algorithm \ref{alg:heuristic}} \label{sec:integrality-test-heuristic-algorithm}
In this section, we test the integrality of the solutions obtained by the second step of Algorithm \ref{alg:heuristic}, that is, the linear relaxation of Problem \eqref{eq:heuristic-closed-form}, in the temporal case with $T$ time periods (see Figure \ref{fig:similarity-graphs}), and using synthetic data. 

Note that, in this setting, the linear relaxation that Algorithm \ref{alg:heuristic} solves can be written as:
\begin{align*}
    & \min_{\bm{z}, \bm s, \bm w} && \sum_{t \in [T],d\in[D]} L^t_d z^t_d\\
    &\text{s.t.}&& z^t_d \leq s_d & & \forall t \in [T],\ d \in [D]\\
    &&& z^{t+1}_d-z^t_d \leq w^t_d & & \forall t \in [T-1],\ d \in [D]\\       
    &&& z^{t}_d-z^{t+1}_d \leq w^t_d & & \forall t \in [T-1],\ d \in [D]\\  
    &&& \sum_{d \in [D]} s_d \leq K_G & & \forall t \in [T],\ d \in [D]\\ 
    && &\sum_{d \in [D]} z^t_d = K_L & & \forall t \in [T]\\
    &&& \sum_{t \in [T-1],d\in[D]} w^t_d\leq K_C & & \forall t \in [T-1],\ d \in [D]\\       
    &&&0\leq z^t_d\leq 1 &&\forall t\in [T],\ d \in [D]\\
    &&&0\leq w^t_d\leq 1 &&\forall t\in [T-1],\ d \in [D]\\
    &&&0\leq s_d\leq 1 &&\forall t\in [T],\ d \in [D]
\end{align*}

We test 3 values for each parameter, namely, $T,D \in \{2,5,10\}$, for a total of 9 combinations. For each combination, we generate the remaining parameters as follows:
\begin{itemize}
    \item $K_L$ is selected uniformly within $[\lfloor D/4 \rfloor, \lfloor D/2 \rfloor]$.
    \item $K_G$ is selected uniformly within $[\lfloor 1.5K_L\rfloor, \lfloor 2.5K_L \rfloor]$.
    \item $K_C=2(K_G-K_L)$ to allow some slack in selecting what variables can be chosen to satisfy the slowly varying constraint. 
    \item The loss grid $L^t_d$ is generated using two different methods:
    \begin{itemize}
        \item \textbf{Uniform}: $L^t_d \sim U[0,1]$.
        \item \textbf{Correlated}: $L^t_d \begin{cases}
        \sim U[0,D-d+1] & t \leq \lfloor T/2\rfloor \\ \sim U[0,d] & t > \lfloor T/2\rfloor\end{cases}$. 
        This simulates a cost function where features with larger indices are more predictive for time periods $\leq \lfloor T/2\rfloor$ and features with smaller indices are more predictive for time periods $\geq \lfloor T/2\rfloor$.
    \end{itemize}
\end{itemize}
For each $T,D$ combination, we simulate the remaining parameters 100 times and record both the percentage of fully integral solutions and the percentage of integral variables obtained by the resulting linear optimization problem. The results are shown in Table \ref{tab:integrality_table} where Full Int. means fully integral solutions and Int. Var. represents the percentage of variables that take integral values. We see that across all experiments and all types of cost grids, a significant portion of the solutions are integral. We further note that, in the case of non-integral solutions, the portion of non-integral entries in $\bm z$ is small (always less than $15\%$ and, typically, even smaller).
\begin{table}[ht!]
    \caption{Results for integrality test of Algorithm \ref{alg:heuristic}.}
    \label{tab:integrality_table}
\centering
\resizebox{0.5\textwidth}{!}{
    \centering
    \begin{tabular}{rrrrrr}
    \toprule
         \multirow{2}{*}{$\bm{D}$} & \multirow{2}{*}{$\bm{T}$} &  \multicolumn{2}{c}{\textbf{Uniform}} & \multicolumn{2}{c}{\textbf{Correlated}} \\
         &&
         \textbf{Full Int.} $\bm{(\%)}$  &  \textbf{Int. Var} $\bm{(\%)}$ & \textbf{Full Int.} $\bm{(\%)}$  &  \textbf{Int. Var} $\bm{(\%)}$  \\\midrule
          2 & 2 & 100 & 100 & 100 & 100\\
         2 & 5 & 69 & 92.2& 63 & 92.8\\
         2 & 10 & 46 & 91.7& 51 & 91.0\\
         5 & 2 & 100 & 100& 100 & 100\\
         5 & 5 & 37 & 88.6& 47 & 92.4\\
         5 & 10 & 8 & 85.1& 9 & 86.5\\
         10 & 2 & 100 & 100& 100 & 100\\
         10 & 5 & 42 & 95.1& 47 & 95.7\\
         10 & 10 & 24& 93.7& 35 & 94.1\\\bottomrule         
    \end{tabular}
    }
\end{table}

\end{APPENDICES}

\end{document}